\documentclass[journal,compsoc]{IEEEtran}
\IEEEoverridecommandlockouts
\usepackage[tight,small]{subfigure}
\usepackage[figuresright]{rotating}
\usepackage{threeparttable}
\usepackage{amssymb}
\usepackage[figuresright]{rotating}
\usepackage{amsfonts}
\usepackage{mathrsfs}
\usepackage{pifont}
\usepackage{amsmath}
\usepackage{amssymb}
\usepackage{graphicx}
\usepackage{multicol}
\usepackage{multirow}
\usepackage{cases}
\usepackage{epstopdf}
\usepackage{amsthm}
\newcommand{\bm}[1]{\mbox{\boldmath{$#1$}}}
\usepackage[tight,small]{subfigure}
\usepackage{ltxtable, filecontents}
\usepackage{algorithmic}
\usepackage{booktabs}
\usepackage{tabularx}
\usepackage{supertabular}
\usepackage{stfloats}
\usepackage{cite}
\usepackage{bm}
\usepackage{float}
\usepackage{diagbox}

\usepackage{graphicx}  
\usepackage{float}  
\usepackage{subfigure}  

\usepackage{bbm}

\usepackage[ruled]{algorithm2e}

\ifCLASSINFOpdf
\else
\fi
\begin{document}

\title{FedDD: Toward Communication-efficient Federated Learning with Differential Parameter Dropout}

\author{Zhiying Feng,
        Xu Chen,~\IEEEmembership{Senior Member,~IEEE,}
        Qiong Wu,
        Wen Wu,~\IEEEmembership{Senior Member,~IEEE,}
        Xiaoxi Zhang,~\IEEEmembership{Member,~IEEE,}
        and~Qianyi Huang,~\IEEEmembership{Member,~IEEE}
        
\IEEEcompsocitemizethanks{\IEEEcompsocthanksitem Zhiying Feng, Xu Chen, Qiong Wu, Xiaoxi Zhang, and Qianyi Huang are with the School of Computer Science and Engineering, Sun Yat-sen University, Guangzhou, 510006, China.  \IEEEcompsocthanksitem Wen Wu is with Peng Cheng Laboratory, Shenzhen, 518055, China.      \IEEEcompsocthanksitem  The corresponding author: Xu Chen.} }

%
%

\markboth{}%
{Shell \MakeLowercase{\textit{et al.}}: Bare Demo of IEEEtran.cls for Computer Society Journals}
%



\IEEEtitleabstractindextext{
\begin{abstract}
Federated Learning (FL) requires frequent exchange of model parameters, which leads to long communication delay, especially when the network environments of clients vary greatly. Moreover, the parameter server needs to wait for the slowest client (i.e., straggler, which may have the largest model size, lowest computing capability or worst network condition) to upload parameters, which may significantly degrade the communication efficiency. Commonly-used client selection methods such as partial client selection would lead to the waste of computing resources and weaken the generalization of the global model. To tackle this problem, along a different line, in this paper, we advocate the approach of model parameter dropout instead of client selection, and accordingly propose a novel framework of Federated learning scheme with Differential parameter Dropout (FedDD). FedDD consists of two key modules: dropout rate allocation and uploaded parameter selection, which will optimize the model parameter uploading ratios tailored to different clients' heterogeneous conditions  and also select the proper set of important model parameters for uploading subject to clients' dropout rate constraints. Specifically, the dropout rate allocation is formulated as a convex optimization problem, taking system heterogeneity, data heterogeneity, and model heterogeneity among clients into consideration. The uploaded parameter selection strategy prioritizes on eliciting important parameters for uploading to speedup convergence. Furthermore, we theoretically analyze the convergence of the proposed FedDD scheme. Extensive performance evaluations demonstrate that the proposed FedDD scheme can achieve outstanding performances in both communication efficiency and model convergence, and also possesses a strong generalization capability to data of rare classes.

\end{abstract}

\begin{IEEEkeywords}
Federated learning, parameter dropout, communication-efficient, parameter selection.
\end{IEEEkeywords}}

 \maketitle

\IEEEdisplaynontitleabstractindextext

%
\IEEEpeerreviewmaketitle

\IEEEraisesectionheading{\section{Introduction}\label{sec:introduction}}
\IEEEPARstart{W}{ith} the widespread adoption of Internet of Things (IoT) devices and the rise of edge computing, a large amount of data is generated on terminal devices or edge devices\cite{ma2021fedsa}. In order to avoid uploading a large amount of private data to the cloud server for centralized AI model training, Mobile Edge Computing (MEC) is proposed to migrate the AI function from the cloud to edge\cite{wu2020fedscr}, and improved hardware performance makes training AI models on mobile devices possible. However, traditional distributed machine learning methods are difficult to meet the requirements of data privacy and security\cite{9660377}. Federated Learning (FL)\cite{mcmahan2017communication} is a distributed machine learning scheme rising in recent years that fits well with data privacy preserving requirement, which enables training of a machine learning model in a distributed network by only sharing their local model updates instead of raw data sharing\cite{wu2022communication},\cite{han2021fedmes}.

Although FL brings benefits in terms of data privacy, it comes at the expense of efficiency, as FL requires frequent exchange of model parameters between the server and clients. Furthermore, due to the heterogeneity in geographic distributions and networking conditions between the clients and the server, the communication time varies significantly even when transmitting models of the same size. In classical FL, all the clients need to synchronize and wait for the slowest one to complete the parameter upload before starting a new round of training, which further degrades the communication efficiency. This phenomenon is called the straggler effect\cite{li2020federated}. To tackle the straggler effect, some researchers change the communication mode from synchronous to asynchronous. But asynchronous FL will bring other problems, such as the staleness effect. To avoid the negative impact of the staleness effect, our research focus is still on synchronous FL.

In synchronous FL, to improve the communication efficiency, many existing efforts resort to the mechanisms of client selection, in which only a portion of clients are selected to participate in global update \cite{nishio2019client,wang2020optimizing,lai2021oort,xu2020client,bonawitz2019towards}. FL based on client selection reduces the number of participating clients based on different rules, including clients being dropout randomly\cite{bonawitz2019towards}, and clients with long processing time\cite{nishio2019client,xu2020client} or low self-defined utility\cite{wang2020optimizing,lai2021oort} being dropout. FL with partial client participation can effectively improve communication efficiency and mitigate the straggler effect, but it inevitably wastes a portion of costly computation resource\cite{ma2021fedsa}. Some clients hold in idle for a long time and have less chance to participate in global updates. Essentially, FL based on client selection reduces the amount of data that the global model can learn indirectly, which is not conducive to global convergence\cite{wang2021federated}. What's more, client selection could further exacerbate data heterogeneity, and even may lead to the absence of certain labels of data, which would weaken the generalization of global model.

In order to overcome the deficiencies of \emph{client selection}, we advocate a different paradigm of \emph{parameter dropout} in this paper, and introduce a novel communication-efficient Federated learning with Differential parameter Dropout (FedDD). In our scheme, the parameter server sets the minimum amount of parameters required for model aggregation for different clients tailored to their heterogeneous conditions. Specifically, the clients will upload sparse models via client-specific model parameter dropout instead of full models to the server according to their assigned parameter dropout rates. FedDD can improve communication efficiency, meanwhile maximizing the utilization of all computing resources without client selection. Furthermore, compared with client selection schemes, FedDD ensures more clients participating in global update and hence greatly boosts the generalization of the learned global model.

Nevertheless, performing differential parameter dropout is challenging, especially in the face of multiple dimensions of heterogeneity, including system heterogeneity (i.e., differences of communication latency and training latency of clients), data heterogeneity (i.e., difference of local data distribution) and model heterogeneity (i.e., different structures of clients' local models). This motivates us to study two key problems: 1) How to determine the proper dropout rates of different clients, with limited allowable parameter transmission amount; 2) Given a specific dropout rate for a client, which parameters are selected for uploading, aiming to speed up convergence of global model.

To answer the above two problems, in this paper we design two key modules, namely dropout rate allocation and uploaded parameter selection. In dropout rate allocation module, we assign lower dropout rates to the clients with low communication latency, low training latency, and high data quality, in order to improve the total communication efficiency while limiting the performance degradation of global model. In the uploaded parameter selection module, model parameters' contribution for global model convergence are theoretically analyzed, and the model parameters with greater contribution are selected for uploading. It is worth noting that, in order to adapt to the dynamic contribution change of local models during FL training process, the dropout rate and uploaded model parameters of each client can be adaptively tuned in each global FL training round instead of static configuration.

In light of the above discussion, we list the main contributions of this paper as follows.

$\bullet$ We propose FedDD, a novel communication-efficient FL scheme with differential parameter dropout, in which dropout rate allocation module and uploaded parameter selection module are designed to enable different clients to dynamically transmit differential sparse models for boosting both the model accuracy and communication efficiency. Theoretical bias brought by differential parameter dropout are analyzed, and the convergence bound is mathematically derived.

$\bullet$ In the dropout rate allocation module, we formulate an efficient convex optimization problem to adaptively assign different dropout rates to heterogeneous clients by taking system heterogeneity, data heterogeneity, and model heterogeneity into consideration, in order to achieve optimized time-to-accuracy performance. In the uploaded parameter selection module, we analyze the importance of the model parameters, and devise an indicator function to select the proper set of important parameters to upload tailored to clients' dropout rate constraints, so as to speed up the convergence. 

$\bullet$ Extensive experiments based on various popular FL datasets demonstrate that FedDD outperforms other state-of-the-art schemes in model accuracy, communication efficiency, and generalization ability. For instance, compared with the FedAvg benchmarks, FedDD can achieve more than 75\% training time reduction in reaching a target accuracy. Moreover, FedDD is much more robust to the diverse communication budget changes, compared with the client-selection-based FL baselines.

The rest of this paper is organized as follows.  Section \ref{section:Preliminaries and Motivation} presents the preliminaries and motivation. Section \ref{section:Methodology} introduces the overview of the proposed FedDD framework. Section \ref{section:Design Optimization for FedDD} presents the details of design optimization for FedDD. In Section \ref{section:Convergence analysis}, we analyse the convergence of FedDD. Section \ref{Experiments} shows the extensive experiments. Section \ref{section:Related Work} reviews the related work. We finally conclude the paper in Section \ref{section:Conclusion and Discussion}.

\section{Preliminaries and Motivation}\label{section:Preliminaries and Motivation}
Federated learning trains a global model with the help of multiple client devices, where a parameter server serving as an aggregator coordinates client model learning. FL usually consists of one parameter server and $N$ clients. Each client $n$ holds local dataset $\mathcal{D}_n$. Let $m_{n}$ be the amount of data samples of client $n$. Here, $m = \sum_{n=1}^{N}m_{n}$ is the total number of data samples. The learning objective of FL can be mathematically formulated as follows.  

\begin{equation}\label{eq:flfmin}
\begin{aligned}
\min_{\bm{W}} F(\bm{W}) = \min_{\bm{W}}\sum_{n=1}^{N} \frac{m_{n}}{m}F_{n}(\bm{W}),
\end{aligned}
\end{equation}
where $F_{n}(\bm{W}) = \frac{1}{m_n}\sum_{j\in \mathcal{D}_n}f_j(\bm{W})$ is the local loss function of client $n$.

Each client performs local update based on the same global model downloaded from the server according to Eq. (\ref{eq:fllocalupdate}). Then clients upload the model parameters to the server for aggregation as Eq. (\ref{eq:flaggregation}) shows. The server and clients keep exchanging the model parameters until the target global round is reached or the global model reaches a desired accuracy.
\begin{equation}\label{eq:fllocalupdate}
\begin{aligned}
\hat{\bm{W}_{n}^t}= \bm{W}_{n}^t-\eta  \nabla  F_{n}(\bm{W}_{n}^t).
\end{aligned}
\end{equation}

\begin{equation}\label{eq:flaggregation}
\begin{aligned}
\bm{W}^{t+1}=\frac{1}{m}\sum_{n=1}^{N}m_n\hat{\bm{W}_{n}^t}.
\end{aligned}
\end{equation}

Traditional FL typically requires all clients to upload and download the full model parameters. The uploading and downloading of parameters in the above process will take up a lot of communication resources and therefore the number of participating clients can be limited under limited communication resources. Moreover, the data transmission speed of different clients can vary greatly, which makes the clients and the server waste a lot of time waiting for the slowest clients to finish transmitting model parameters. These observations motivate us to consider two problems: how to involve more clients to participate in FL and how to mitigate the straggler effect. For the former problem, we intend to let clients upload and download sparse models via parameter dropout instead of the full models for data communication overhead reduction. For the latter problem, we will devise efficient differential parameter dropout scheme by assigning different dropout rates for different clients, such that capable clients (with stronger computation ability, faster transmission speed, and higher data quality) would handle more tasks of model parameter upload. We will further conduct fine-grained important model parameter selection for a client, in order to boost the convergence performance given its dropout rate constraint.

We should emphasize that the proposed parameter dropout scheme in FedDD is different from the structure sparsity schemes such as model pruning \cite{li2021hermes} in FL. For FL with model pruning, the structures of clients' local models keep changing as training progresses, aiming at training personalized models for different clients. While, in FedDD, the structure of clients' local models do not change (only the uploaded model parameters change via differential and dynamic parameter dropout), because our goal is to train a common global FL model with high model accuracy and communication efficiency.

\section{Overview of FedDD Framework}\label{section:Methodology}

In this section, we introduce the FedDD framework, in which different clients are assigned with different dropout rates and adaptively transmit differential sparse models to the server. The key variables and their physical meanings are summarized in TABLE \ref{tab:Summary_MainVariable}.

\begin{table}[h]
	\centering
    \setlength{\abovecaptionskip}{0pt}
     \setlength{\belowcaptionskip}{0pt}
\scriptsize
\caption{Summary of main notations.}\label{tab:Summary_MainVariable}
\begin{tabular}{cc}
  \hline
   Notations & Meanings  \\
  \hline
  $n$    &     The index of a client  \\
  $N$    &     The number of clients\\
  $\bm{W}_{n}^t$   &   The model parameters of client $n$ in global \\
           &    round $t$ before local update\\
  $\hat{\bm{W}}_{n}^t$   &  The model parameters of client $n$ in global\\
                  & round $t$ after local update\\
  $\bm{M}_{n}^t$    &   The mask that determines which parameters of client $n$\\ 
     &        can be uploaded in round $t$ \\
    $t_{n}^{\rm u}$   &   Time for client $n$ to upload model parameters \\
  $t_{n}^{\rm d}$   &   Time for client $n$ to download model parameters \\
  $t_{\rm server}$  &   The processing time of one global round\\
   $D_{n}^{t}$    &   Dropout rate of client $n$ in round $t$ \\
  $D_{\rm max}$    &   The maximal dropout rate of clients\\
  $U$     &  Size of global model \\
  $U_{n}$     &  Size of the local model of client $n$ \\
  $A_{\rm server}$     &   The proportion of parameters amount\\
          &    required by the parameter server      \\
  $\delta$   &    The penalty factor\\
 \hline
\end{tabular}
\end{table}

\begin{figure}[htbp]
\centering
\includegraphics[scale=0.43]{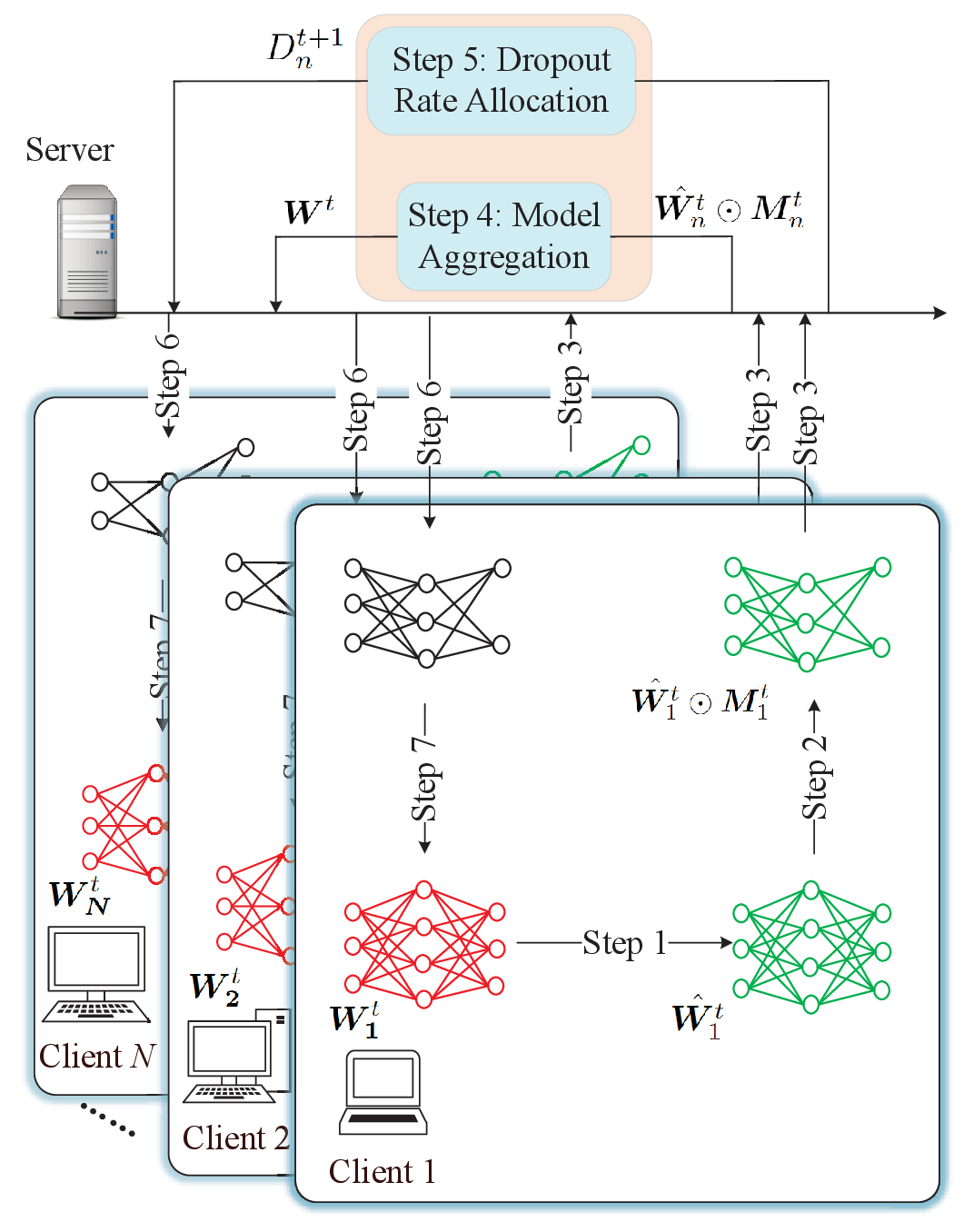}
\caption{Model training process of FedDD.}
\label{fig:overview}
\end{figure}


We now present Fig. \ref{fig:overview} to illustrate the detailed model training process of FedDD. As shown in Fig. \ref{fig:overview}, the training process of FedDD contains seven main steps, which are elaborated as follows:

\textbf{Step 1. Local model training.} Clients train their local models using their local datasets and set their local models as $\hat{\bm{W}}_{n}^{t}= \bm{W}_{n}^{t}-\eta  \nabla  F_{n}(\bm{W}_{n}^{t})$.

\textbf{Step 2. Selection of uploaded model parameters.} After local update, all the clients calculate the important indices of the model parameters and select the parameters with high important indices to form the mask $\bm{M}_{n}^t$ based on the dropout rates $D_n^t$ assigned to them. $D_n^t$ is defined as the proportion of the size of model parameters being dropouted, i.e., not for uploading, and the size of the local full model. That is, $1-D^t_n$ is the ratio of model parameters of the uploaded sparse model by client $n$ for global aggregation.

\textbf{Step 3. Model parameters and training loss upload.} Clients upload $\hat{\bm{W}}_{n}^t \odot \bm{M}_{n}^t$ and the training losses to the server, where $\odot$ is the Hardmard product operator, and $\bm{M}_{n}^t$ is the mask that indicates the sparse subnetwork of client $n$ uploaded to the server. We define $U_n$ as the size of the local model of client $n$ and the number of nonzero elements in $\bm{M}_{n}^t$ is $\left\| \bm{M}_{n}^t \right\|_0 = U_n (1-D_n^t)$, where $\| \cdot \|_0$ denotes $L_0$-norm operator. 

\textbf{Step 4. Global model aggregation.} At this step, the server receives the updated sparse model parameters from the connected clients and then performs model aggregation using the local models received from the connected clients as Eq. (\ref{eq:Wedgek}) shows.

\begin{equation}\label{eq:Wedgek}
\begin{aligned}
\bm{W}^t =\frac{ \sum_{{n}=1}^{N}{m_n\hat{\bm{W}_{n}^t}} \odot \bm{M}_{n}^t }{\sum_{{n}=1}^{N}m_n{\bm{M}_{n}^t}}.
\end{aligned}
\end{equation}
where the division of two matrices represents the element-wise division of the corresponding positions of the two matrices.

The difference between Eq. (\ref{eq:Wedgek}) and Eq. (\ref{eq:flaggregation}) is that every parameter of the global model in Eq. (\ref{eq:Wedgek}) is aggregated from the uploaded sparse models containing this parameter, rather than all the uploaded models.

\textbf{Step 5. Dropout rate allocation.} When global aggregation in one global FL training round is finished, the server recalculates the dropout rates $D_n^{t+1}$ assigned to different clients in the next round.

\textbf{Step 6. Global model and dropout rates download.} The server sends the dropout rates with the corresponding subnetwork to clients. In order to avoid overfitting of some local models, the server broadcasts the full model to all clients every $h$ rounds. Specifically, at round $t$, if $t\, {\rm mod}\, h \neq 0$, the server sends the global sparse models $\bm{W}^{t} \odot \bm{M}_{n}^t$ and the dropout rates $D_n^{t+1}$ to clients. If $t \,{\rm mod}\, h= 0$ (i.e., after every $h$ rounds), the server sends the global full model $\bm{W}^{t}$ and the dropout rates $D_n^{t+1}$ to clients.

\textbf{Step 7. Local model update.} At round $t$, if $t\, {\rm mod}\, h \neq 0$, each client $n$ update its local model using the received global sparse model and the remaining part of its current local model as Eq. (\ref{eq:wnkt1}) shows. If $t\, {\rm mod}\, h= 0$, clients update the local models using the received global full model as Eq. (\ref{eq:wnkt2}) shows. After that, a new global training round begins.

\begin{equation}\label{eq:wnkt1}
\begin{aligned}
\bm{W}_{n}^{t+1}= \bm{W}^{t} \odot \bm{M}_{n}^t + \hat{\bm{W}}_{n}^t \odot (\bm{1}-\bm{M}_{n}^t). \\
\end{aligned}
\end{equation}

\begin{equation}\label{eq:wnkt2}
\begin{aligned}
\bm{W}_{n}^{t+1}= \bm{W}^{t}. \\
\end{aligned}
\end{equation}

The above steps are executed repeatedly until the target number of global rounds is reached. The details of the proposed procedure are summarized in \textbf{Algorithm \ref{alg:total}}.

\begin{algorithm}[htbp]
\caption{FedDD Training Procedure}
\small
\label{alg:total}
		\KwIn{\\
		$\bullet$   Initialized model on the parameter server $\bm{W}^0$.\\
        $\bullet$   Learning rate $\eta$. \\

        $\bullet$   Interval of full model broadcast $h$.}

\textbf{The server executes:}\\
Initializes and broadcasts $\bm{W}^0$ and $D_n^{1}=0$ to all clients $n=1,2,...,N$.\\
\textbf{Clients execute:}\\
$\bm{W}_n^{1} = \bm{W}^0$.\\
\For{$t=1,2,...,T $}
    {\For{{\rm each client} $n=1,2,...,N$ {\rm in parallel}}{
     \textbf{Clients execute:}\\
       Local training: $\hat{\bm{W}_{n}^t}= \bm{W}_{n}^t-\eta  \nabla  F_{n}(\bm{W}_{n}^t)$;\\
       parameter selection: Calculate Eq. (\ref{eq:importance_index_heterogeneous}) and formulate $\bm{M}_{n}^t$;\\
       Upload $\hat{\bm{W}_{n}^t} \odot \bm{M}_{n}^t$ and $loss_{n}$ to the server. \\
     }
     \textbf{The server executes:}\\
     Global aggregation:  $\bm{W}^t =\frac{ \sum_{{n}=1}^{N}{m_n\hat{\bm{W}_{n}^t}} \odot \bm{M}_{n}^t }{\sum_{{n}=1}^{N}m_n{\bm{M}_{n}^t}}$;\\
     Dropout rate calculation: Solve Eq. (\ref{eq:tmin3}) with constraints Eq. (\ref{eq:constraints3}) and get $D_n^{t+1}, n=1,2,...,N$. \\
     \eIf{$t\, {\rm mod}\, h\neq 0$}{\textbf{The server executes:}\\
     Send $\bm{W}^t \odot \bm{M}_{n}^t$ and $D_{n}^{t+1}$ to client $n$ for  $n=1,2,...,N$ in parallel.\\
             \For{{\rm each client} $n = 1,2,...,N$ {\rm in parallel}}{
       \textbf{Clients update model:}\\
        $\bm{W}_{n}^{t+1}= \bm{W}^{t} \odot \bm{M}_{n}^t + \hat{\bm{W}}_{n}^t \odot (\bm{1}-\bm{M}_{n}^t).$
       }}
     {\textbf{The server executes:}\\
     Send $\bm{W}^t$ and $D_{n}^{t+1}$ to client $n$ for  $n=1,2,...,N$ in parallel.\\
             \For{{\rm each client} $n=1,2,...,N$ {\rm in parallel}}{
       \textbf{Clients update model:}\\
        $\bm{W}_{n}^{t+1}= \bm{W}^{t}$.
       }}
     
     }

       \KwOut {$\bm{W}^T$.}
\end{algorithm}

\section{Design Optimization for FedDD}\label{section:Design Optimization for FedDD}
In this section, we will address the key design optimization issues of dropout rate allocation and uploaded parameter selection for FedDD. 

\subsection{Dropout Rate Allocation}\label{subsection:Dropout Rate Allocation}

In this subsection, we discuss how to assign different dropout rates to clients considering the impact of system heterogeneity, data heterogeneity and model heterogeneity.

\textbf{1) System heterogeneity:} Clients' hardware are heterogeneous and have different computing capability. The computation latency for local model training is determined by the computation workload and the computing capability of a device\cite{9598845}. As the number of CPU cycles to process one data sample is related to the model structure, let $c_{n}$ be the CPU cycles for client $n$ to process one data sample, $b_n$ be the batch size of one epoch, and $f_{n}$ be the CPU frequency of client $n$. The computation latency is given by

\begin{equation}\label{eq:tcop}
\begin{aligned}
t_{n}^{\rm cmp} = \frac{c_{n}b_{n}}{f_{n}}.
\end{aligned}
\end{equation}

Let $r_{n}^{\rm u}$ denote the uplink data rate of client $n$ which is defined as

\begin{equation}\label{eq:rnkup}
\begin{aligned}
r_{n}^{\rm u}=B_n^{\rm u} {\rm log_2}\left(1+\frac{p_{n}^{\rm u} h_{n} }{N_0}\right),\\
\end{aligned}
\end{equation}
where $B_n^{\rm u}$ is the uplink bandwidth between client $n$ and the server, $h_{n}$ is the channel gain between client $n$ and the server, $p_{n}^{\rm u}$ is the uplink transmission power of client $n$, and $N_0$ is the background noise.

The communication time for client $n$ to upload model parameters to the parameter server is defined as

\begin{equation}\label{eq:tnkup}
\begin{aligned}
t_{n}^{\rm u} = \frac{U_{n}(1-D_{n}^{t})}{r_{n}^{\rm u}},
\end{aligned}
\end{equation}
where $U_{n}$ is the data size of model parameters of client $n$, $D_{n}^{t}$ is the dropout rate of client $n$ in round $t$.

The downlink data rate and communication time between the server and client $n$ is defined as follows:

\begin{equation}\label{eq:redgekdown}
\begin{aligned}
r_{n}^{\rm d}=B_n^{\rm d} {\rm log_2}\left(1+\frac{ p_{{\rm server}}^{\rm d} h_{n} }{N_0}\right),\\
\end{aligned}
\end{equation}

\begin{equation}\label{eq:tedgekdown}
\begin{aligned}
t_{n}^{\rm d} = \frac{U_{n}(1-D_{n}^{t})}{r_{n}^{\rm d}},
\end{aligned}
\end{equation}
where $B_n^{\rm d}$ is the downlink bandwidth between the client $n$ and server, $p_{{\rm server}}^{\rm d}$ is the downlink transmission power of the server. 

Taking the moment when the parameter server broadcasts the global model as the starting point of one global round, the time required to complete a round of global training can be written as

\begin{equation}\label{}
\begin{aligned}
t_{\rm server} = \max_{n}(t_{n}^{\rm d} + t_{n}^{\rm cmp} + t_{n}^{\rm u}), 
\end{aligned}
\end{equation}
where $t_{\rm server}$ denotes the processing time of one global round, and is determined by the slowest client. To reduce $t_{\rm server}$, clients with long training time and transmission time should be assigned high dropout rates.

\begin{figure*}[htbp]
	\centering  
	\subfigbottomskip=2pt 
	\subfigcapskip=-5pt 
	\subfigure[MNIST]{
		\includegraphics[width=0.3\linewidth]{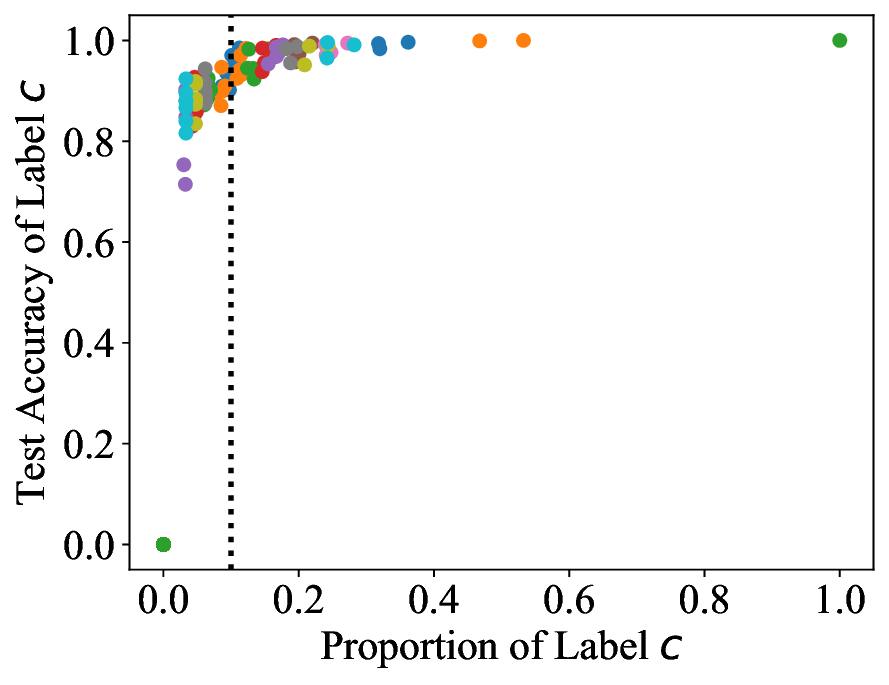}}
  	\subfigure[FMNIST]{
		\includegraphics[width=0.3\linewidth]{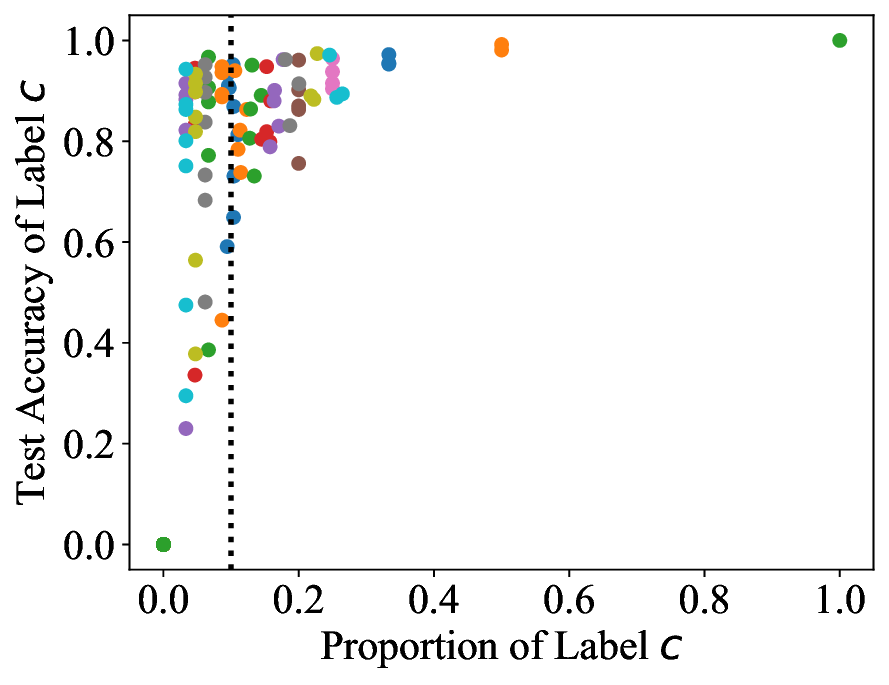}}
	\subfigure[CIFAR10]{
		\includegraphics[width=0.3\linewidth]{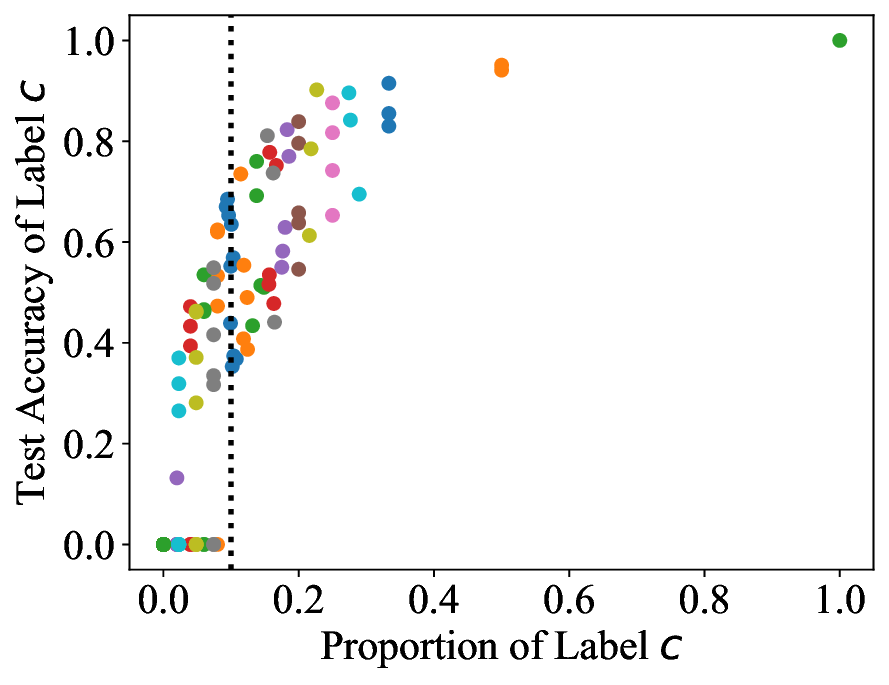}}
	\caption{The performance under different data distributions.}
	\label{fig:distribution_accuracy}
\end{figure*}

\textbf{2) Data heterogeneity:} Data heterogeneity refers to differences in data distribution and data quality among clients. Due to data heterogeneity, each client's contribution to global model convergence is different. Clients with more potential contribution to global model should be assigned to lower dropout rate. We analyze clients' contribution from three perspectives: data amount, data distribution, training loss. 

\textbf{Data amount:} A large amount of training data leads to a good model performance in one global iteration\cite{9842363}. So clients with larger amount of training data should be assigned lower dropout rates. 

\textbf{Data distribution:} Clients with more data labels have stronger generalization of their local models. Besides, the uniformity of data distribution also affects the generalization of local model. For example, both client A and client B have data of three kinds of labels, and the data distributions of their three kinds of labels are \{0.3, 0.3, 0.4\} and \{0.48, 0.48, 0.04\}, respectively. Local model of client B is more biased towards labels 1 and 2, while local model of client A has strong generalization of all three kinds of labels. We conduct some preliminary experiments on MNIST, FMNIST, and CIFAR10 to observe how the data distribution affects the testing performance. As shown in Fig. \ref{fig:distribution_accuracy}, when the proportion of a label gets greater, the corresponding test accuracy get higher. But this change is not linear, and we can observe a key proportion 0.1, i.e., 1/$C$, where $C$ is the total number of classes. The growth rate of test accuracy keeps slowing down and gradually approaches 1 when the proportion is greater than 0.1, while the test accuracy drops sharply even to 0 when the proportion is less than 0.1. Therefore, we design $\sum_{c=1}^{C}\min(Cdis_n^c,1)$ to reflect contribution of the local data distribution, where $dis_n^c$ denotes the proportion of label $c$ in client $n$'s dataset. The physical meaning of $\min(Cdis_n^c,1)$ is that the local model $n$ has good generalization toward class $c$ when $dis_n^c \geq 1/C$, and when $dis_n^c < 1/C$, the generalization of local model $n$ get lower with the decrease of $dis_n^c$. It is worth noting that, a client can calculate $\sum_{c=1}^{C}\min(Cdis_n^c,1)$ and reports to the server as a whole, without revealing the specific data distribution of each label, which is very mild in privacy risk.

\textbf{Training loss:} A larger gradient norm often attributes to a larger training loss \cite{lai2021oort}, which can represent the importance for future training rounds\cite{luo2022tackling}. So clients with larger training losses $loss_n^t$ should be assigned with lower dropout rates.

\begin{figure}[htbp]
\centering
\includegraphics[scale=0.35]{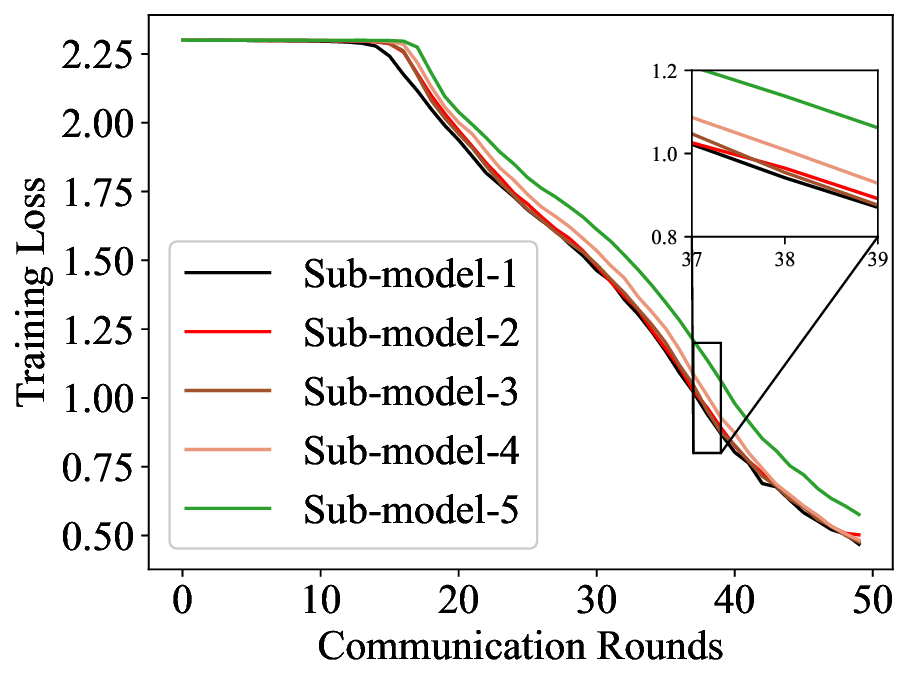}
\caption{The relationship between training loss and model size.}
\label{fig:loss_submodel_B}
\end{figure}

\textbf{3) Model heterogeneity:} Model heterogeneity refers to differences in size and structure of local models among clients. Traditional FL assumes that all the local models have the same structure. However, considering the heterogeneity of clients' devices (storage size, battery capacity, etc.) and the complexity of its own tasks, it is more reasonable to allow clients to train local models of different sizes \cite{diao2021heterofl}. In our scheme, the parameter server holds and processes the global model, whose size is denoted as $U$. Each client can deploy a sub-model (e.g., via model pruning) as the full local model that is suitable for itself before training, whose size is denoted as $U_n$ and $U \geq U_n$ for all $n$. To simplify global aggregation, following the idea of \cite{diao2021heterofl}, local sub-models have similar structure but can shrink their number of channels or neurons within the same layer. When local models of clients are heterogeneous, training losses are varied among heterogeneous models even when trained on the same dataset. Fig. \ref{fig:loss_submodel_B} shows the training losses of 5 heterogeneous models (the model structures are showed in TABLE \ref{tab:modelheterogeneousB} in Appendix \ref{Configurations of Model-Heterogeneous-b Setting} in the separate supplementary file) under IID setting. We can see that training loss of big model is lower than small model in the same round. If we directly compare the training losses of heterogeneous models, this leads to a misleading result: big models with lower losses will be assigned large dropout rates, which is detrimental to the convergence of global model. Therefore, we multiply training loss by a coefficient of normalized model size to rectify the naive training loss. The rectified training loss is denoted as $\frac{U_n}{U}loss_n^t$.

Based on the above analysis, we formulate the following regularized term (Eq. (\ref{eq:re})), which is related to data amount, data distribution, training loss, and size of local model, to reflect the effect of data heterogeneity and model heterogeneity toward the convergence of global model.
\begin{equation}\label{eq:re}
\begin{aligned}
re_n = \frac{m_n}{m}\left(\sum_{c=1}^{C}\min(Cdis_n^c,1)\right)\frac{U_n}{U}loss_{n}^{t}.
\end{aligned}
\end{equation}

Intuitively, clients with larger $re_n$ should be assigned lower dropout rates.

\textbf{4) Optimization of dropout rates:} In our scheme, the parameter server requires the connected clients to upload a certain amount of parameters for model aggregation. Our goal is to minimize the total processing time while ensuring the generalization of the model. So we not only need to reduce the training time and transmission time (affected by system heterogeneity), but also need to consider the contribution of different clients (affected by data heterogeneity and model heterogeneity). To achieve this goal, we formulate the optimization problem as, given the constraints on the amount of parameters, minimizing the processing time of one global round along with the regularization terms, which aims at adjusting the dropout rate dynamically as clients' contributions change in FedDD training process. This optimization problem is expressed as 
\begin{equation}\label{eq:tmin_reclient_reedge}
\begin{aligned}
\min_{D_{n}^{t}}& ( t_{\rm server} + \delta  \sum_{n=1}^{N}re_n D_n^t) =\\
 \min_{D_{n}^{t}}& \Bigg( \max_{n}\left( t_{n}^{\rm cmp} + \frac{U_{n}(1-D_{n}^{t})}{r_{n}^{\rm u}} + \frac{U_{n}(1-D_{n}^{t})}{r_{n}^{\rm d}} \right) \\
& + \delta \sum_{n=1}^{N} \frac{m_n}{m}\left(\sum_{c=1}^{C}\min(Cdis_n^c,1)\right)\frac{U_n}{U}loss_{n}^{t} D_n^t \Bigg),\\
\end{aligned}
\end{equation}

\begin{equation}\label{eq:constraints}
\begin{aligned}
{\rm s.t.}\; & 0\leq D_{n}^{t} \leq D_{\rm max}, \forall  n\in\{1,2,..,N\}\\
             &\sum\limits_{n=1}^{N}{U_{n}(1-D_{n}^{t} )} = A_{\rm server} \sum\limits_{n=1}^{N}U_{n},\\
\end{aligned}
\end{equation}
where $D_{\rm max}$ denotes the maximal dropout rate of clients, $\delta$ is a penalty factor, and $A_{\rm server}$ denotes the proportion of parameters amount required by the parameter server\footnote{The server determines the value of $A_{\rm server}$ according to the available communication resources (e.g., the total bandwidth constraint or transmission cost budget at the server). Due to space constraints, we leave the discussion on how to determine the value of $A_{\rm server}$ in the future work. In this paper we focus on the scheme comparison under the same amount of parameter transmission.}.

\noindent\textbf{Theorem 1.} \emph{The optimization problem Eq. (\ref{eq:tmin_reclient_reedge}) with constraints Eq. (\ref{eq:constraints}) are convex.}\\
\textbf{\emph{Proof.}} The problem of Eq. (\ref{eq:tmin_reclient_reedge}) with constraints Eq. (\ref{eq:constraints}) can be rewritten as Eq. (\ref{eq:tmin3}) with constraints Eq. (\ref{eq:constraints3}).
\begin{equation}\label{eq:tmin3}
\begin{aligned}
\min_{D_{n}^{t}, t_{\rm server} } & \Bigg( t_{\rm server} + \\
 &\delta \sum_{n=1}^{N} \frac{m_n}{m}\left(\sum_{c=1}^{C}\min(Cdis_n^c,1)\right)\frac{U_n}{U}loss_{n}^{t}  D_n^t\Bigg),
\end{aligned}
\end{equation}

\begin{equation}\label{eq:constraints3}
\begin{aligned}
{\rm s.t.}\; & 0\leq D_{n}^{t} \leq D_{\rm max},\;\; \forall  n\in\{1,2,..,N\}\\
             &  A_{\rm server} \sum\limits_{n=1}^{N}U_{n} - \sum\limits_{n=1}^{N}{U_{n}(1-D_{n}^{t} )} = 0 \\
              &      t_{n}^{\rm cmp} + \frac{U_{n}(1-D_{n}^{t})}{r_n^{\rm d}}  +  \frac{U_{n}(1-D_{n}^{t})}{r_n^{\rm u}}  - t_{\rm server} \leq  0. \\
\end{aligned}
\end{equation}
The objective function of Eq. (\ref{eq:tmin3}) is a convex function (a linear function) with respect to optimization variables $t_{\rm server} $ and $ D_n^t$. Moreover, all constraints in Eq. (\ref{eq:constraints3}) are affine. So the optimization problem Eq. (\ref{eq:tmin_reclient_reedge}) with constraints Eq. (\ref{eq:constraints}) is convex.
$\hfill\qedsymbol$

The above optimization problem considers system heterogeneity, data heterogeneity, and model heterogeneity simultaneously, aiming at reducing the total FL processing time, rather than minimizing the process time of a single round. Though it's not easy to calculate closed-form solution of $D_{n}^{t} $ from the above optimization problem, it can be solved by some convex optimization  solvers (e.g., CVXOPT and GUROBI) to obtain optimal solution with low computational complexity. Therefore, only a small computational burden is added to the parameter server.

\subsection{Uploaded parameter selection}\label{subsection:Selection of Uploaded Parameters}

Given a dropout rate, a client needs to determine which part of its model parameters to be uploaded. Neural network models often have hundreds of thousands of parameters, so it is not practical to calculate and compare the relative importance of each model parameter one by one. The number of channels or neurons of each layer are generally tens to thousands, which is much smaller than the number of model parameters. Referring to the structured pruning method\cite{626754c85aee126c0fbcdca4}, we set the same dropout rate for each layer, and perform dropout at channel-wised manner (or neuron-wised manner) within the same layer. Since the relative importance of channels or neurons is only compared within the same layer, the additional computational burden is not large.

Therefore, subject to the dropout rate constraints, the clients should select the most important parameters for uploading. Specifically, in FedDD training, the model parameters of client $n$ are updated as $\hat{\bm{W}_{n}^t}= \bm{W}_{n}^t + \Delta \bm{W}_{n}^t$, where $\Delta \bm{W}_{n}^t = -\eta \nabla  F_{n}(\bm{W}_{n}^t)$. And the parameters uploaded to the server is $\hat{\bm{W}_{n}^t} \odot \bm{M}_{n}^t$. Thus, the error caused by parameter dropout (i.e., $F_{n}((\bm{W}_{n}^t+ \Delta \bm{W}_{n}^t)\odot \bm{M}_{n}^t) - F_{n}(\bm{W}_{n}^t+ \Delta \bm{W}_{n}^t) $) should be minimized. We assume the loss function $F_{n}(\bm{W})$ is $L$-smooth and can derive the following upper bound of the induced error:


\begin{equation}\label{eq:Lsmooth2}
\begin{aligned}
      &  F_{n}\left((\bm{W}_{n}^t+ \Delta \bm{W}_{n}^t)\odot \bm{M}_{n}^t\right) - F_{n}(\bm{W}_{n}^t+ \Delta \bm{W}_{n}^t) \\ 
\leq  &   \left<(\bm{W}_{n}^t+ \Delta \bm{W}_{n}^t)\odot (\bm{M}_{n}^t-\bm{1}), \nabla  F_{n}(\bm{W}_{n}^t+ \Delta \bm{W}_{n}^t) \right>\\
&  +     \frac{L}{2}\left\|(\bm{W}_{n}^t+ \Delta \bm{W}_{n}^t)\odot (\bm{M}_{n}^t-\bm{1}) \right\|^2\\
\leq  & \left\| (\bm{W}_{n}^t+ \Delta \bm{W}_{n}^t)\odot (\bm{M}_{n}^t-\bm{1}) \right\|  \left\| \nabla  F_{n}(\bm{W}_{n}^t+ \Delta \bm{W}_{n}^t)  \right\|\\
 & +    \frac{L}{2}\left\|(\bm{W}_{n}^t+ \Delta \bm{W}_{n}^t)\odot (\bm{M}_{n}^t-\bm{1}) \right\|^2,
\end{aligned}
\end{equation}
where $\left< \cdot, \cdot \right>$ is the inner product operator.

In order to reduce the induced error caused by parameter dropout, we should minimize the corresponding upper bound. We use $[\bm{M}_{n}^t]_k$ to represent the $k$th parameter of $\bm{M}_{n}^t$. From the upper bound in (\ref{eq:Lsmooth2}), it is easy to minimize the term $\frac{L}{2}\|(\bm{W}_{n}^t+ \Delta \bm{W}_{n}^t)\odot (\bm{M}_{n}^t-\bm{1})\|^2$ and $\| \nabla  F_{n}(\bm{W}_{n}^t+ \Delta \bm{W}_{n}^t)  \|\| (\bm{W}_{n}^t+ \Delta \bm{W}_{n}^t)\odot (\bm{M}_{n}^t-\bm{1}) \| $ by setting $[\bm{M}_{n}^t]_k = 1$ corresponding to the largest values of $\| \bm{W}_{n}^t+ \Delta \bm{W}_{n}^t\|$.

On the other hand, in order to speed up the convergence of model training, clients want to upload the parameters with the largest change compared with the parameters before local update. With this purpose, $\|\bm{W}_{n}^t - (\bm{W}_{n}^t+ \Delta \bm{W}_{n}^t)\odot \bm{M}_{n}^t\|^2$ should be maximized. $\|\bm{W}_{n}^t - (\bm{W}_{n}^t+ \Delta \bm{W}_{n}^t)\odot \bm{M}_{n}^t\|^2$ can be rewritten as:

\begin{equation}\label{eq:Lsmooth1}
\begin{aligned}
&  \left\|\bm{W}_{n}^t - \left(\bm{W}_{n}^t+ \Delta \bm{W}_{n}^t\right)\odot \bm{M}_{n}^t \right\|^2\\
=  & \left\|\bm{W}_{n}^t\odot\left(\bm{1}-\bm{M}_{n}^t\right) - \Delta \bm{W}_{n_k}^t\odot \bm{M}_{n}^t  \right\|^2\\
=  & \left \|  \bm{W}_{n}^t\odot\left(\bm{1} - \bm{M}_{n}^t\right) \right\|^2  +   \left\|  \Delta \bm{W}_{n}^t\odot \bm{M}_{n}^t  \right\|^2,
\end{aligned}
\end{equation}
where the second equation of Eq. (\ref{eq:Lsmooth1}) follows from $\left<\bm{1}-\bm{M}_{n}^t, \bm{M}_{n}^t\right> = 0$.

Eq. (\ref{eq:Lsmooth1}) gives us two guidelines for determining mask $\bm{M}_{n}^t$: a) setting $[\bm{M}_{n}^t]_k = 1$ corresponding to the smallest values of $\bm{W}_{n}^t$; b) setting $[\bm{M}_{n}^t]_k = 1$ corresponding to the largest values of $\Delta \bm{W}_{n}^t$.

Based on the analysis from (\ref{eq:Lsmooth2}) and (\ref{eq:Lsmooth1}), next we define the importance index to quantify the importance of parameter $k$ (i.e., channel/neuron $k$) in training process, i.e.,

\begin{equation}\label{eq:importance_index}
\begin{aligned}
\mathcal{I}_n^k=\left\| \Delta \bm{W}_{n}^t  \frac{\bm{W}_{n}^t+\Delta \bm{W}_{n}^t}{\bm{W}_{n}^t}\right\|_{(k)}.
\end{aligned}
\end{equation}
 
Then given a specific dropout rate, the parameters of the largest $\mathcal{I}_n^k$ will be upload.

The above analysis is based on the condition that the local model of clients is homogeneous. When the local models of clients are heterogeneous, some common parameters are owned by all clients and some parameters are owned by only a few clients. In other words, the probability of different parameters being selected for uploading can be different. We define a variable $CR(k)$ to represent the coverage rate of model parameter $k$ among multiple clients (i.e., the proportion of clients possessing such common parameter $k$). In the first global round, all clients upload their full local models to the server, so the server can know the structure and size of each local model, and then calculate and broadcast the $CR(k)$ to all clients. Note that in FL with model averaging mechanism, clients with heterogeneous models will only upload the model parameters which are common among multiple clients. In practice, the heterogeneous client models are usually generated from a big foundation model by using methods such as model pruning tailored to their resource constraints.

Since model parameters with high coverage rates will be shared by more clients, in order to encourage the sharing of the model parameters with low coverage rates to promote the generalization capability of the global model, we rectify Eq. (\ref{eq:importance_index}) to the following form for heterogeneous client model cases:

\begin{equation}\label{eq:importance_index_heterogeneous}
\begin{aligned}
\mathcal{\tilde{I}}_n^{k}=\frac{\left\| \Delta \bm{W}_{n}^t  \frac{\bm{W}_{n}^t+\Delta \bm{W}_{n}^t}{\bm{W}_{n}^t}\right\|_{(k)}}{CR(k)}.
\end{aligned}
\end{equation}

Based on the dropout rate assigned by the server, clients can calculate the number of channels (or neurons) to upload for each layer. After clients perform local update of their local models, they can easily calculate the importance indices of the model parameters according to Eq. (\ref{eq:importance_index_heterogeneous}) and then select the channels (or neurons) with high importance indices to meet the required uploaded number. The procedure of the uploaded parameter selection module is summarized in \textbf{Algorithm \ref{alg:Selection}}.

\begin{algorithm}[htbp]
\caption{Uploaded parameter selection}

\label{alg:Selection}
		\KwIn{\\
		$\bullet$   Dropout rate $D_n^t$.\\
        $\bullet$   The model parameters before local update $\bm{W}_{n}^t$. \\
        $\bullet$   The model parameters after local update $\hat{\bm{W}_{n}^t}$.\\
        $\bullet$   The coverage rate of model parameters $CR=[CR(1), CR(2), ..., CR(K)]$.   /* $K$ is the total channels and neurons of global model.*/}
\textbf{Client $n$ executes:}\\
Calculate $\Delta \bm{W}_{n}^t = \hat{\bm{W}_{n}^t} - \bm{W}_{n}^t$;\\
Calculate $\mathcal{\tilde{I}}_n^{k} =\| \Delta \bm{W}_{n}^t  \frac{\bm{W}_{n}^t+\Delta \bm{W}_{n}^t}{\bm{W}_{n}^t}\|_{(k)}/CR(k)$, for $k = 1,2,...,K$.     \\

\For{{\rm each layer} l}
{ 
Calculate the number of uploaded parameters of layer $l$: $n_l^{\rm up} = N_l * D_n^t$; 
/* $N_l$ is the number of channels or neurons in layer $l$. */  \\
Select the top $n_l^{\rm up}$ parameters with the largest $\mathcal{\tilde{I}}_n^{k}$ value to upload and update $\bm{M}_n^t$.\\

     }

       \KwOut {$\hat{\bm{W}_{n}^t} \odot \bm{M}_n^t $}

\end{algorithm}

\section{Convergence analysis}\label{section:Convergence analysis}

In this section, we theoretically characterize the convergence of FedDD with non-IID data distributions, and derive a convergence bound.

In order to simplify the derivation process, we assume the number of data samples for each client to be the same. We make some commonly used assumptions as follows.

\noindent \textbf{Assumption 1}. $F_1,...,F_{N}$ are $L$-smooth: $\forall \bm{V},\bm{W}, F_{n}(\bm{V}) \leq F_{n}(\bm{W}) + (\bm{V}-\bm{W})^{\rm T}\nabla F_{n}(\bm{W})+\frac{L}{2}\|\bm{V}-\bm{W}\|^2$.\\

\noindent \textbf{Assumption 2}. Variance between local loss function and global loss function is bounded: $\mathbb{E}\| \nabla F_{n}(\bm{W}) - \nabla F(\bm{W})  \|^2 \leq \sigma_n^2$.\\

\noindent \textbf{Assumption 3}. Mask-induced error in global aggregation is bounded: $\mathbb{E}\left\|\frac{ \sum_{{n}=1}^{N}\hat{\bm{W}}_{n}^t \odot \bm{M}_{n}^t }{\sum_{{n}=1}^{N}{\bm{M}_{n}^t}} - \frac{1}{N}\sum_{n=1}^{N}\hat{\bm{W}}_{n}^t\right\|^2 \leq \epsilon \left\|\frac{1}{N}\sum_{n=1}^{N}\hat{\bm{W}}_{n}^t\right\|^2 $, $\forall t$.\\

Under these above assumptions, we have the following theorem, which guarantees the convergence of the proposed FedDD algorithm.

\noindent\textbf{Theorem 2.} \emph{Suppose Assumptions 1-3 hold. Assuming after global aggregation, the server sends full model to the connected clients every $h$ rounds. When $\eta$ is small sufficiently, i.e., $\eta < \frac{2}{L+L\epsilon + 4(\epsilon + 1)\epsilon}$, let $T=Kh$, and we have the following convergence bound:}
\begin{equation}\label{eq:Theoremt0}
\begin{aligned}
          &\frac{1}{T}\sum_{t=0}^{T-1}\mathbb{E} \left \| \nabla F(\bm{W}^{t}) \right\|^2  \leq \\
&  \frac{2\left(F(\bm{W}^{0}) - F(\bm{W}^{*})\right) }{Kh\left(2\eta  -L \eta^2-  L\epsilon \eta^2- 4( \epsilon +1)\epsilon \eta^2\right)}\\
& + \frac{L\epsilon \eta^2 \frac{1}{N}\sum_{n=1}^{N}\sigma_n^2 (h-1) \left( 2\epsilon + 2\epsilon\eta^2 L^2 +  2\eta^2 L^2  +3 \right)}{h\left(2\eta  -L \eta^2-  L\epsilon \eta^2- 4( \epsilon +1)\epsilon \eta^2\right)}\\
& + \frac{ L\epsilon \eta^2 \frac{1}{N}\sum_{n=1}^{N}\sigma_n^2}{h\left(2\eta  -L \eta^2-  L\epsilon \eta^2- 4( \epsilon +1)\epsilon \eta^2\right)}.
\end{aligned}
\end{equation}

The proof is lengthy and omitted here due to space limit. Please refer to Appendix B in the separate supplementary file for detailed proof.  Our convergence analysis takes into account both the sparsification of model upload and model download, while the previous researches only
consider the sparsification of the model upload process. So, the convergence analysis in this paper is applicable to a wider range.

From the theorem above, we see that in the right-hand side of Eq. (\ref{eq:Theoremt0}), the first term tends to zero as $T$ or $K$ approaches infinity. If all the clients upload the full model and the server broadcast the full model every round, FedDD will degenerate into FedAvg, where the second and the third terms in Eq. (\ref{eq:Theoremt0}) become 0. In this case, we show that FedDD converges to the global optimum at a rate of $\mathcal{O}(1/T)$, which is consistent with FedAvg \cite{liconvergence}. 

The second and the third terms of Eq. (\ref{eq:Theoremt0}) together can be viewed as a residual error term, which is dependent on the variance between local loss function and global loss function $\sigma_n$, mask-induced error bound in global aggregation $\epsilon$, and the full model broadcast period $h$. 

Specifically, the parameter $\sigma_n$ can reflect the degree of model heterogeneity among local models and global model. A bigger local model leads to a lower $\sigma_n$. Based on this consideration, in dropout rate allocation module, we multiply training loss by a coefficient of normalized model size to rectify the naive training loss to let clients with big models have more chance to upload more model parameters.

Also, it is easy to check by the first-order condition that the residual error term is a monotone increasing function on $h$. When the server broadcasts full model to clients in every round, i.e., $h=1$, the residual error term reduces to $\frac{L\epsilon \eta^2 \frac{1}{N}\sum_{n=1}^{N}\sigma_n^2}{2\eta  -L \eta^2-  L\epsilon \eta^2- 4( \epsilon +1)\epsilon \eta^2}$. The server can set the value of $h$ based on its available communication resources and operational cost for periodic full model broadcasting. 

On the other hand, the best and worst cases are that all clients upload the full models, and all clients upload nothing, respectively. The corresponding values of $\epsilon$ are 0 and 1, respectively. The residual error term shrink with the decrease of $\epsilon$. When all the clients upload and download full model, i.e., $\epsilon = 0$, the residual error terms reduce to 0. However, such convergence improvement requires more communication cost for the clients' model uploading. Hence, a reasonable $\epsilon$ helps balance the trade-off between communication efficiency and convergence improvement. In FedDD, dropout rate allocation and uploaded parameter selection will affect the value of $\epsilon$. In dropout rate allocation module, a smaller $A_{\rm server}$ usually means a larger $\epsilon$, leading to larger residual error terms, which is confirmed in the sensitivity analysis of Section \ref{Experiments}. Under a fixed $A_{\rm server}$, we set the maximal dropout rate of clients $D_{\rm max}$ to avoid an excessively large $\epsilon$. In uploaded parameter selection module, selecting the model parameters with the largest important indices for uploading also helps to limit the value of $\epsilon$. Thus, our algorithm design can get a better trade-off between the convergence speed and the solution bias.

\section{Performance Evaluation}\label{Experiments}

\subsection{Experimental Setup}

To show the effectiveness of our proposed scheme, we conduct two extensive evaluations: 1) 100 clients simulated in a lab server cluster. 2) a 10-client testbed deployed at 10 geo-distributed virtual machine (VM) instances rented from Alibaba Cloud.

a) Simulation setting:

In the simulation experiment, we consider image classification as the FL task and evaluate the performance of FedDD with three datasets: MNIST\cite{mnistcite}, FMNIST\cite{xiao2017fashion} and CIFAR10\cite{cifar10cite}. To simulate the model heterogeneity in real world, we carry out simulation experiments in two scenarios: 1) \textbf{Model-homogeneous scenarios:} Clients hold and train models with the same structure. 2) \textbf{Model-heterogeneous scenarios:} Clients hold and train models with heterogeneous structures.  In model-homogeneous scenarios, we perform experiments for MLP on MNIST, CNN1 on FMNIST and CNN2 on CIFAR10. The model configurations of this three models are shown in TABLE \ref{tab:model}. In model-heterogeneous scenarios, we perform experiments for five sub-models on CIFAR10 and each sub-model is hold by the same number of clients. To investigate how the degree of model heterogeneity affects the performance of FL, we set up two heterogeneous scenarios, namely model-heterogeneous-a and model-heterogeneous-b. Model-heterogeneous-a represents low model heterogeneity, while model-heterogeneous-b represents high model heterogeneity, as the differences among these five sub-models in model-heterogeneous-b setting are greater. The model configurations of model-heterogeneous-a and model-heterogeneous-b are shown in TABLE \ref{tab:modelheterogeneousA} and TABLE \ref{tab:modelheterogeneousB} (shown in Appendix \ref{Configurations of Model-Heterogeneous-b Setting} in the separate supplementary file), respectively.

\begin{table}[htbp]
\centering
\scriptsize
\setlength{\abovecaptionskip}{0pt}
\setlength{\belowcaptionskip}{0pt}
\caption{Summary of model configurations.}\label{tab:model}
\begin{tabular}{c|c|c}
 \hline
\textbf{MLP} & \textbf{CNN1} & \textbf{CNN2}  \\
\hline
FC(784, 100) & Conv(1, 10, kernel=5) &  Conv(3, 16, kernel=3) \\
ReLU        & Maxpool                    &  ReLU      \\
FC(100, 64)  & ReLU                       &  Maxpool   \\
ReLU     & Conv-(10, 20, kernel=5)&  Conv(16, 32, kernel=3)     \\
FC(64, 10)        & Maxpool                    &  ReLU       \\
Softmax           & ReLU                       &  Maxpool    \\
            & FC(320, 50)                &  Conv(32, 64, kernel=3)  \\
            & ReLU                       &  ReLU     \\
            & FC(50, 10)                 &  Maxpool  \\
            & Softmax                    &  FC(1024, 500)  \\
            &                            &  ReLU  \\
            &                            &  FC(500, 100)  \\
            &                            &  ReLU      \\
            &                            &  FC(100, 10)  \\
            &                            &  Softmax    \\
 \hline
\end{tabular}
\end{table}

\begin{table*}[htbp]
\centering
\scriptsize
\setlength{\abovecaptionskip}{0pt}
\setlength{\belowcaptionskip}{0pt}
\caption{Summary of Model-Heterogeneous-a Configurations.}\label{tab:modelheterogeneousA}
\begin{tabular}{c|c|c|c|c}
\hline
\textbf{Full model (Sub-model-1)} & \textbf{Sub-model-2} & \textbf{Sub-model-3}& \textbf{Sub-model-4}& \textbf{Sub-model-5}\\
\hline
Conv(3, 64, kernel=3) & Conv(3, 64, kernel=3)  &  Conv(3, 64, kernel=3)  &  Conv(3, 32, kernel=3)  &   Conv(3, 32, kernel=3)    \\
ReLU   &  ReLU &   ReLU    &   ReLU  &  ReLU    \\
Maxpool  & Maxpool  & Maxpool  &  Maxpool   &   Maxpool   \\
Conv(64, 128, kernel=3) & Conv(64, 128, kernel=3)  & Conv(64, 128, kernel=3) & Conv(32, 128, kernel=3)    &    Conv(32, 128, kernel=3)    \\
ReLU   &  ReLU &   ReLU    &  ReLU   &  ReLU    \\
Maxpool  & Maxpool  &   Maxpool    & Maxpool    & Maxpool     \\
Conv(128, 256, kernel=3) & Conv(128, 256, kernel=3)  & Conv(128, 256, kernel=3)    &   Conv(128, 256, kernel=3)  &  Conv(128, 128, kernel=3)      \\
ReLU   &  ReLU &   ReLU    &  ReLU   &  ReLU    \\
Maxpool  &  Maxpool &    Maxpool   &  Maxpool   & Maxpool     \\
Conv(256, 512, kernel=3) &  Conv(256, 256, kernel=3) &  Conv(256, 256, kernel=3)     &  Conv(256, 256, kernel=3)   & Conv(128, 256, kernel=3)       \\
ReLU   & ReLU   &   ReLU    &  ReLU   &    ReLU  \\
Maxpool  & Maxpool  &  Maxpool   & Maxpool    &  Maxpool    \\
Conv(512, 512, kernel=3) & Conv(256, 512, kernel=3)  & Conv(256, 512, kernel=3)   &  Conv(256, 512, kernel=3)   &     Conv(256, 512, kernel=3)   \\
ReLU   &  ReLU & ReLU   &  ReLU   &  ReLU    \\
Maxpool  &  Maxpool &  Maxpool     & Maxpool    &    Maxpool  \\
FC(512, 100) & FC(512, 100)  &   FC(512, 80)    &   FC(512, 80)  &  FC(512, 80)      \\
ReLU   &  ReLU & ReLU &  ReLU   &  ReLU    \\
FC(100, 100) & FC(100, 100)  &    FC(80, 100)   &   FC(80, 100)  &  FC(80, 100)      \\
ReLU   & ReLU  &   ReLU    &   ReLU  &  ReLU   \\
FC(100, 10) & FC(100, 10)  &   FC(100, 10)    &  FC(100, 10)   & FC(100, 10)       \\
Softmax   & Softmax   & Softmax      &  Softmax   & Softmax \\
 \hline
\end{tabular}
\end{table*}

The factors that affect system heterogeneity include the computing performance of clients' devices and network connection. Typical parameters of clients' devices, network environment and other experiment setting are provided in Table \ref{tab:Simulation_settings}. All experiments were conducted using PyTorch version 1.10.1.

\begin{table}[htb]
\centering
\scriptsize
\setlength{\abovecaptionskip}{0pt}
\setlength{\belowcaptionskip}{0pt}
\caption{Simulation Settings.}\label{tab:Simulation_settings}
\begin{tabular}{c|c}
\hline
\textbf{Parameter} & \textbf{Value} \\
\hline

$r_n^{\rm u}$     &     [1, 5]   $\times 10^4$ bit/s    \\
$r_n^{\rm d}$   &     [4, 20]   $\times 10^4$ bit/s          \\
  $D_{\rm max}$    &   80\%\\
  $A_{\rm server}$     &   60\%\\
  $f_{n}$ & [1, 10] GHz\\
 $c_{n}$ & [1, 10] Megacycles/sample\\
 Local epoch        &  1 (MNIST), 3 (FMNIST), 5 (CIFAR10)\\
 $h$         &     5\\
 \hline
\end{tabular}
\end{table}

b) Testbed setting:

In the testbed experiment, we rent 11 geo-distributed virtual machine instances, including one parameter server and 10 clients, to constitute a real distributed FL system. These 11 virtual machine instances are located in Guangzhou, Shanghai, Zhangjiakou, Beijing, Nanjing, and Ulanqab, respectively. The specific configuration and geographical location are shown in Table \ref{tab:Testbed_settings}. Due to space limitation, we evaluate the performance of FedDD in testbed for CNN2 on CIFAR10 under Model-homogeneous setting. The hyperparameters $h$, $D_{\rm max}$, and $A_{\rm server}$ are set to 1, 80\% and 60\%.

\begin{table}[htb]
\centering
\scriptsize
\setlength{\abovecaptionskip}{0pt}
\setlength{\belowcaptionskip}{0pt}
\caption{Testbed Settings.}\label{tab:Testbed_settings}
\begin{tabular}{c|c|c}
\hline
\textbf{VM} & \textbf{Configuration} & \textbf{Location} \\
\hline
Parameter server  & 4-vCPU and 1 NVDIA T4 GPU & Ulanqab \\
Client 0  &  8-vCPU and 1 NVDIA P100 GPU   &  Guangzhou  \\
Client 1  &  8-vCPU and 1 NVDIA T4 GPU  &  Nanjing\\
Client 2  &  8-vCPU and 1 NVDIA T4 GPU  &  Nanjing   \\
Client 3  &  4-vCPU and 1 NVDIA T4 GPU  &  Beijing\\
Client 4  &  4-vCPU and 1 NVDIA T4 GPU  &  Beijing \\
Client 5  &  4-vCPU and 1 NVDIA T4 GPU  &  Zhangjiakou\\
Client 6  &  4-vCPU and 1 NVDIA T4 GPU  &  Zhangjiakou\\
Client 7  &  4-vCPU and 1 NVDIA T4 GPU  &  Guangzhou\\
Client 8  &  4-vCPU and 1 NVDIA T4 GPU  &  Guangzhou\\
Client 9  &  8-vCPU and 1 NVDIA P100 GPU   &  Shanghai\\
 \hline
\end{tabular}
\end{table}

To simulate the data heterogeneity in real world, we set up three kinds of data heterogeneity, namely IID, Non-IID-a and Non-IID-b. In IID setting, all classes of data in the datasets are assigned to 100/10 clients with uniform distribution. In Non-IID-a setting, each client is assigned with different classes of data and the number of classes owned by each client is randomly generated from 2 to 10. In Non-IID-b setting, each client is assigned with three randomly selected classes.

\subsection{Evaluation Metrics and Baselines}

\textbf{Performance Metrics.} a) \textbf{Final test accuracy}: We use final testing accuracy to evaluate the generalization performance of the global model. b) \textbf{Time to target accuracy (T2A)}: We use time to target accuracy to evaluate the communication efficiency of FedDD. For clear comparison, we set the time required to attain the target accuracy of FedAvg without dropout as 1, and T2A of other methods is computed by dividing the time of FedAvg without dropout. A smaller T2A represents higher communication efficiency.

\textbf{Baselines.} We compare the proposed scheme with the following 3 schemes. For fairness consideration, the latter 2 schemes and the proposed scheme are compared under the same transmission amount of model parameters. Compared with FedAvg, the proportion of uploaded parameters of other baselines (i.e., communication budget) is set to 60\%, unless otherwise specified.

\begin{itemize}

\item \textbf{FedAvg \cite{mcmahan2017communication}}: In FedAvg, clients can upload and download full model parameters without the communication budget constraints.

\item \textbf{FedCS \cite{nishio2019client}}: In this scheme, the clients with the long communication time will be dropout, in order to satisfy the communication budget.

\item \textbf{Oort\cite{lai2021oort}}: In this scheme, the clients with the lowest self-defined utility will be dropout, subject to the the communication budget constraint. We set straggler penalty $ \alpha $ to 2\cite{lai2021oort}.
\end{itemize}

To verify the rationality of our design choice, we will also conduct comparisons with a multitude of FedDD variant schemes as follows: 
\begin{itemize}
\item \textbf{FedDD w. random selection}: In this scheme, clients select the uploaded parameters randomly.

\item \textbf{FedDD w. max selection}: In this scheme, clients select the uploaded parameters with large amplitude.

\item \textbf{FedDD w. delta selection\cite{aji-heafield-2017-sparse}}: In this scheme, clients select the uploaded parameters with large change.

\item \textbf{FedDD w. ordered selection\cite{horvath2021fjord}}: In this scheme, clients select the uploaded parameters in ordered manner.
\end{itemize}

\subsection{Performance Comparison under Model-homogeneous Setting}

\begin{figure*}[htb]
	\centering  
	\subfigbottomskip=2pt 
	\subfigcapskip=-5pt 
	\subfigure[MNIST]{
		\includegraphics[width=0.29\linewidth]{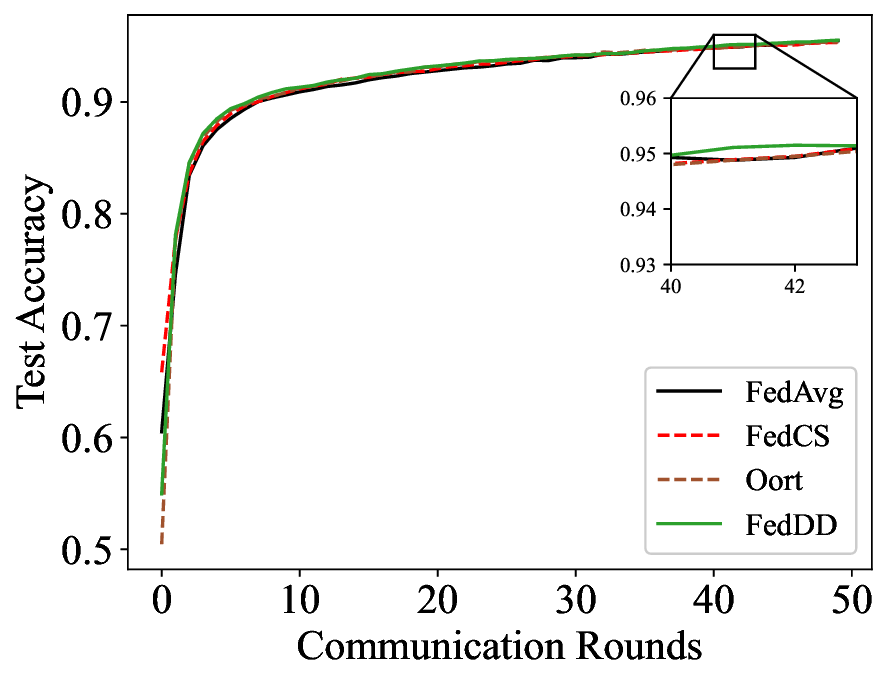}}
	\subfigure[FMNIST]{
		\includegraphics[width=0.29\linewidth]{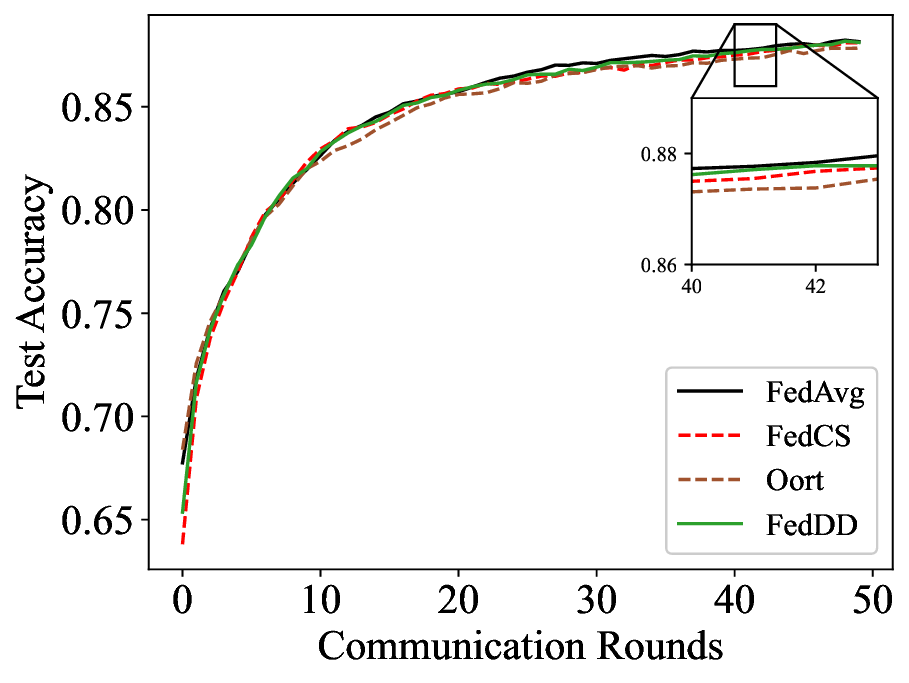}}
	\subfigure[CIFAR10]{
		\includegraphics[width=0.29\linewidth]{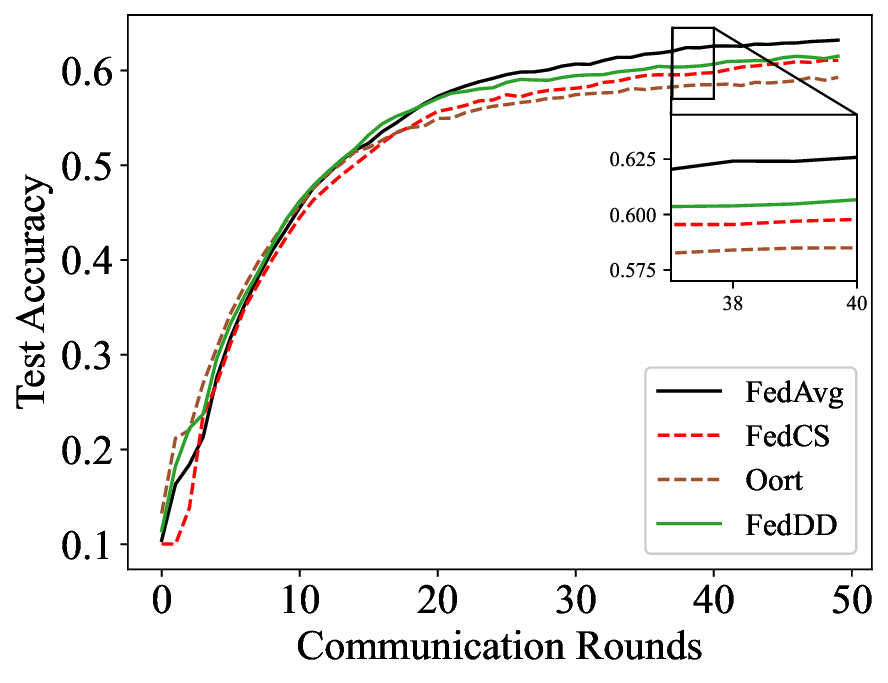}}
	\caption{Curves of top-1 accuracy under IID and model-homogeneous setting in simulation.}
	\label{accuracyIID-homogeneous}
\end{figure*}

We first compare the test accuracy of global model between FedDD and other baseline schemes under model-homogeneous setting in simulation. Fig. \ref{accuracyIID-homogeneous} shows the test accuracy of different schemes under IID and model-homogeneous setting. FedDD and other baseline schemes have almost the same final test accuracy on MNIST and FMNIST dataset, due to the holistic homogeneous settings in FL and the relatively simple classification task of MNIST and FMNIST. On CIFAR10 dataset, FedDD and FedCS have similar final test accuracy, which is 3.6\% higher than Oort.

\begin{figure*}[htb]
	\centering  
	\subfigbottomskip=2pt 
	\subfigcapskip=-5pt 
	\subfigure[MNIST]{
		\includegraphics[width=0.29\linewidth]{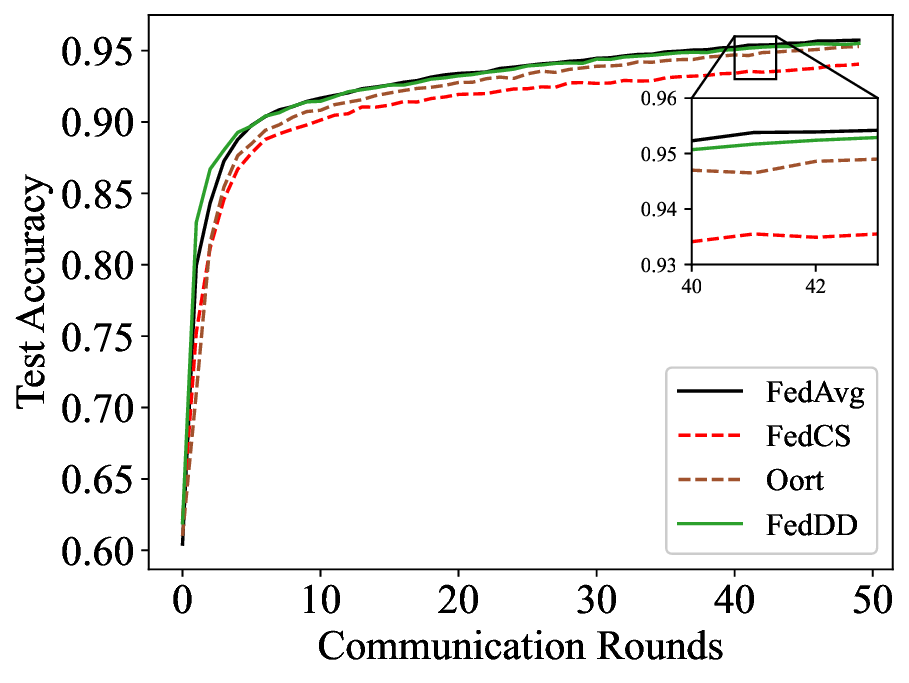}}
	\subfigure[FMNIST]{
		\includegraphics[width=0.29\linewidth]{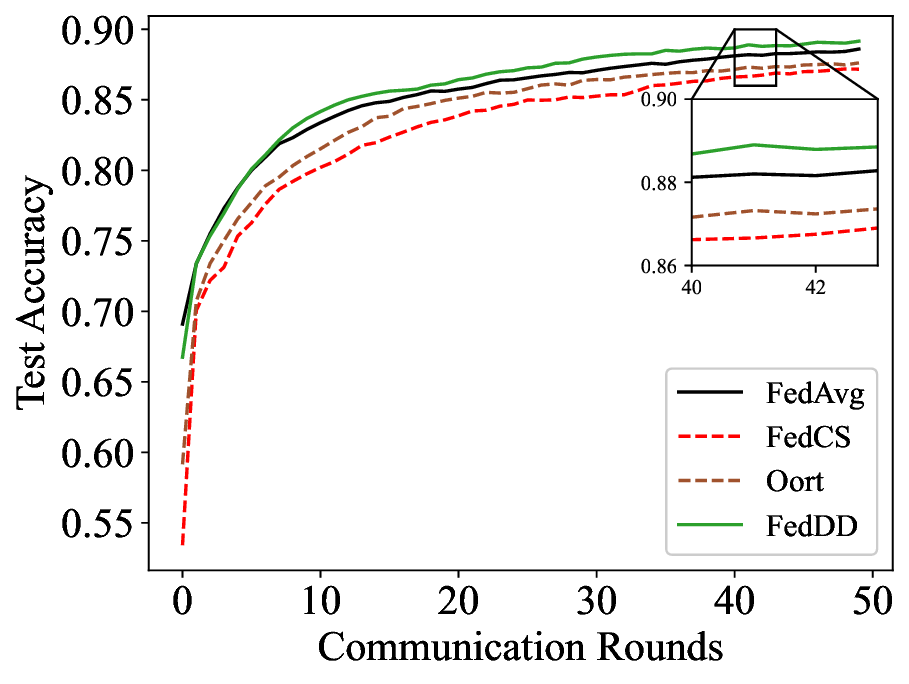}}
	\subfigure[CIFAR10]{
		\includegraphics[width=0.29\linewidth]{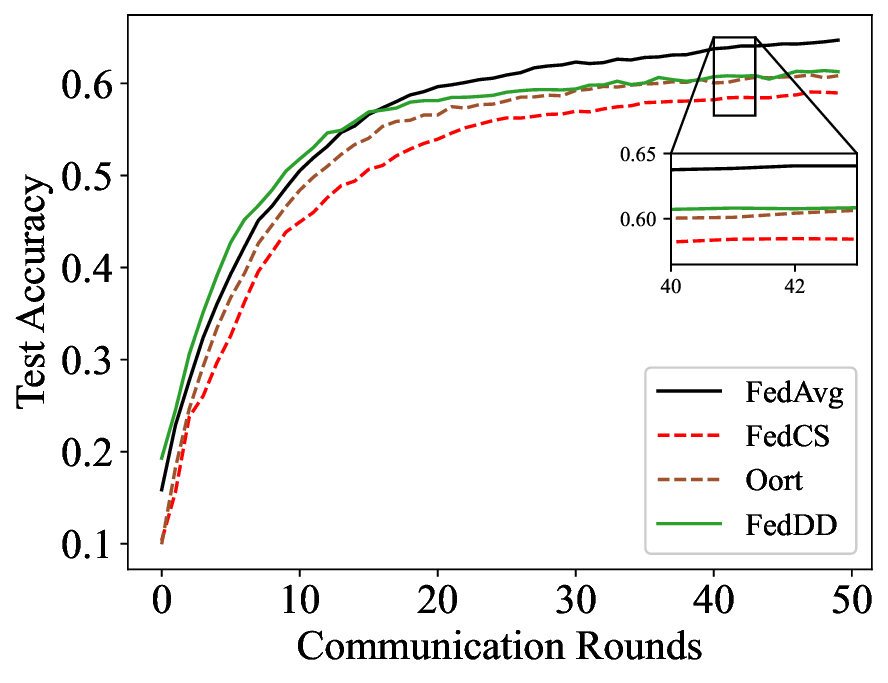}}
	\caption{Curves of top-1 accuracy under Non-IID-a and model-homogeneous setting in simulation.}
	\label{fig:acc_nonIIDa_homogeneous}
\end{figure*}

Fig. \ref{fig:acc_nonIIDa_homogeneous} illustrates the test accuracy of different schemes under Non-IID-a and model-homogeneous setting. On MNIST dataset, the final test accuracy of FedDD is similar to Oort and FedAvg, which is 1.5\% higher than FedCS. On FMNIST dataset, FedDD has highest test accuracy, achieving 1.7\% and 2.2\% better final test accuracy compared with Oort and FedCS. On CIFAR10 dataset, FedDD and Oort have similar final test accuracy, which is 3.8\% higher than FedCS, but FedDD has a faster convergence speed than Oort.

\begin{figure*}[htb]
	\centering  
	\subfigbottomskip=2pt 
	\subfigcapskip=-5pt 
	\subfigure[MNIST]{
		\includegraphics[width=0.29\linewidth]{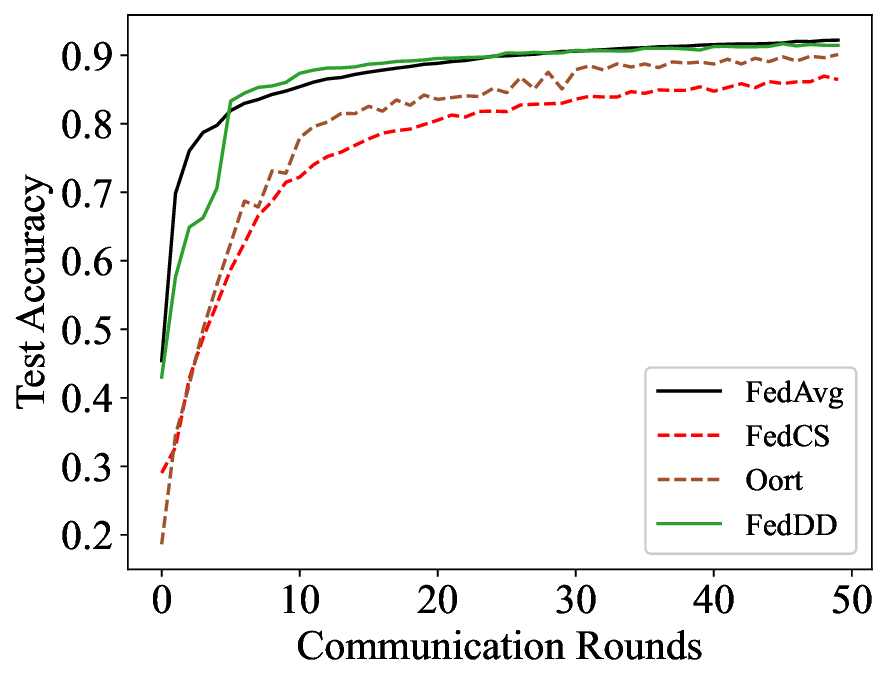}}
	\subfigure[FMNIST]{
		\includegraphics[width=0.29\linewidth]{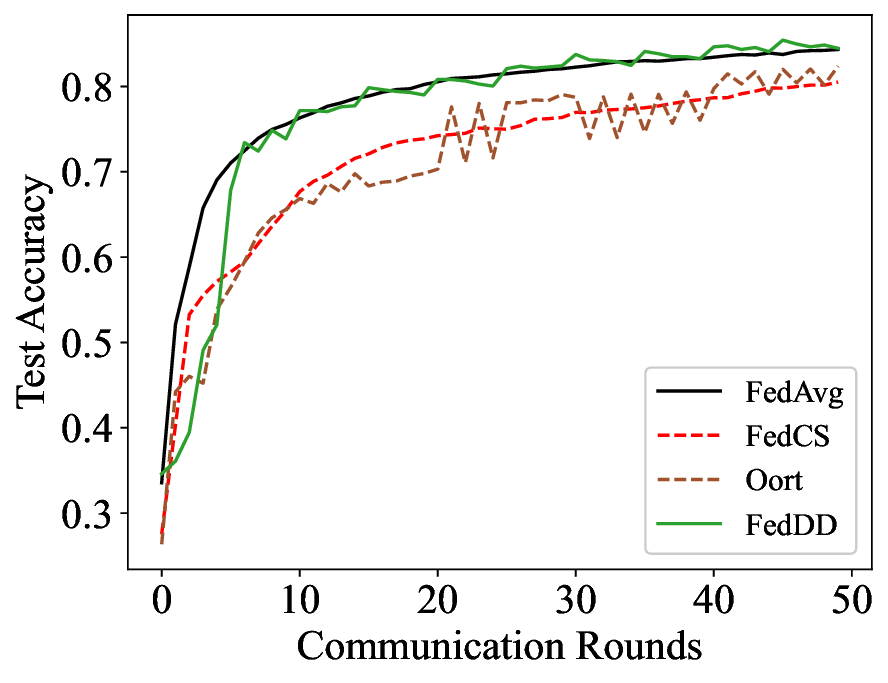}}
	\subfigure[CIFAR10]{
		\includegraphics[width=0.29\linewidth]{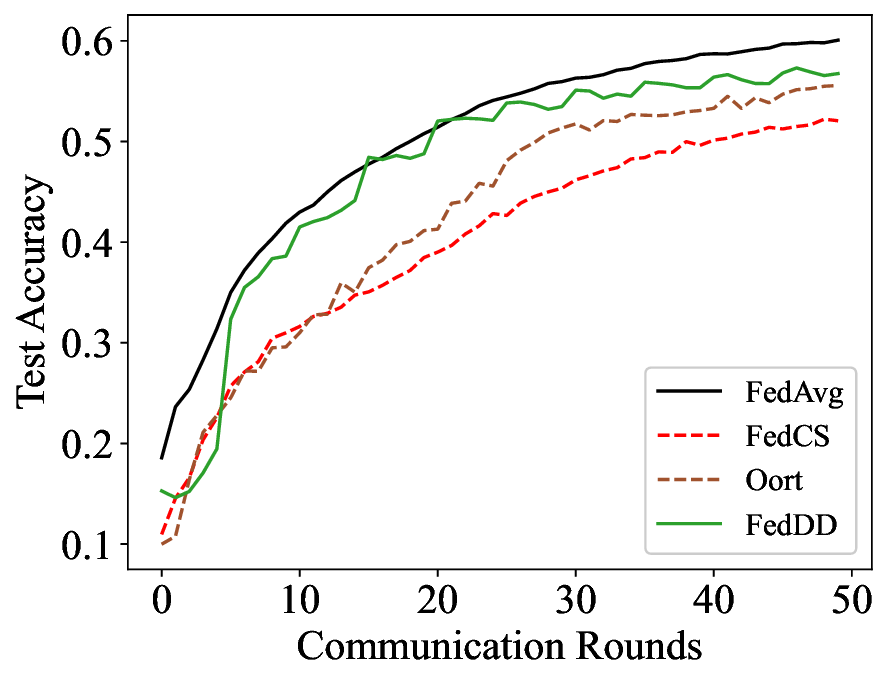}}
	\caption{Curves of top-1 accuracy with Non-IID-b and model-homogeneous setting in simulation.}
	\label{accuracynonIIDb}
\end{figure*}

Fig. \ref{accuracynonIIDb} shows the test accuracy of different schemes in Non-IID-b and model-homogeneous setting. The difference of testing performance is more obvious under this setting with more severe Non-IID degree. The final test accuracy of client-selection-based schemes (FedCS and Oort) are significantly lower than that of our FedDD scheme. On MNIST dataset, FedDD achieves 1.5\% and 5.5\% higher final test accuracy compared with Oort and FedCS. On FMNIST dataset, FedDD achieves 2.5\% and 4.7\% higher final test accuracy compared with Oort and FedCS. On CIFAR10 dataset, FedDD achieves 2.1\% and 8.3\% higher final test accuracy compared with Oort and FedCS. The reason behind this is that client-selection-based method would learn from less data due to client-scale dropout (i.e., the dropout clients' full model parameters will be dropout), which is more likely to cause the global model to be unable to access data of certain labels. The ability of client-selection-based methods to deal with data heterogeneity is not as good as that of FedDD.

\begin{figure}[htbp]
	\centering  
	\subfigbottomskip=2pt 
	\subfigcapskip=-5pt 
	\subfigure[MNIST]{
		\includegraphics[width=0.48\linewidth]{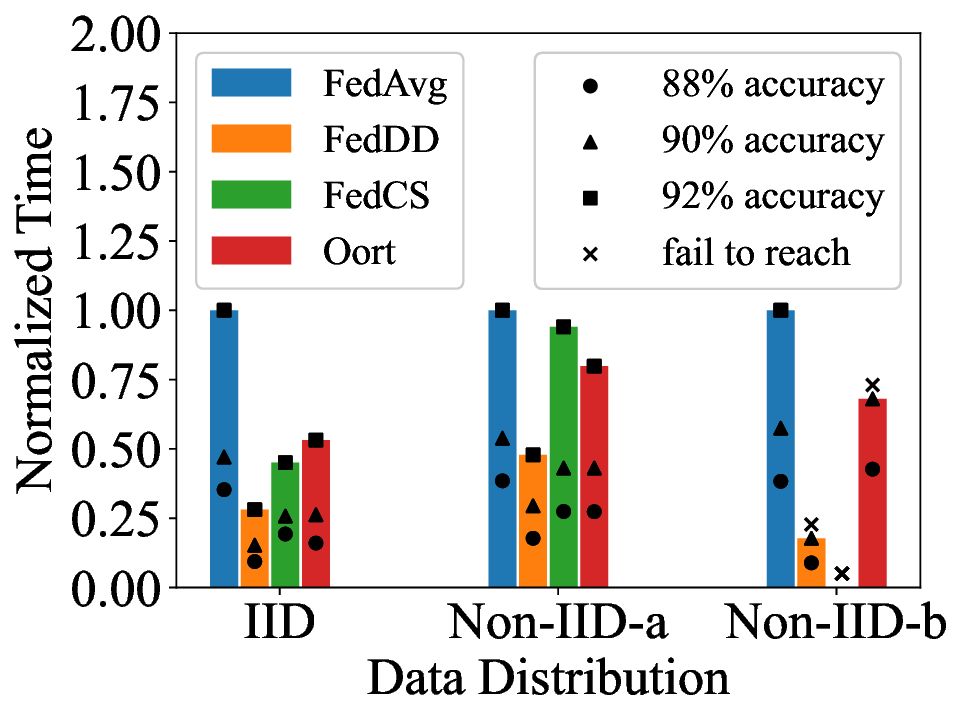}}
	\subfigure[FMNIST]{
		\includegraphics[width=0.48\linewidth]{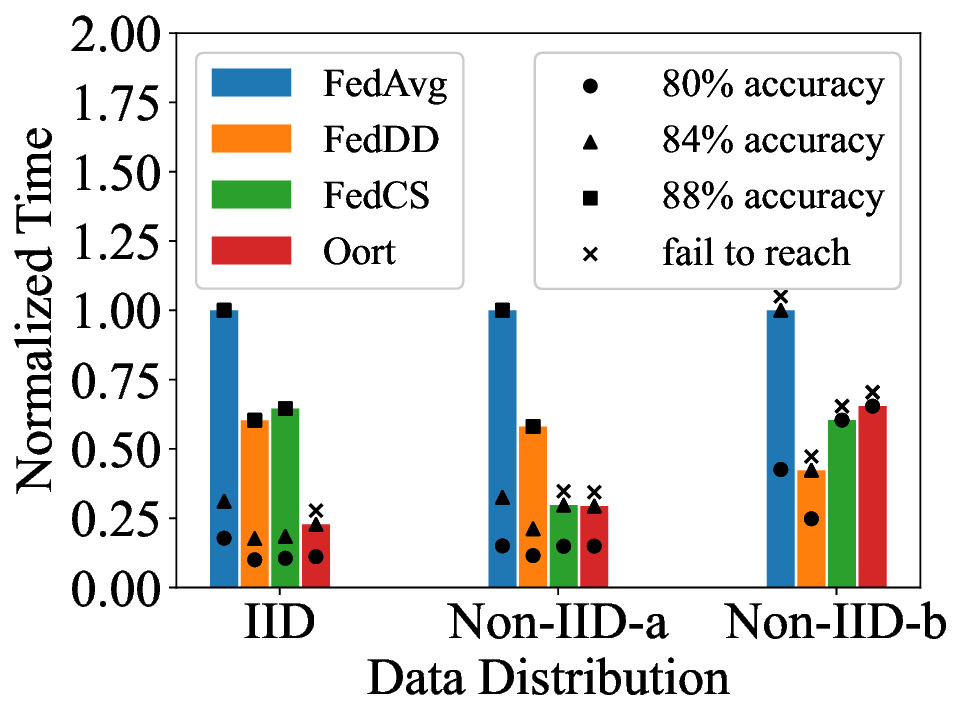}}
	\subfigure[CIFAR10]{
		\includegraphics[width=0.48\linewidth]{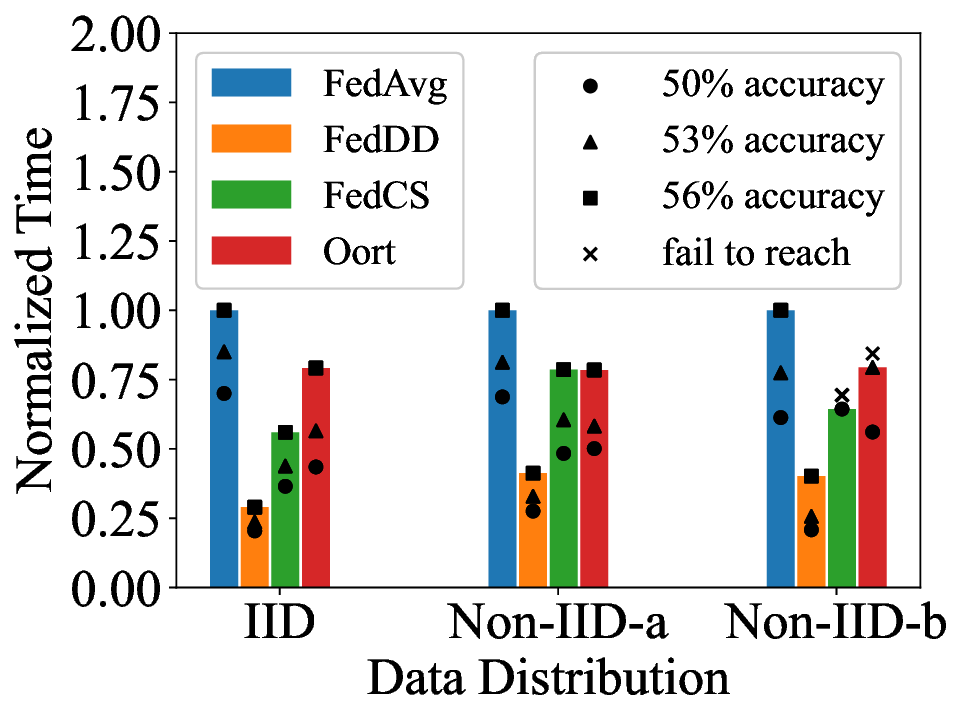}}
	\caption{Time to target accuracies performance of different FL methods under model-homogeneous setting in simulation.}
	\label{fig:time_to_accuracy_homo}
\end{figure}

T2A is an important index to reflect the communication efficiency in FL, we investigate the time to reach several test accuracies for all the FL based schemes. Fig. \ref{fig:time_to_accuracy_homo} records the time to several target test accuracies under model-homogeneous setting. The results show that FedDD can significantly reduces the processing time to reach some target accuracies under all setting, compared with FedAvg and other schemes. For example, under IID distribution, FedDD can reduce the training time by up to 73.5\%  than FedAvg, 51.5\% than FedCS, and 41.5\% than Oort on MNIST dataset to reach the target accuracy of 88\%. Under No-IID-b distribution, FedDD can reduce the training time by up to 59.9\% than FedAvg on CIFAR10 dataset to reach the target accuracy of 56\%, while FedCS and Oort both fail to reach this target accuracy. In all cases, client-selection-based schemes (FedCS and Oort) are inferior to FedDD in T2A performance, and Oort even spends more time than FedAvg in some cases, which indicates that client-selection-based will deteriorate the generalization of global model.

\begin{figure}[htbp]
	\centering  
	\subfigbottomskip=2pt 
	\subfigcapskip=-5pt 
	\subfigure[IID]{
		\includegraphics[width=0.48\linewidth]{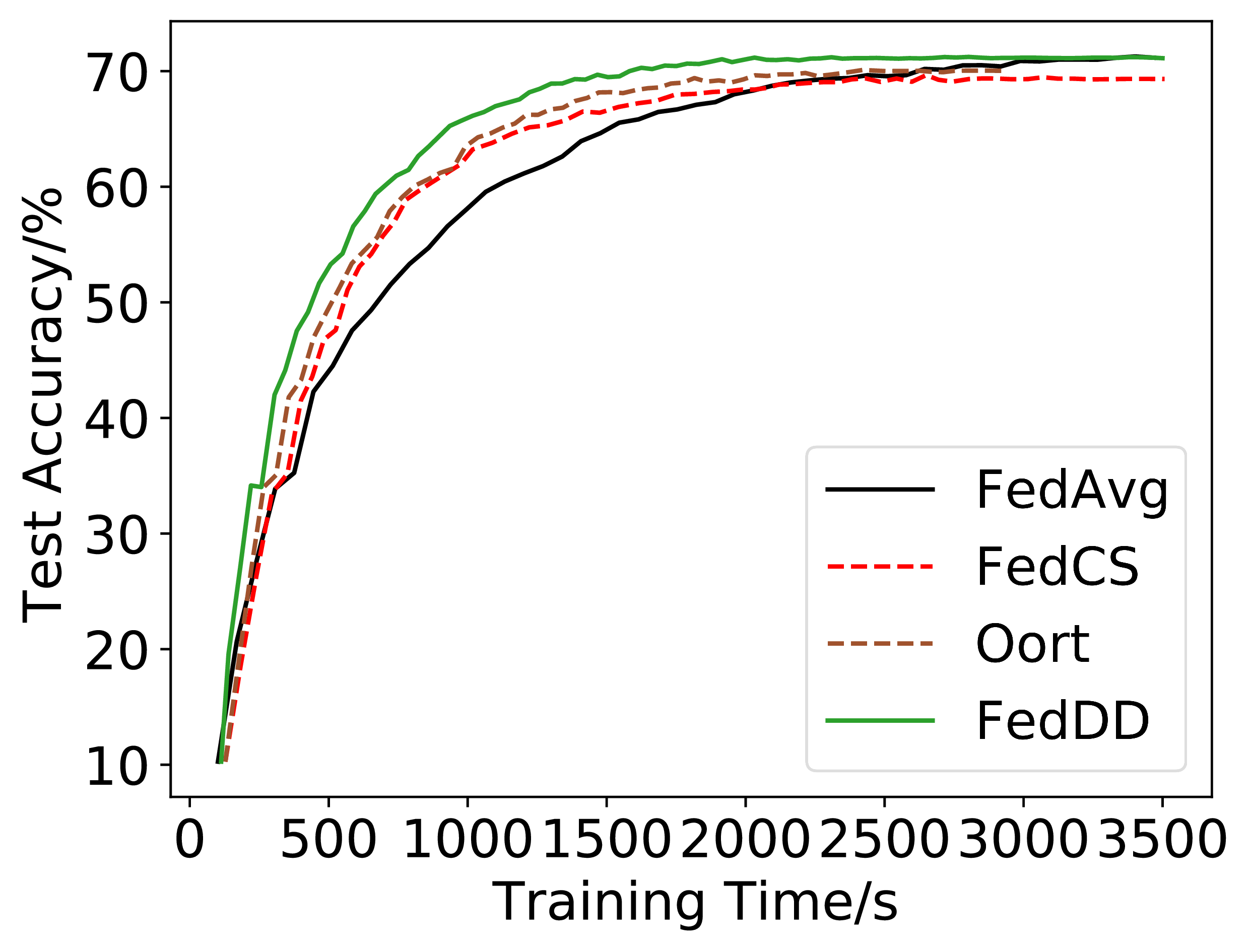}}
	\subfigure[Non-IID-a]{
		\includegraphics[width=0.48\linewidth]{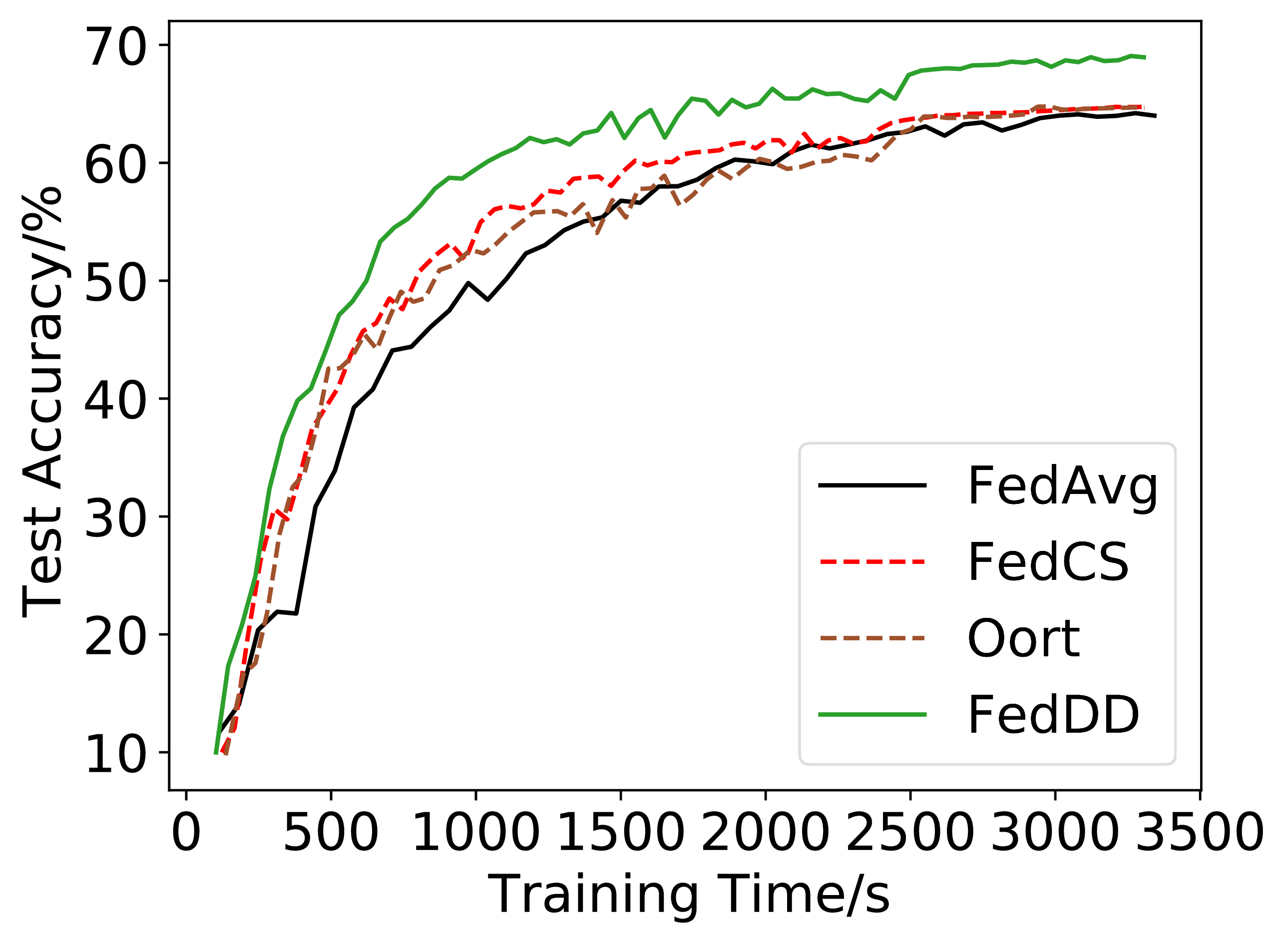}}
	\subfigure[Non-IID-b]{
		\includegraphics[width=0.48\linewidth]{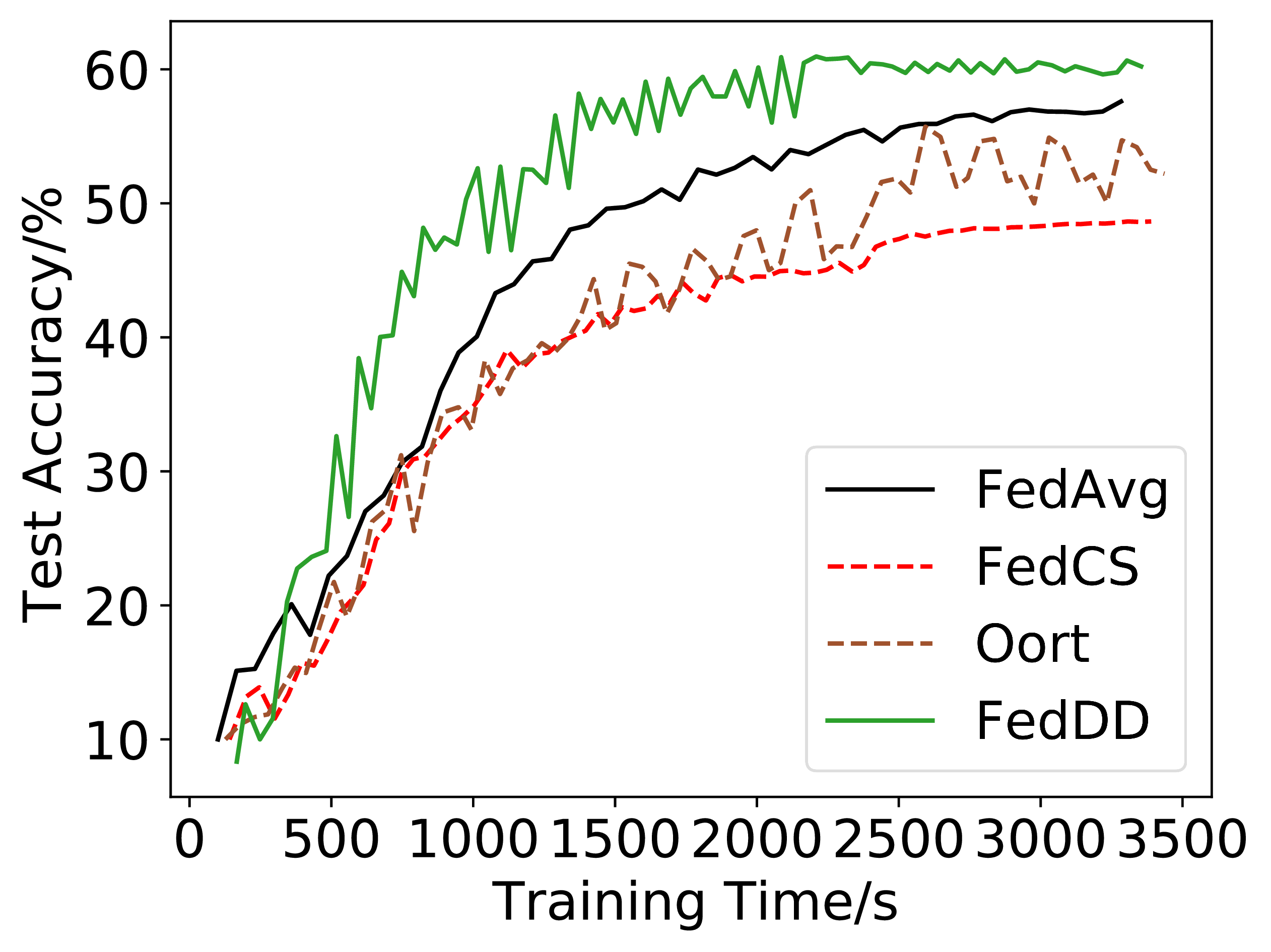}}
	\caption{Curves of top-1 accuracy on CIFAR10 with model-homogeneous setting in testbed.}
	\label{fig:Acc_cifar10_testbed}
\end{figure}

Fig. \ref{fig:Acc_cifar10_testbed} shows the test accuracy of global model between FedDD and other baseline schemes under model-homogeneous setting in testbed. Under three kinds of data heterogeneity, FedDD has the fastest convergence speed. Under IID setting, FedDD reaches the final accuracy 71\% in about 2000 seconds, while FedAvg and Oort takes about 3500 seconds to reach the same final accuracy. FedCS converges faster than FedAvg, but has a lower final accuracy. Under Non-IID-a setting, FedDD reaches the final accuracy 69\% in about 3000 seconds, while other baselines do not converge within 3500 seconds. Under Non-IID-b setting, FedDD reaches the final accuracy 60\% in about 2300 seconds, while other baselines do not converge within 3500 seconds. Moreover, FedCS and Oort have much lower convergence speed than FedDD and FedAvg.

From the experiment results above, we can conclude that when clients' models and server's model are homogeneous, the testing performance of FedDD sightly underperform FedAvg, but can significantly improve the T2A performance (i.e., has much faster converge speed) if we assign suitable dropout rate to different clients.


\subsection{Performance Comparison under Model-heterogeneous Setting}

\begin{figure}[htb]
	\centering  
	\subfigbottomskip=2pt 
	\subfigcapskip=-5pt 
 	\subfigure[IID and model-heterogeneous-a setting]{
		\includegraphics[width=0.48\linewidth]{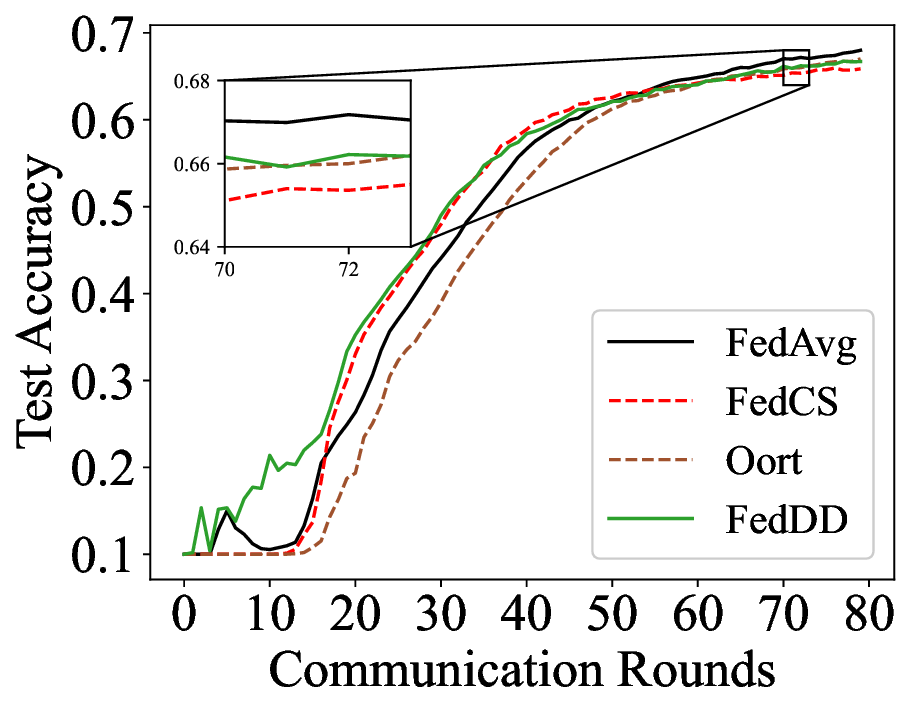}}
    \subfigure[IID and model-heterogeneous-b setting]{
		\includegraphics[width=0.48\linewidth]{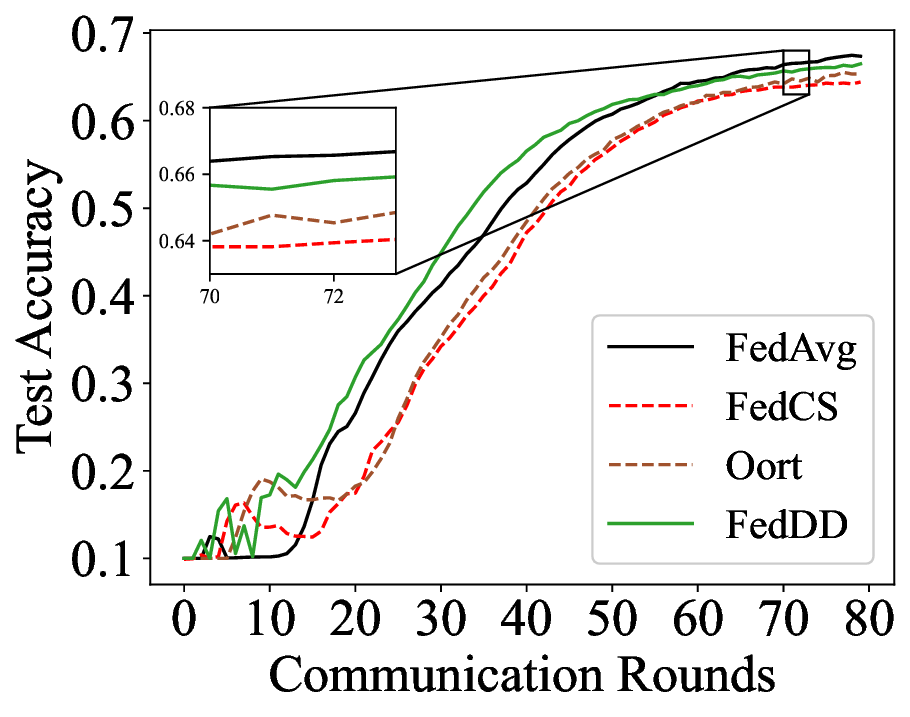}}
	\subfigure[Non-IID-a and model-heterogeneous-a setting]{
		\includegraphics[width=0.48\linewidth]{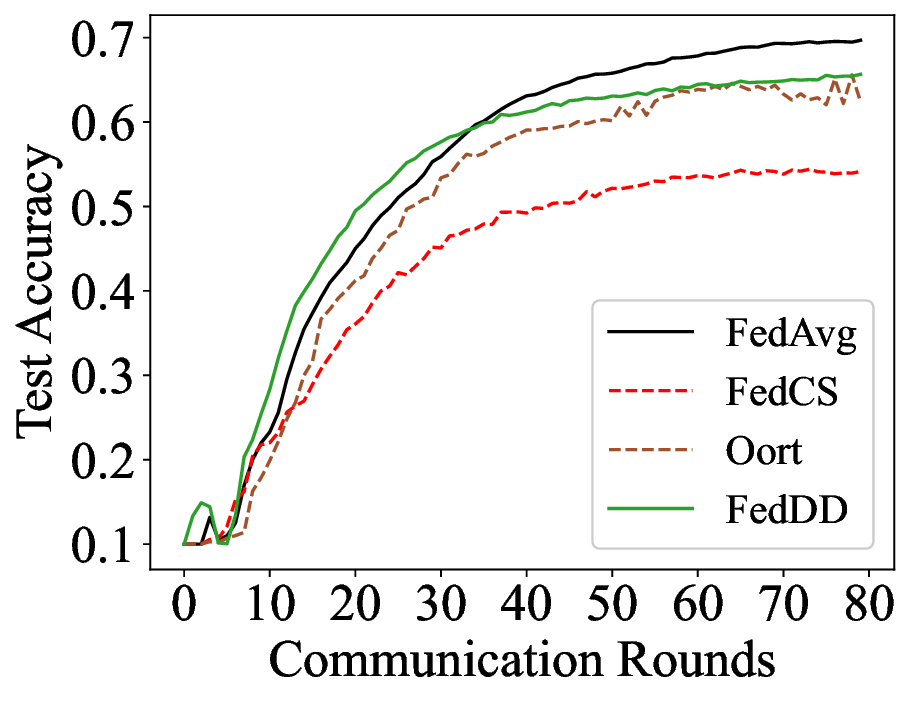}}
  	\subfigure[Non-IID-a and model-heterogeneous-b setting]{
		\includegraphics[width=0.48\linewidth]{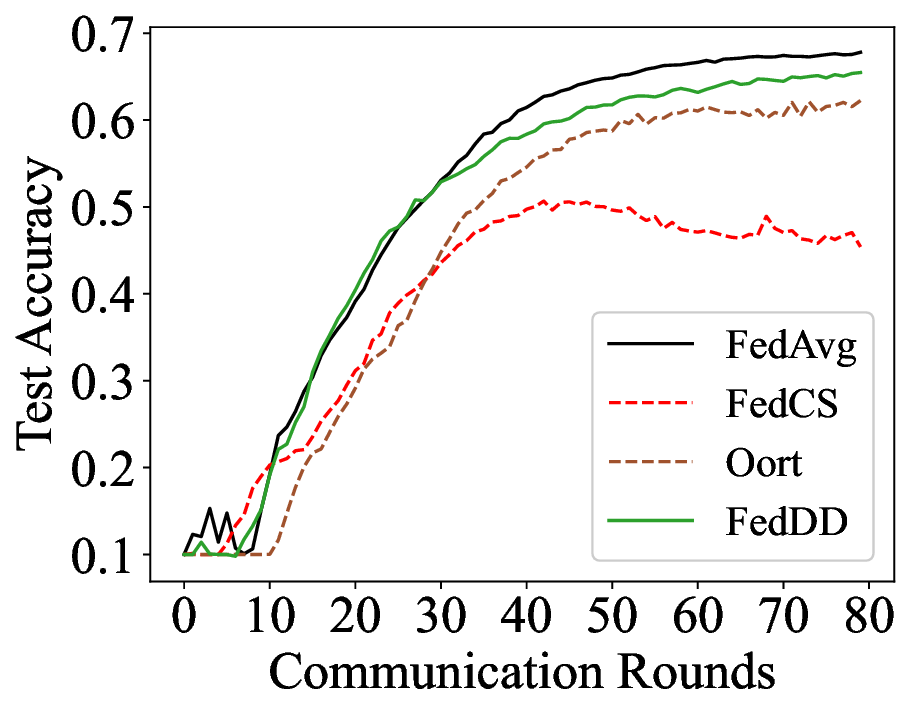}}
	\subfigure[Non-IID-b and model-heterogeneous-a setting]{
		\includegraphics[width=0.48\linewidth]{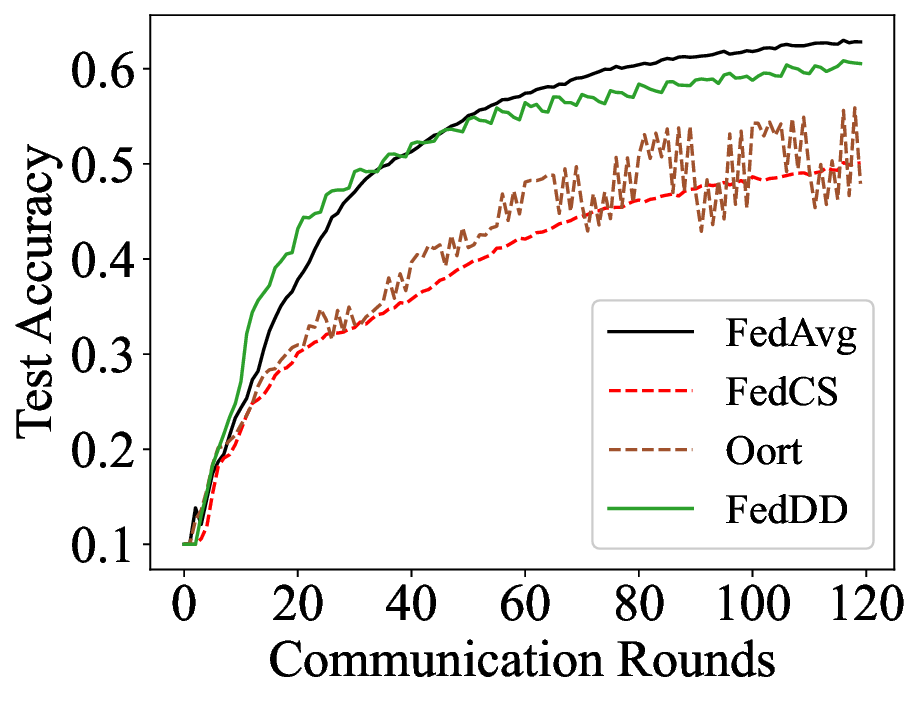}}
	\subfigure[Non-IID-b and model-heterogeneous-b setting]{
		\includegraphics[width=0.48\linewidth]{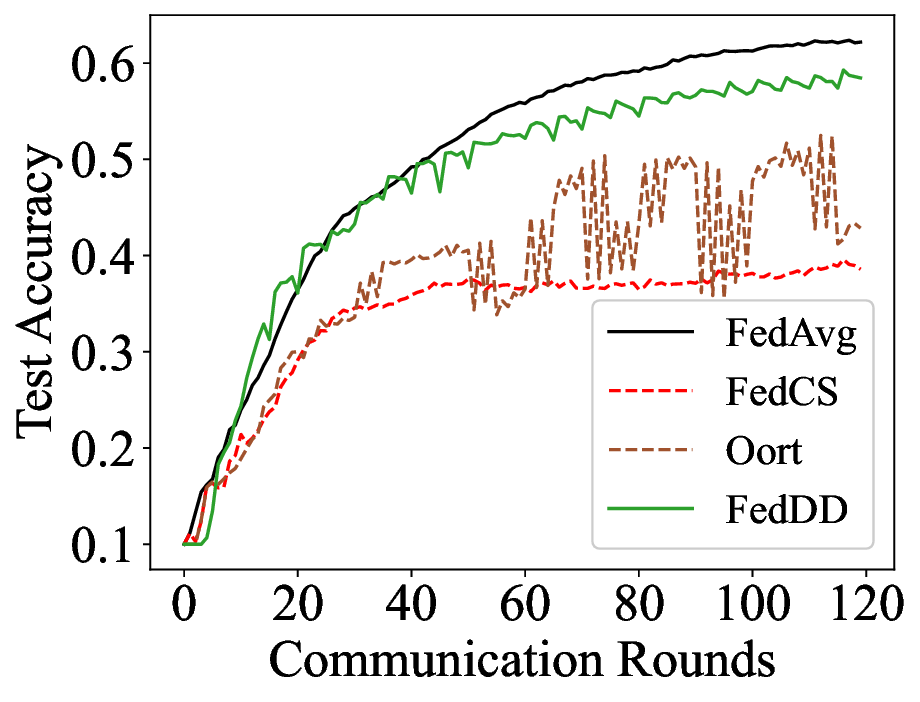}}
	\caption{Curves of top-1 accuracy on CIFAR10 under model-heterogeneous setting in simulation.}
	\label{fig:cifar10_heterogeneous}
\end{figure}

Fig. \ref{fig:cifar10_heterogeneous} illustrates the test accuracy of different schemes on CIFAR10 under model-heterogeneous-a and model-heterogeneous-b setting. We can observe some interesting phenomena different from model-homogeneous setting.

From Fig. \ref{fig:cifar10_heterogeneous}(a), we can see that FedDD achieve similar final test accuracy with other baseline schemes under IID and model-heterogeneous-a setting. As shown in Fig. \ref{fig:cifar10_heterogeneous}(c), FedDD and Oort achieve similar final accuracy, which is 17.5\% higher than FedCS. Fig. \ref{fig:cifar10_heterogeneous}(e) shows that FedDD has the similar convergence speed with FedAvg, and achieves the final test accuracy of 60.6\%, which is 3.6\% lower than FedAvg, 20.9\% higher than Oort, and 17.2\% higher than FedCS. Moreover, Oort has a very unstable performance under model-heterogeneous-b setting.

When the degree of model heterogeneity becomes larger, the advantage of FedDD is more obvious. Compared with Fig. \ref{fig:cifar10_heterogeneous}(a), Fig. \ref{fig:cifar10_heterogeneous}(b) shows that FedCS and Oort underperform FedDD in convergence speed. Fig. \ref{fig:cifar10_heterogeneous}(d) shows that FedDD achieves 4.9\% and 30.6\% higher final test accuracy compared with Oort and FedCS. From Fig. \ref{fig:cifar10_heterogeneous}(f), under model-heterogeneous-b setting, FedCS have the worst final test accuracy, which is 33.6\% lower than FedDD, while the final test accuracy of Oort is 26.2\% lower than FedDD. Compared with model-heterogeneous-a setting, the convergence performance of Oort is more unstable.

\begin{figure}[htb]
	\centering  
	\subfigbottomskip=2pt 
	\subfigcapskip=-5pt 
	\subfigure[Model-heterogeneous-a]{
		\includegraphics[width=0.48\linewidth]{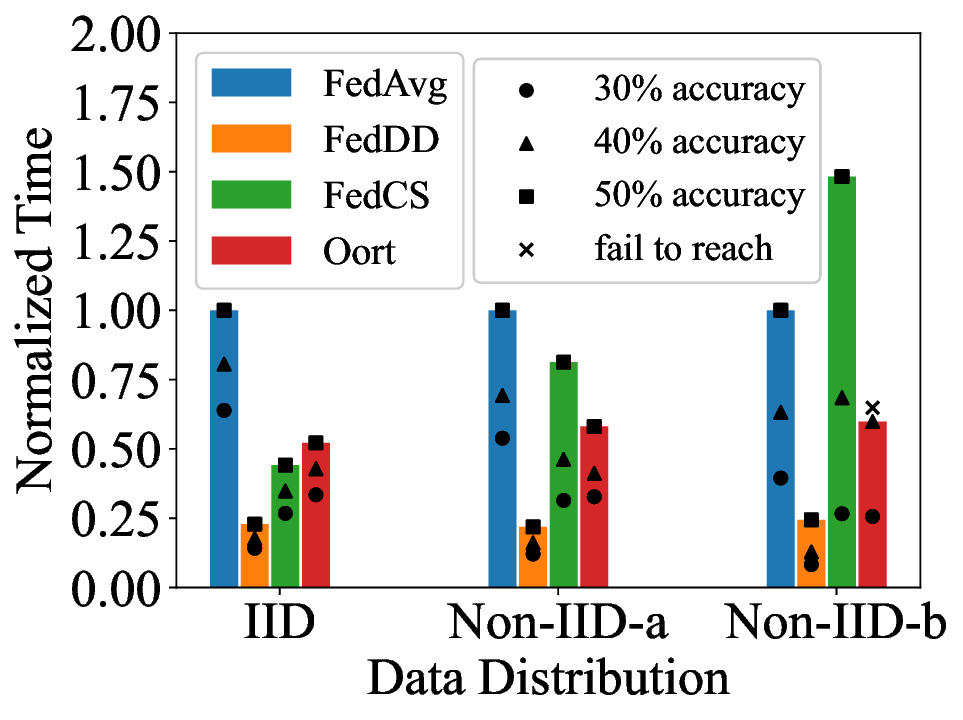}}
	\subfigure[Model-heterogeneous-b]{
		\includegraphics[width=0.48\linewidth]{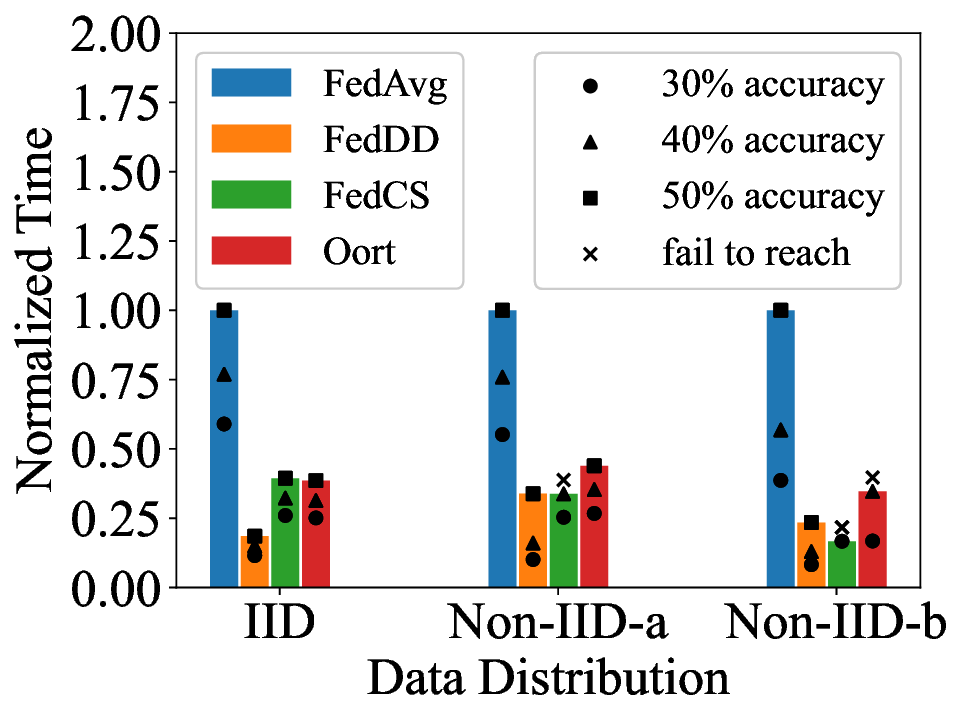}}
	\caption{Time to target accuracies performance of different FL methods under model-heterogeneous setting in simulation.}
	\label{fig:time_to_accuracy_heterogeneous}
\end{figure}

As depicted in Fig. \ref{fig:time_to_accuracy_heterogeneous}, FedDD have the best T2A performance and can significantly reduce the time to several target accuracies compared with FedAvg without dropout. Under model-heterogeneous-a setting, FedDD can reduce the training time by 75.6\%-79.7\% than FedAvg, 47.1\%-83.5\% than FedCS, and 56.2\%-78.6\% than Oort to reach those target accuracies. Under model-heterogeneous-b setting, FedDD can reduce the training time by 66.2\%-81.7\% than FedAvg, 50.6\%-60.2\% than FedCS, and 23.0\%-62.7\% than Oort to reach those target accuracies. It is worth noting that, under Non-IID distribution, client-selection-based schemes do not necessarily improve the performance of T2A. For example, under model-heterogeneous-a and Non-IID-b setting, FedCS spend more time to reach the target accuracies of 40\% and 50\% compared with FedAvg, which Oort fail to reach the target accuracy of 50\%. The reason behind is that client-selection-based schemes can reduce the processing time of a single round, but increase the number of global rounds to reach those target accuacies, due to their ignorance of model heterogeneity and the reduction of accessible data samples.

From the results above, we can see that FedDD can significantly improve the T2A performance with acceptable degradation of final test accuracy, and avoids the damage to the generalization of global model caused by client selection.

\begin{figure}[htb]
	\centering
	\subfigbottomskip=2pt 
	\subfigcapskip=-5pt 
	\subfigure[IID]{
		\includegraphics[width=0.48\linewidth]{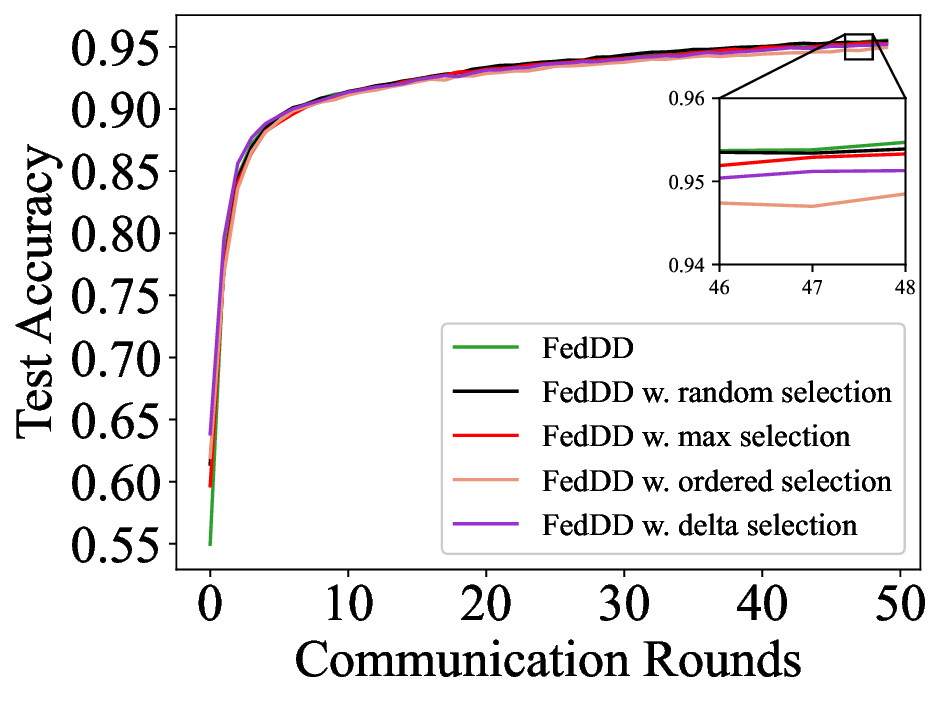}}
	\subfigure[Non-IID-a]{
		\includegraphics[width=0.48\linewidth]{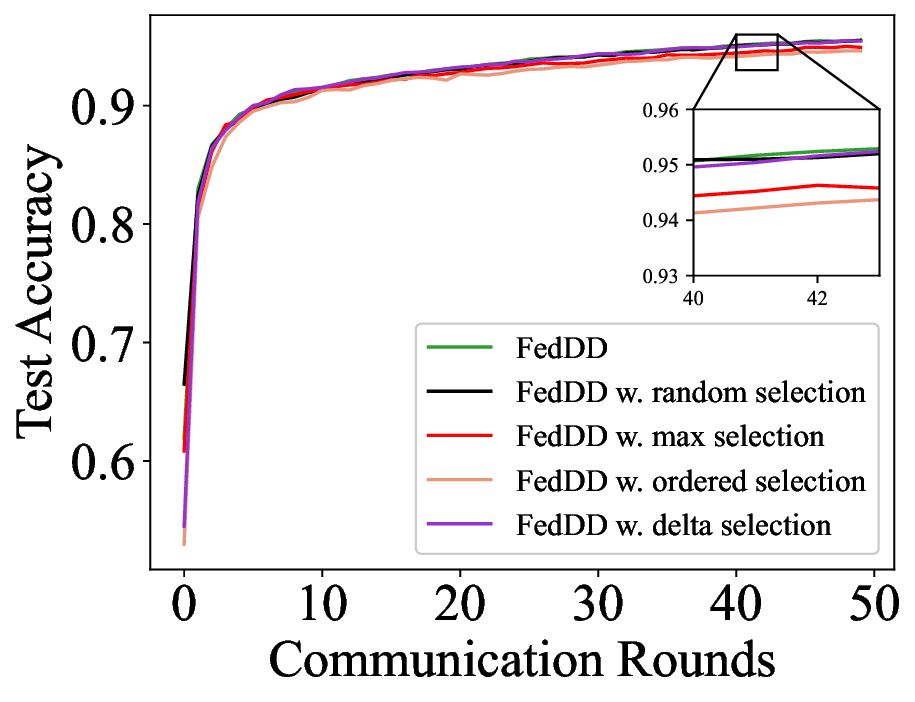}}
	\subfigure[Non-IID-b]{
		\includegraphics[width=0.48\linewidth]{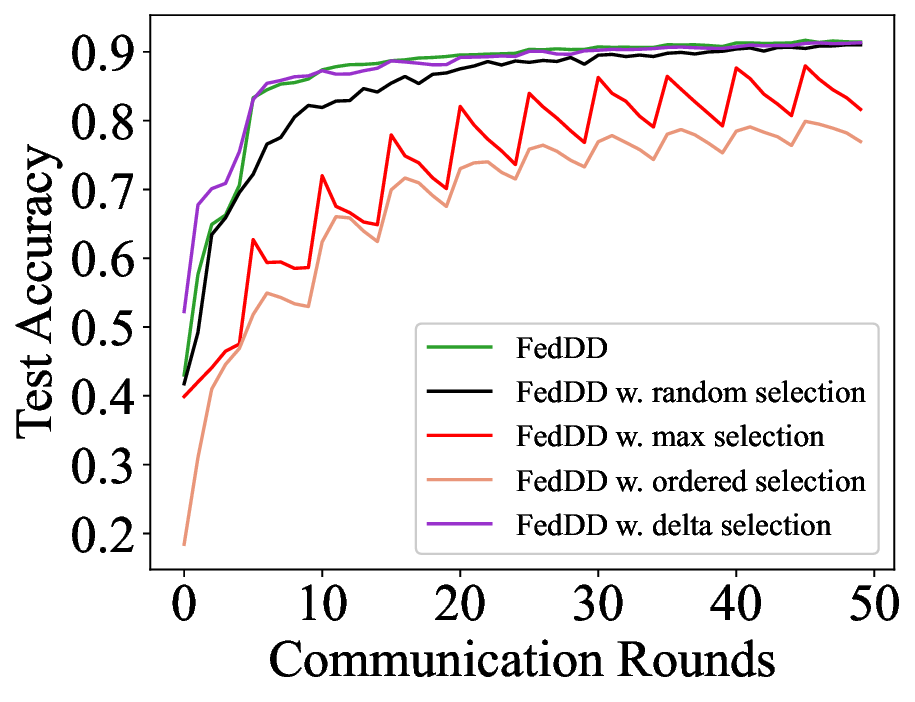}}
	\caption{Curves of top-1 accuracy with different parameter selection schemes on MNIST under model-homogeneous setting in simulation.}
	\label{MNIST_homogeneous_Ablation}
\end{figure}

\begin{figure}[htb]
	\centering
	\subfigbottomskip=2pt 
	\subfigcapskip=-5pt 
	\subfigure[IID]{
		\includegraphics[width=0.48\linewidth]{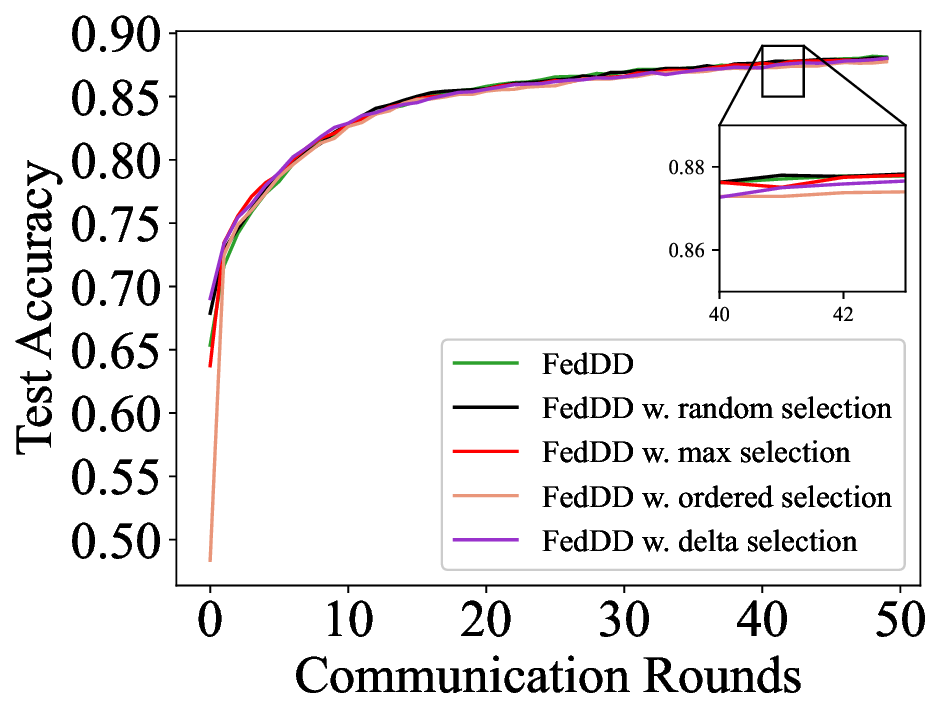}}
	\subfigure[Non-IID-a]{
		\includegraphics[width=0.48\linewidth]{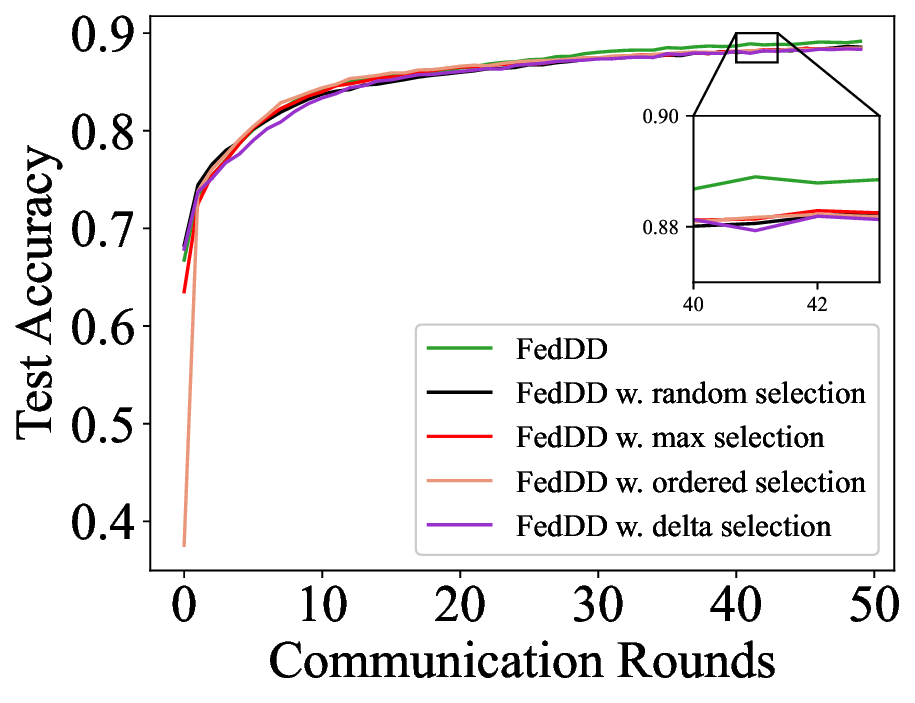}}
	\subfigure[Non-IID-b]{
		\includegraphics[width=0.48\linewidth]{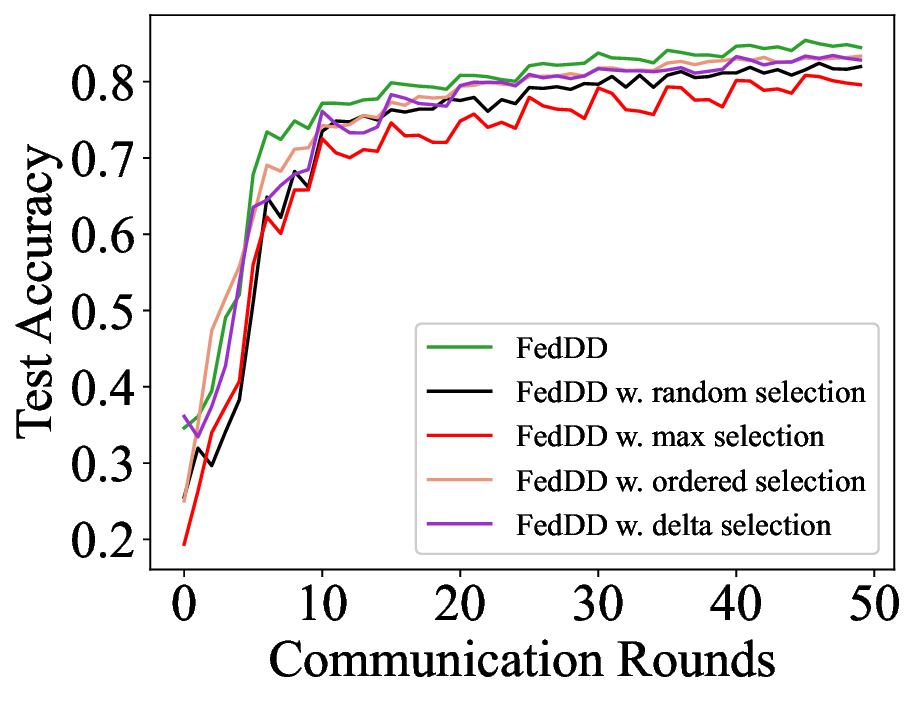}}
	\caption{Curves of top-1 accuracy with different parameter selection schemes on FMNIST under model-homogeneous setting in simulation.}
	\label{FMNIST_homogeneous_Ablation}
\end{figure}

\begin{figure}[htb]
	\centering
	\subfigbottomskip=2pt 
	\subfigcapskip=-5pt 
	\subfigure[IID]{
		\includegraphics[width=0.48\linewidth]{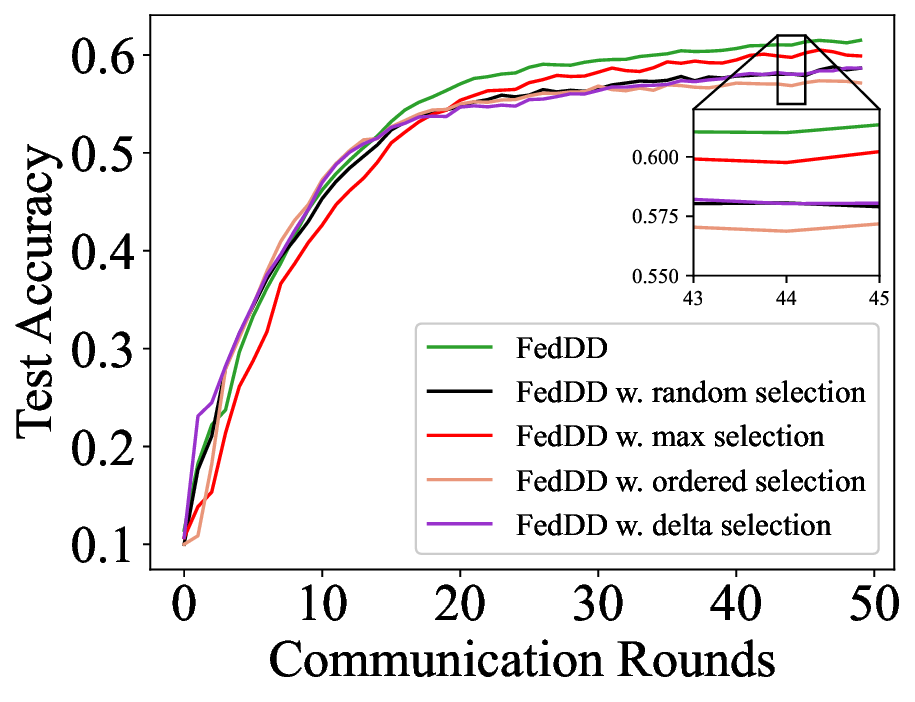}}
	\subfigure[Non-IID-a]{
		\includegraphics[width=0.48\linewidth]{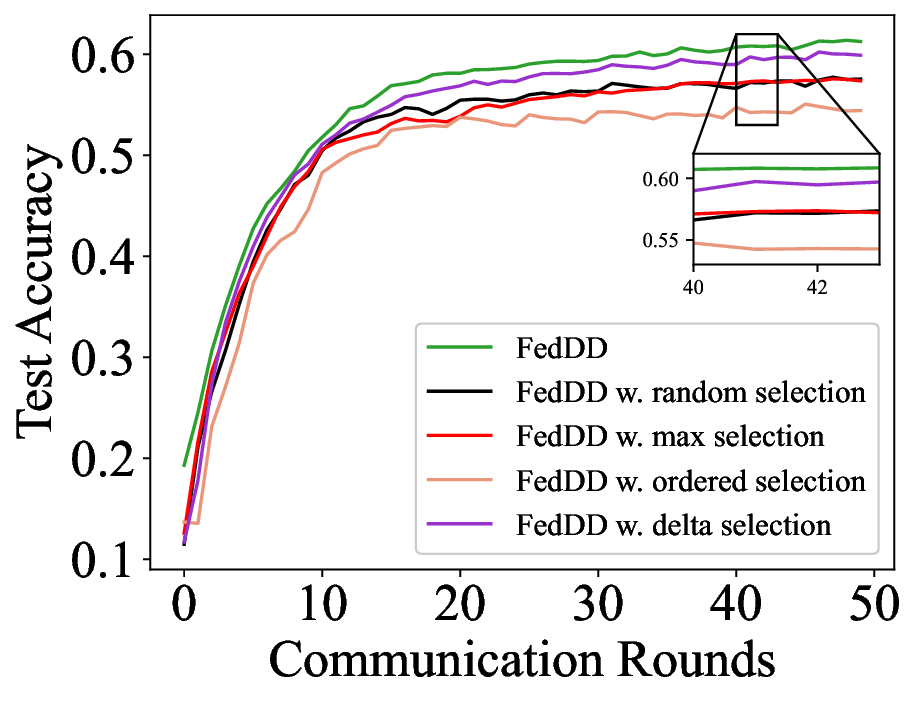}}
	\subfigure[Non-IID-b]{
		\includegraphics[width=0.48\linewidth]{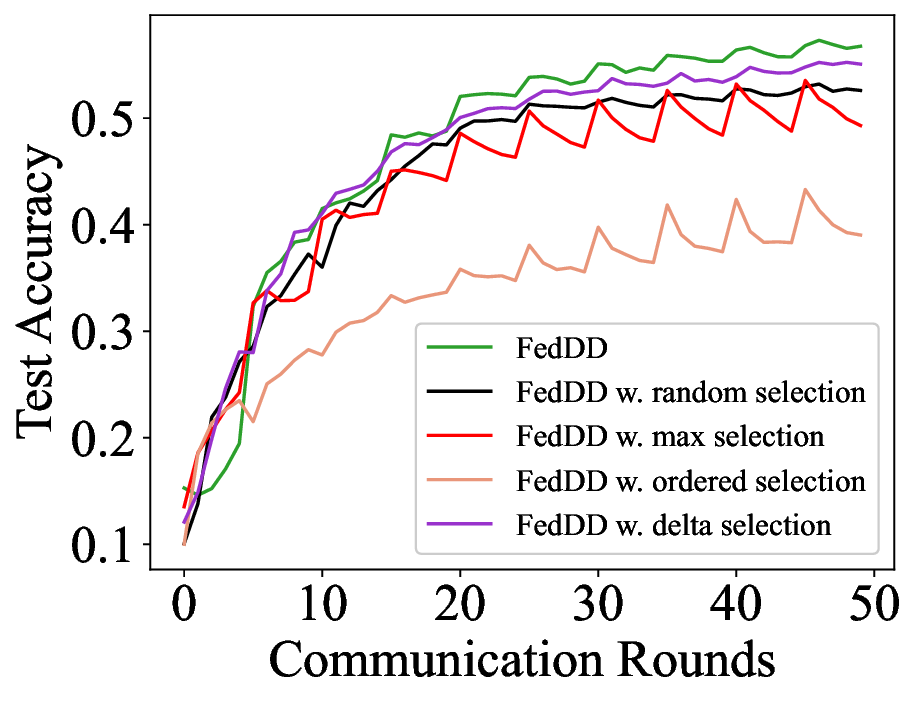}}
	\caption{Curves of top-1 accuracy with different parameter selection schemes on CIFAR10 under model-homogeneous setting in simulation.}
	\label{CIFAR10_homogeneous_Ablation}
\end{figure}

\begin{figure}[htb]
	\centering
	\subfigbottomskip=2pt 
	\subfigcapskip=-5pt 
	\subfigure[IID]{
		\includegraphics[width=0.48\linewidth]{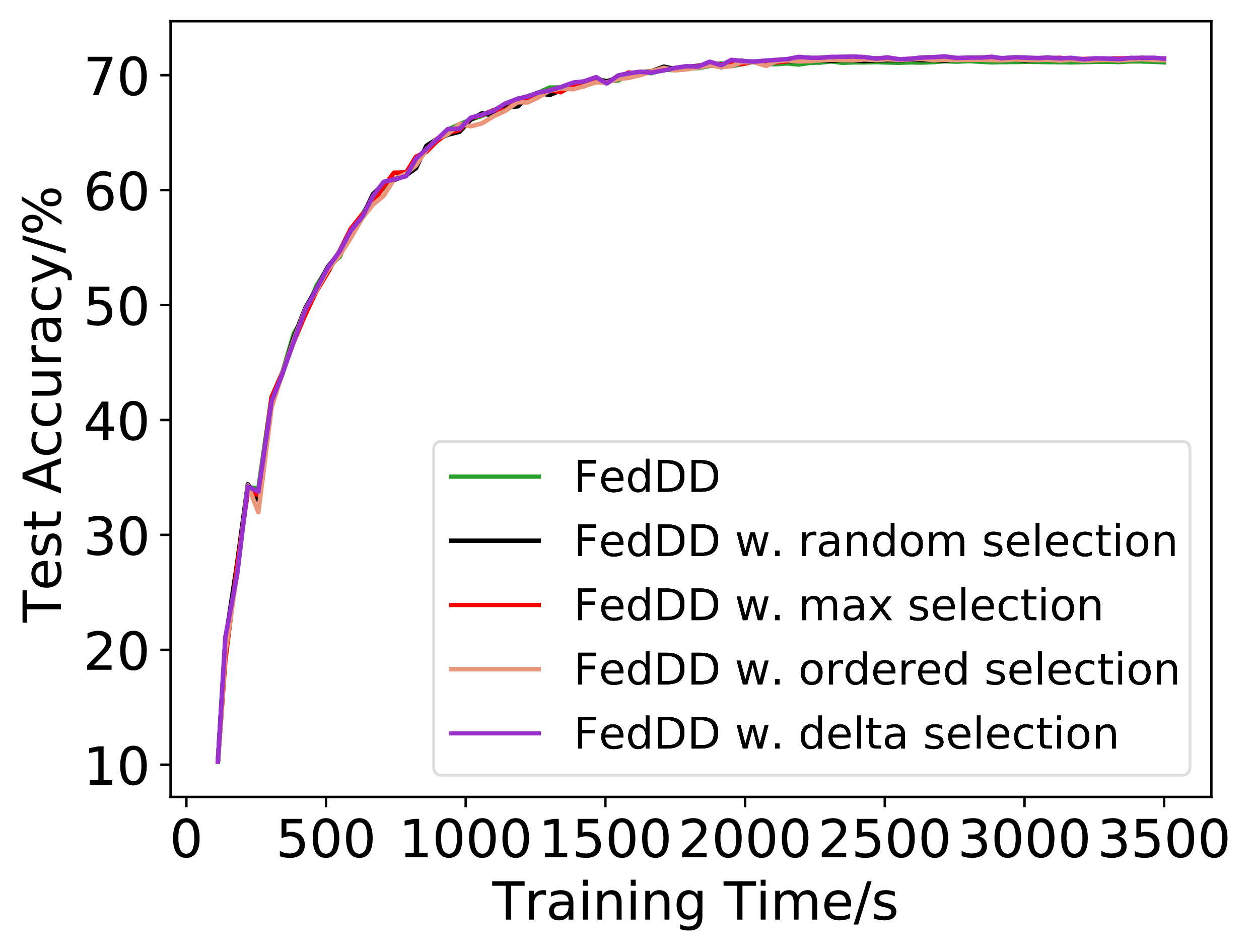}}
	\subfigure[Non-IID-a]{
		\includegraphics[width=0.48\linewidth]{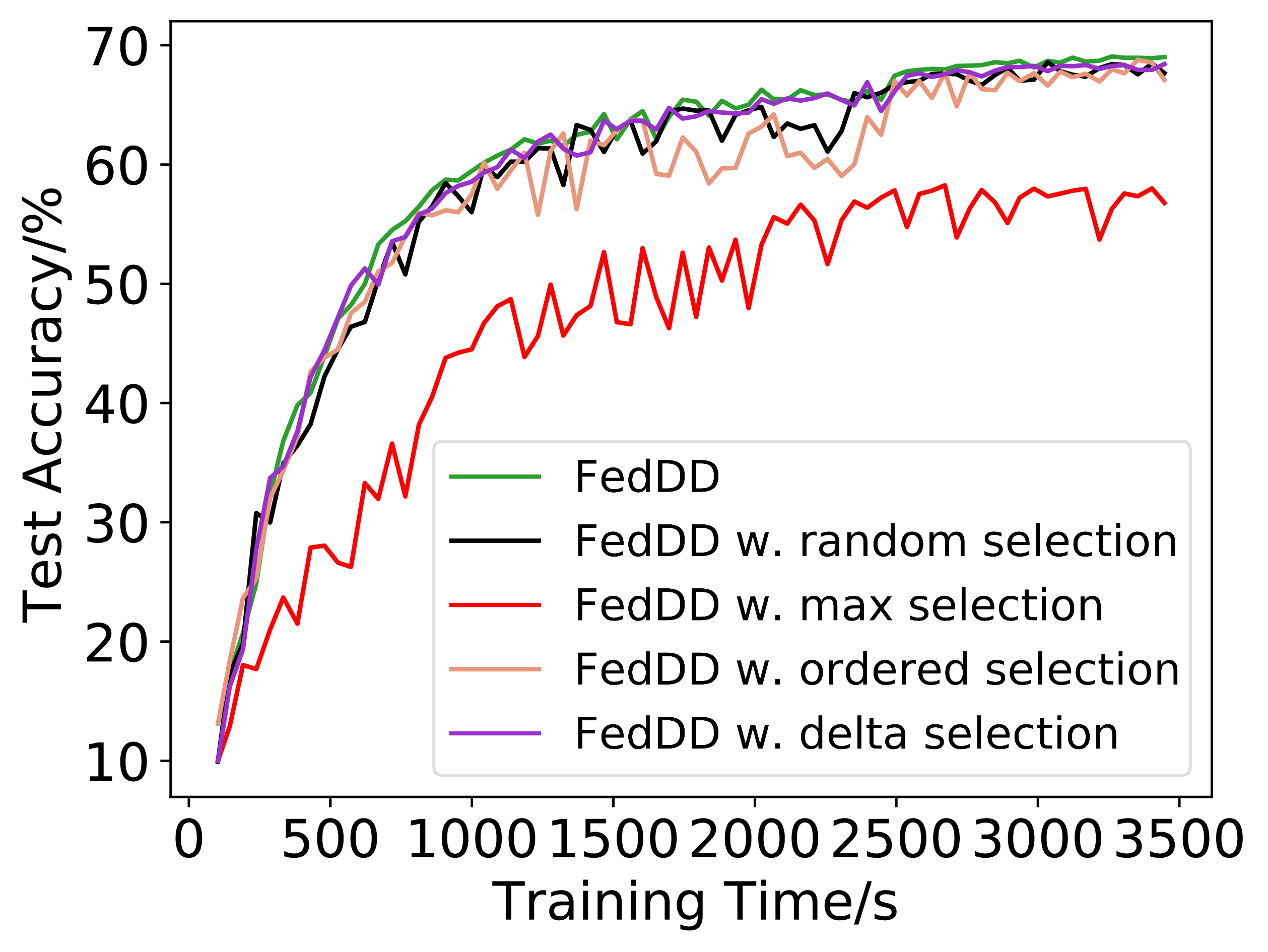}}
	\subfigure[Non-IID-b]{
		\includegraphics[width=0.48\linewidth]{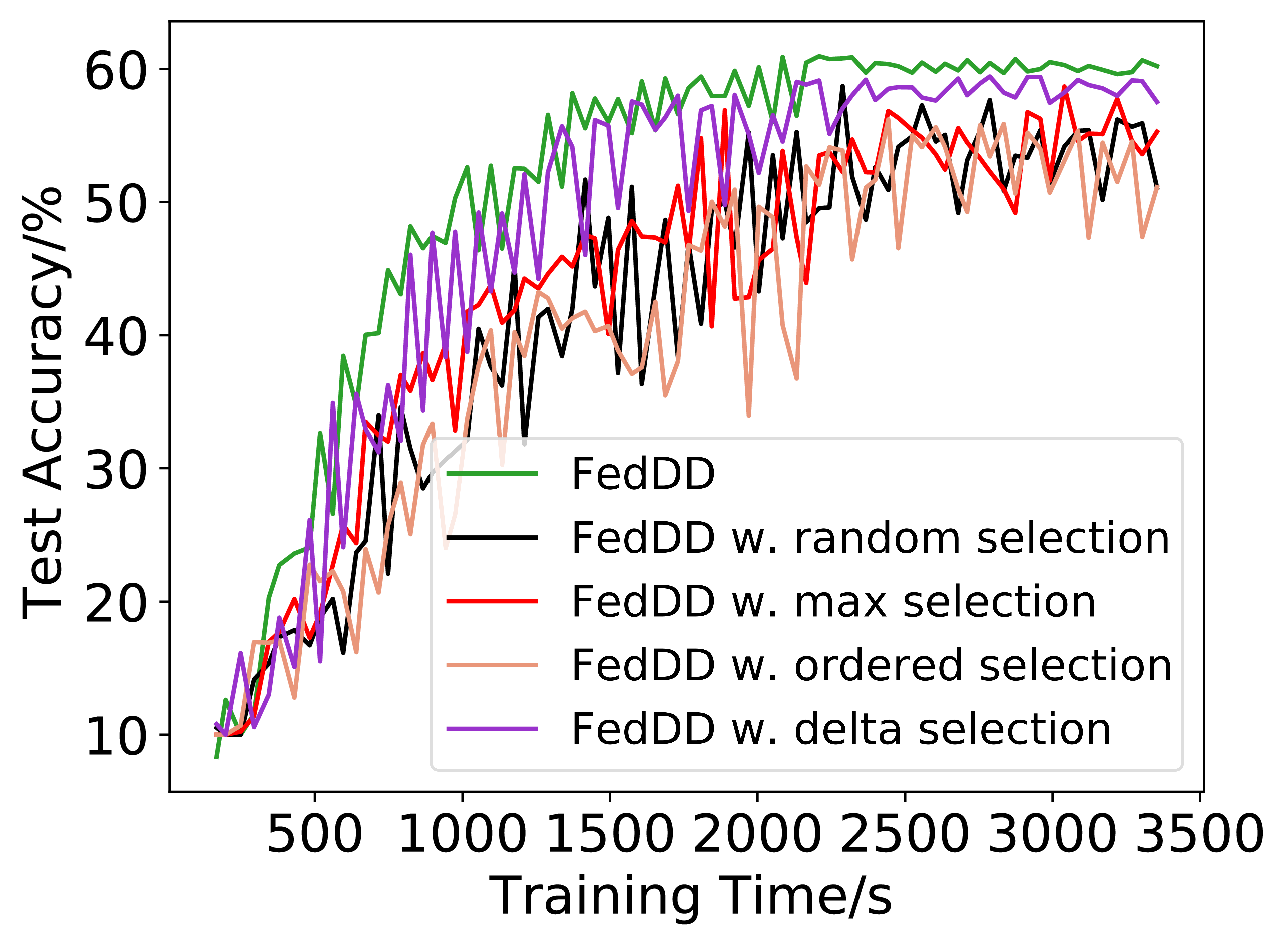}}
	\caption{Curves of top-1 accuracy with different parameter selection schemes on CIFAR10 under model-homogeneous setting in testbed.}
	\label{CIFAR10_homogeneous_Ablation_testbed}
\end{figure}

\subsection{Performance Comparisons with FedDD Variant Schemes}

Fig. \ref{MNIST_homogeneous_Ablation} shows the test accuracy of FedDD with different parameter selection schemes on MNIST under model-homogeneous setting. The results of Fig. \ref{MNIST_homogeneous_Ablation}(a) and Fig. \ref{MNIST_homogeneous_Ablation}(b) show that all the parameter selection schemes have the similar performance under IID and Non-IID-a setting, because the classification task of MNIST is relatively simple and the degree of data heterogeneity is low. Under Non-IID-b (high data heterogeneity), the convergence speed of random selection is lower than our proposed scheme and delta selection, while max selection and ordered selection perform worst and converge to two significantly low test accuracies.

Fig. \ref{FMNIST_homogeneous_Ablation} shows the test accuracy of FedDD with different parameter selection schemes on FMNIST under model-homogeneous setting. From Fig. \ref{FMNIST_homogeneous_Ablation}(a) and Fig. \ref{FMNIST_homogeneous_Ablation}(b), we can see that all the parameter selection schemes have similar performance on FMNIST under IID and Non-IID-a setting. Fig. \ref{FMNIST_homogeneous_Ablation}(c) shows our proposed parameter selection scheme has the best performance on FMNIST under Non-IID-b setting, followed by ordered selection, delta selection, random selection, and max selection.

Fig. \ref{CIFAR10_homogeneous_Ablation} shows the test accuracy of FedDD with different parameter selection schemes on CIFAR10 under model-homogeneous setting. The results of Fig. \ref{CIFAR10_homogeneous_Ablation} show that our selection scheme has the best convergence performance under different data distribution setting, while max selection and delta selection perform the second best under IID and Non-IID-a(b) setting, respectively.

Fig. \ref{CIFAR10_homogeneous_Ablation_testbed} shows the test accuracy of FedDD with different parameter selection schemes on CIFAR10 under model-homogeneous setting in testbed. From Fig. \ref{CIFAR10_homogeneous_Ablation_testbed}(a), we can see that all the parameter selection schemes have similar performance on CIFAR10 under IID setting. From Fig. \ref{CIFAR10_homogeneous_Ablation_testbed}(b), we can see that our proposed parameter selection scheme and delta selection have the fastest convergence speed and stable convergence performance, while the performance of ordered selection and random selection are not stable. Moreover, max selection converges to the lowest final accuracy. From Fig. \ref{CIFAR10_homogeneous_Ablation_testbed}(c), we can see that our proposed parameter selection scheme has the best performance, while the performance of other schemes is unstable.

\begin{figure}[htbp]
	\centering  
	\subfigbottomskip=2pt 
	\subfigcapskip=-5pt 
 	\subfigure[IID and model-heterogeneous-a setting]{
		\includegraphics[width=0.48\linewidth]{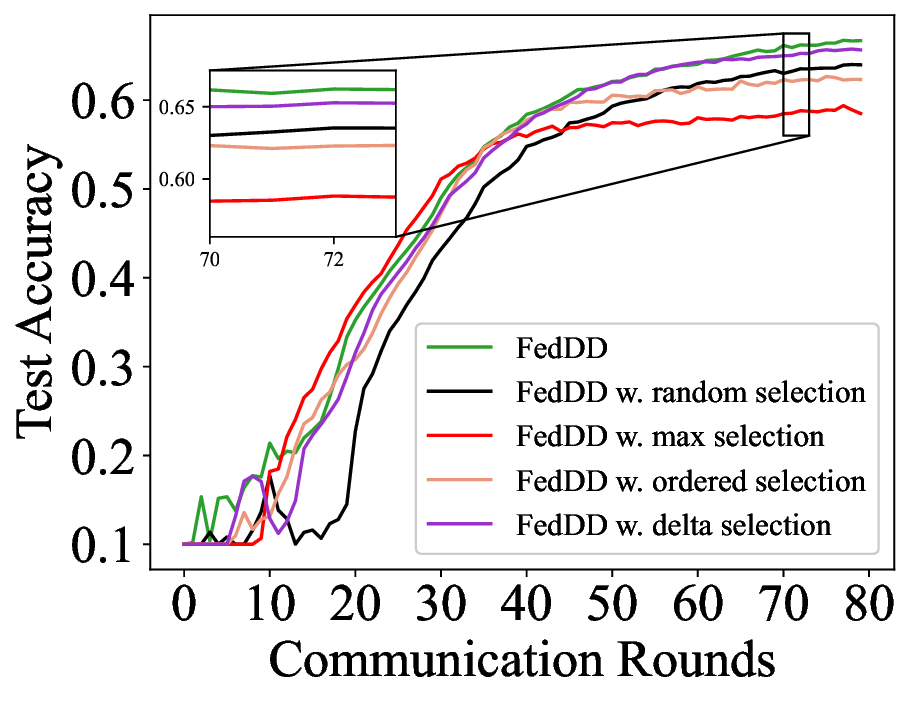}}
    \subfigure[IID and model-heterogeneous-b setting]{
		\includegraphics[width=0.48\linewidth]{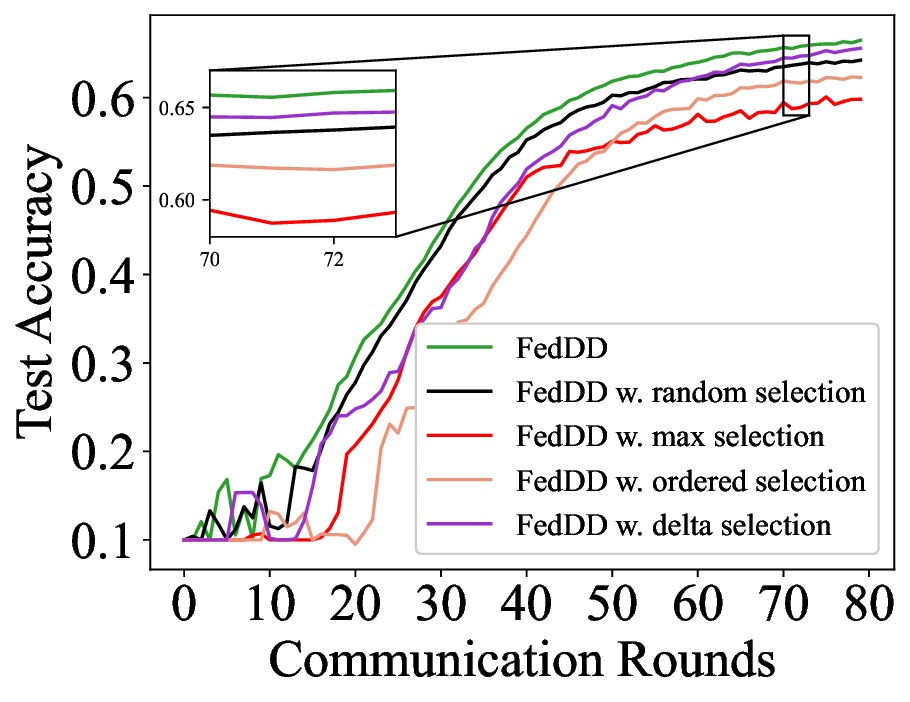}}
	\subfigure[Non-IID-a and model-heterogeneous-a setting]{
		\includegraphics[width=0.48\linewidth]{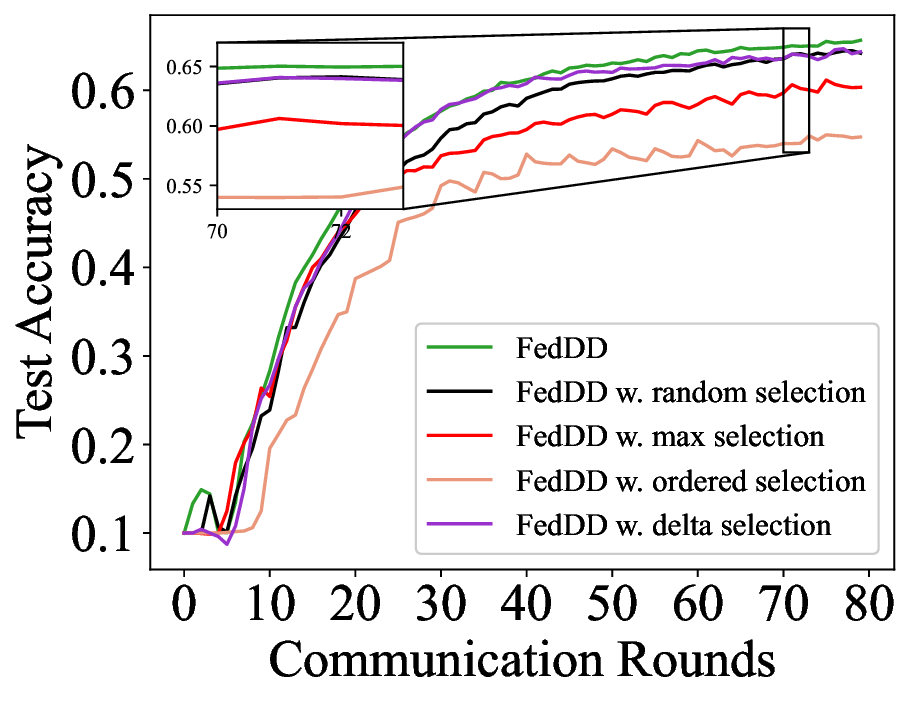}}
    \subfigure[Non-IID-a and model-heterogeneous-b setting]{
		\includegraphics[width=0.48\linewidth]{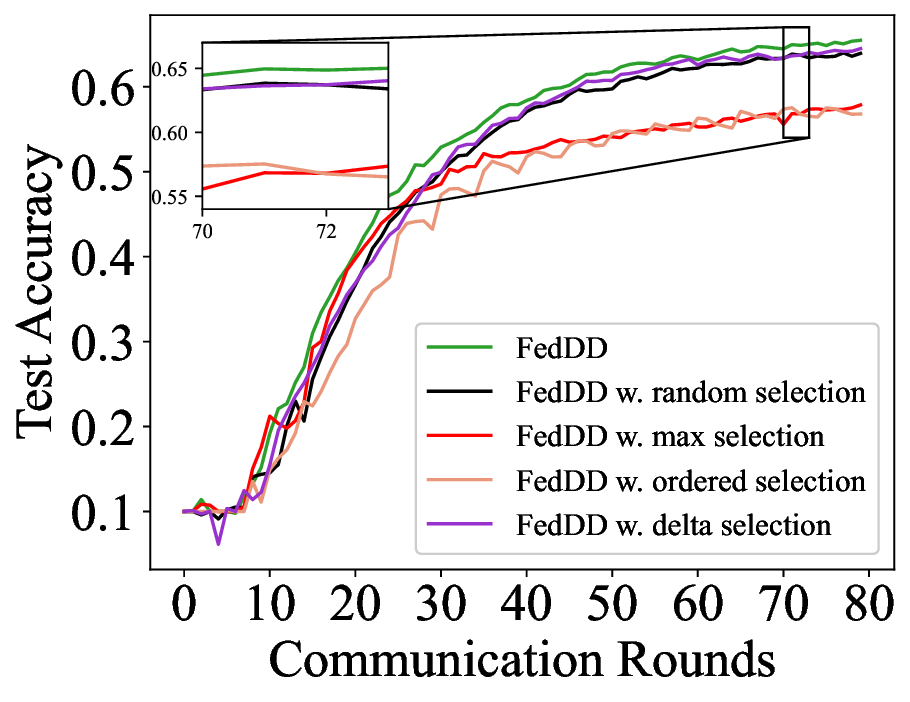}}
	\subfigure[Non-IID-b and model-heterogeneous-a setting]{
		\includegraphics[width=0.48\linewidth]{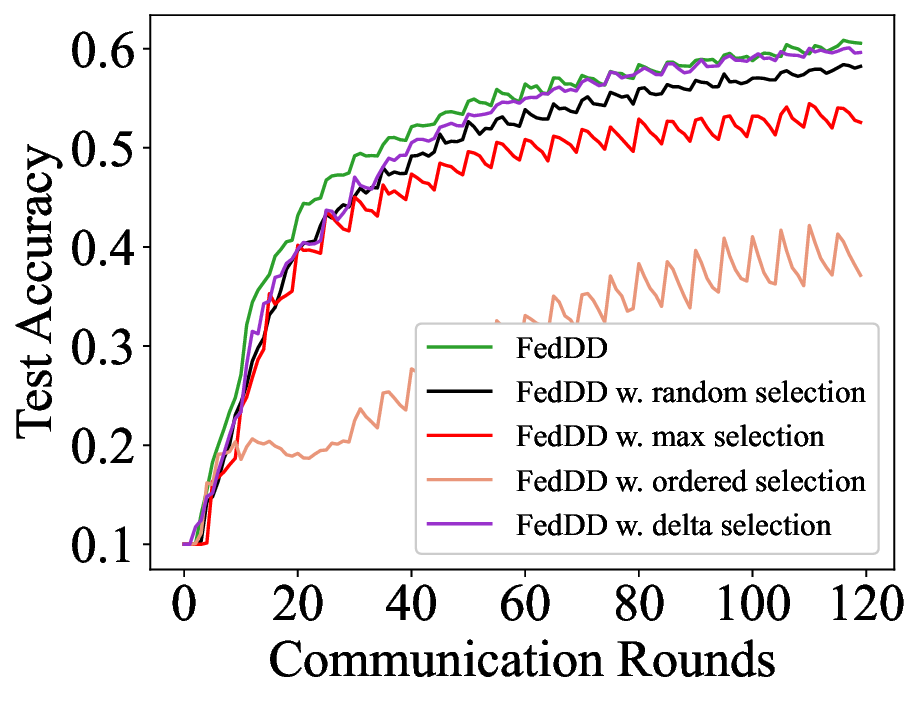}}
	\subfigure[Non-IID-b and model-heterogeneous-b setting]{
		\includegraphics[width=0.48\linewidth]{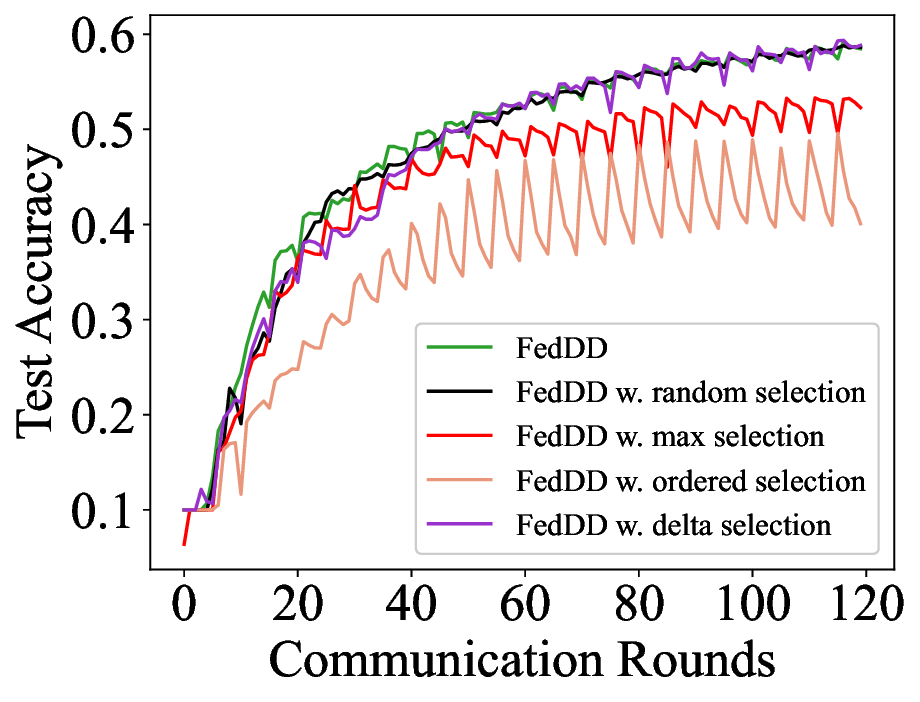}}
	\caption{Curves of top-1 accuracy with different parameter selection schemes on CIFAR10 under model-heterogeneous setting in simulation.}
	\label{ABLATION_heterogeneous}
\end{figure}

Fig. \ref{ABLATION_heterogeneous} shows that the performance difference of different selection schemes is more obvious under model-heterogeneous setting. Our parameter selection scheme outperforms other selection schemes under all data distribution setting. Other selection schemes perform differently in different scenarios. For example, random selection has similar performance with our schemes under Non-IID-a and model-heterogeneous-b setting (Fig. \ref{ABLATION_heterogeneous}(d)), but performs worse under Non-IID-b and model-heterogeneous-a setting (Fig. \ref{ABLATION_heterogeneous}(e)). We can conclude that our selection scheme is robust to all scenarios.

\begin{figure}[htbp]
	\centering  
	\subfigbottomskip=2pt 
	\subfigcapskip=-5pt 
 	\subfigure[MNIST under Non-IID-a setting]{
		\includegraphics[width=0.45\linewidth]{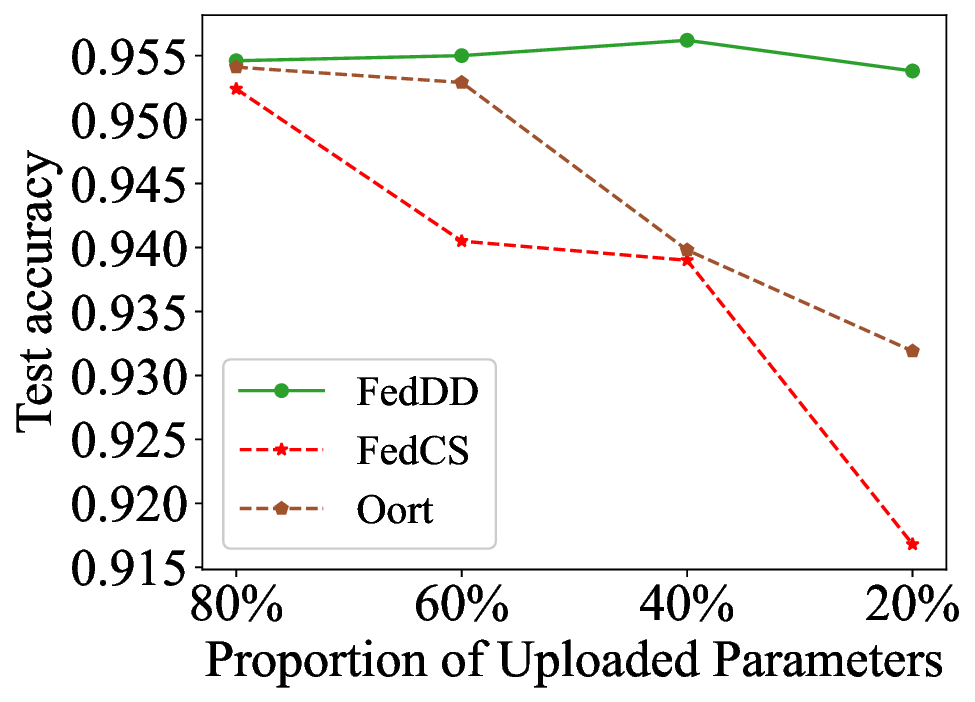}}
  	\subfigure[MNIST under Non-IID-b setting]{
		\includegraphics[width=0.45\linewidth]{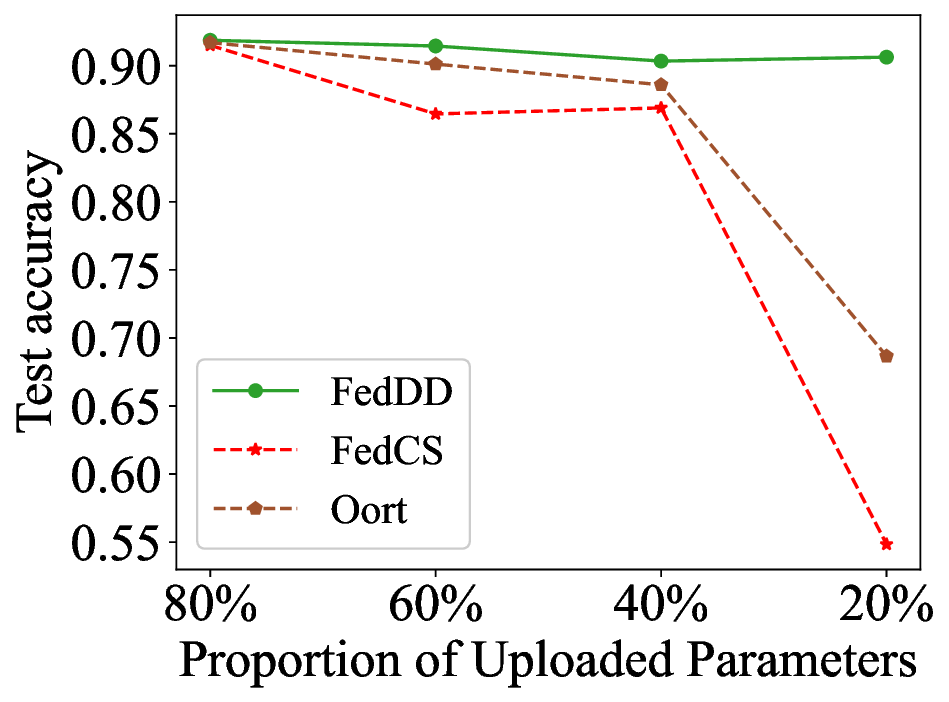}}
	\subfigure[FMNIST under Non-IID-a setting]{
		\includegraphics[width=0.45\linewidth]{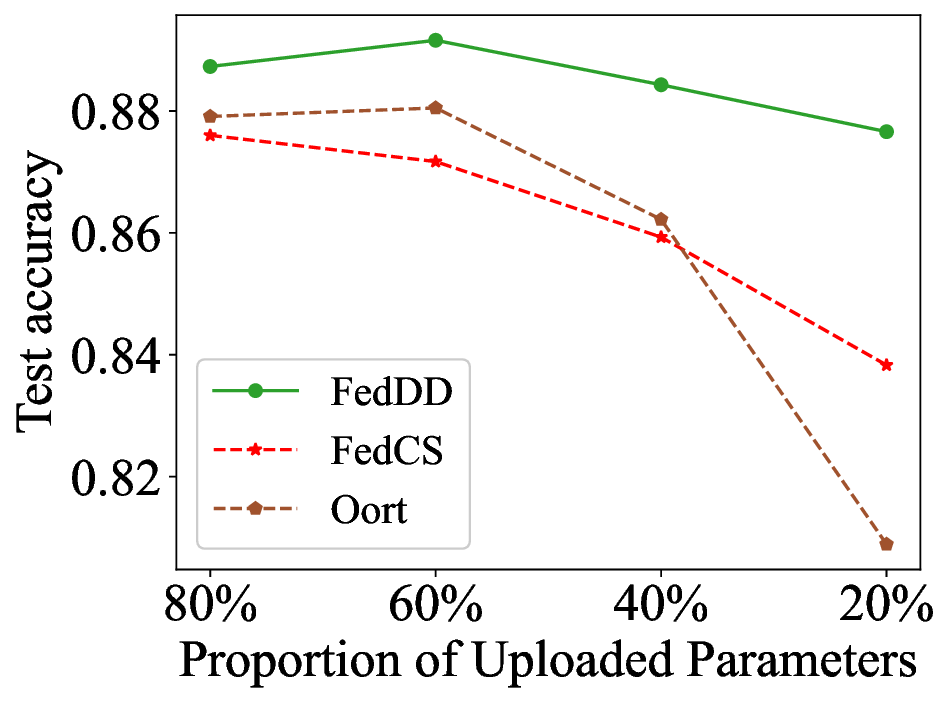}}
  	\subfigure[FMNIST under Non-IID-b setting]{
		\includegraphics[width=0.45\linewidth]{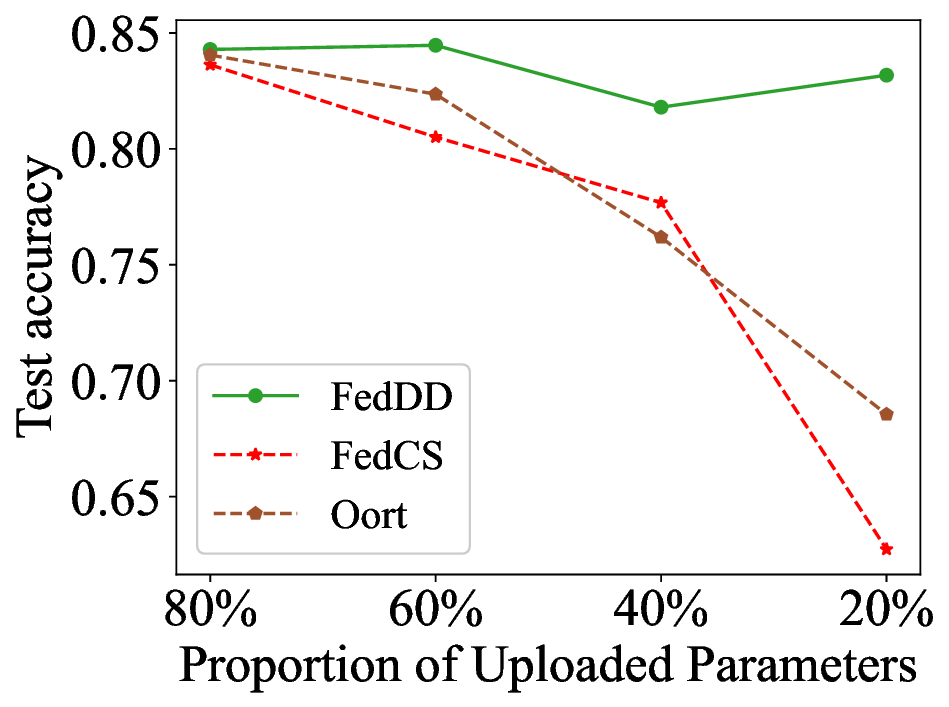}}
	\subfigure[CIFAR10 under Non-IID-a setting]{
		\includegraphics[width=0.45\linewidth]{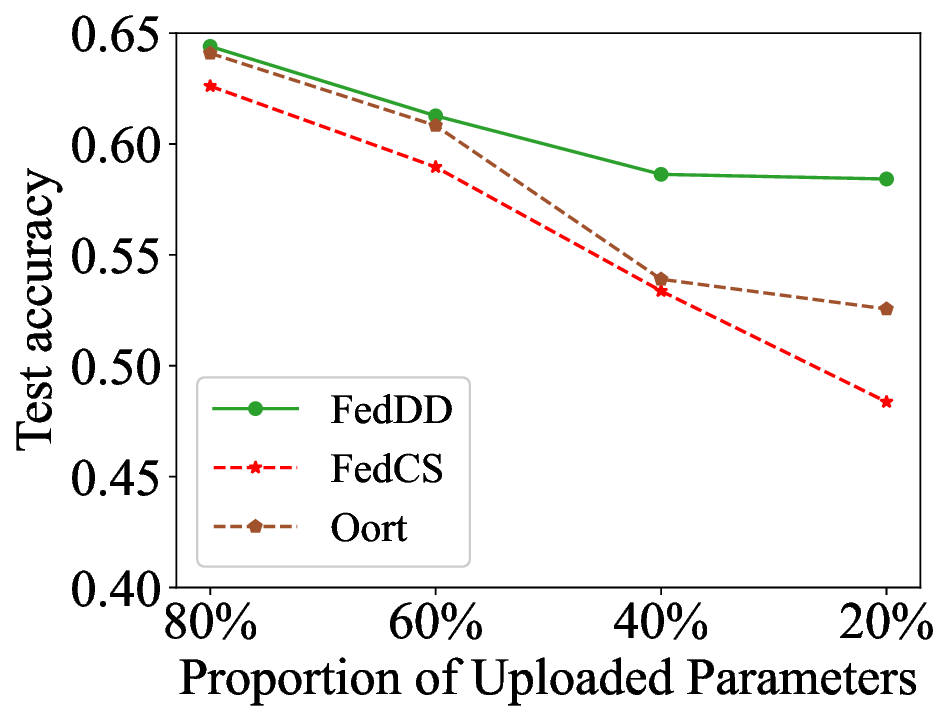}}
	\subfigure[CIFAR10 under Non-IID-b setting]{
		\includegraphics[width=0.45\linewidth]{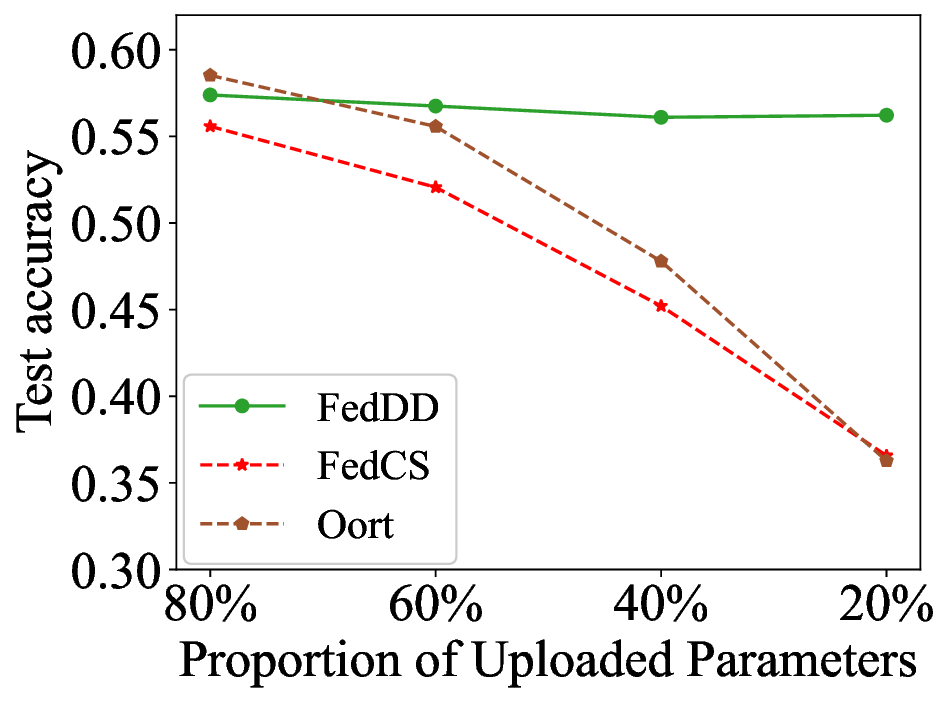}}
	\caption{The final test accuracy with different proportion of uploaded parameters under model-homogeneous setting in simulation.}
	\label{Sensitivity_homogeneity_noniid}
\end{figure}

\begin{figure}[htb]
	\centering  
	\subfigbottomskip=2pt 
	\subfigcapskip=-5pt 
 	\subfigure[Non-IID-a and model-heterogeneous-a setting]{
		\includegraphics[width=0.45\linewidth]{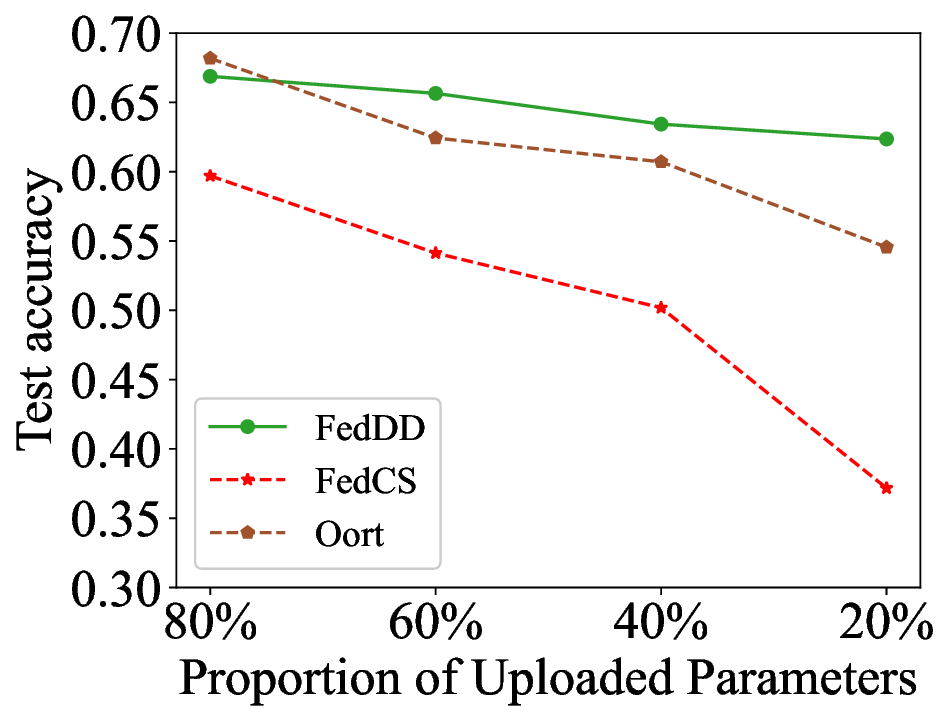}}
  	\subfigure[Non-IID-b and model-heterogeneous-a setting]{
		\includegraphics[width=0.45\linewidth]{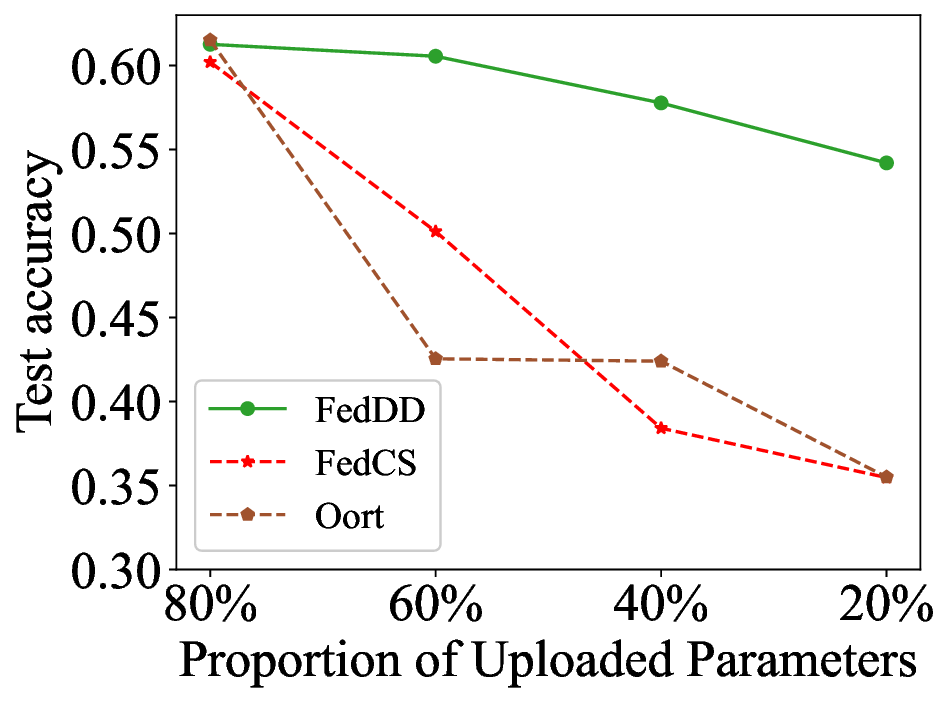}}
   	\subfigure[Non-IID-a and model-heterogeneous-b setting]{
		\includegraphics[width=0.45\linewidth]{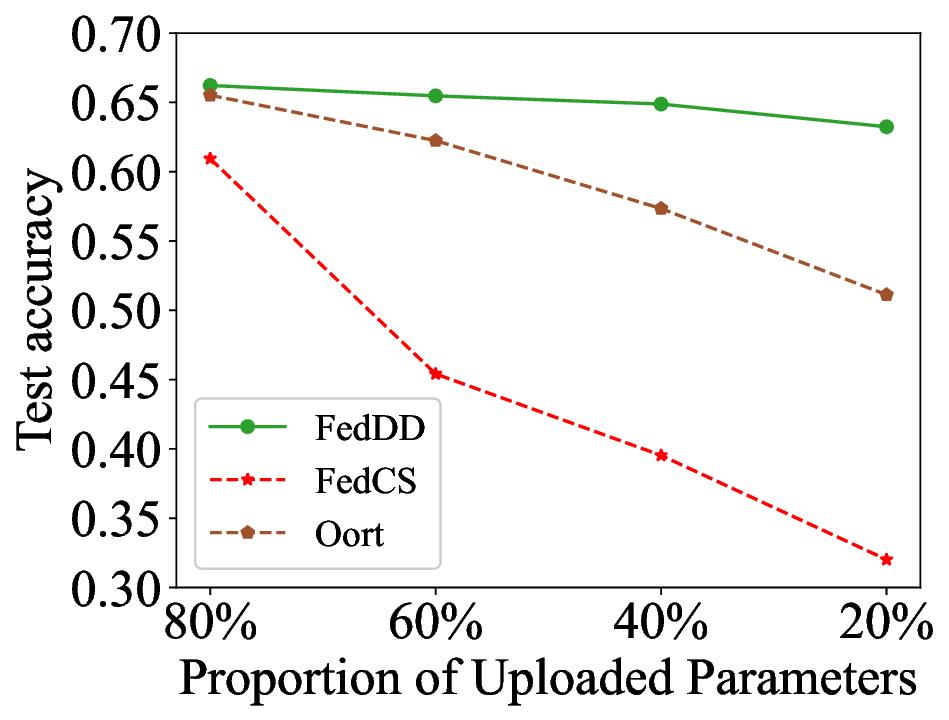}}
     	\subfigure[Non-IID-b and model-heterogeneous-b setting]{
		\includegraphics[width=0.45\linewidth]{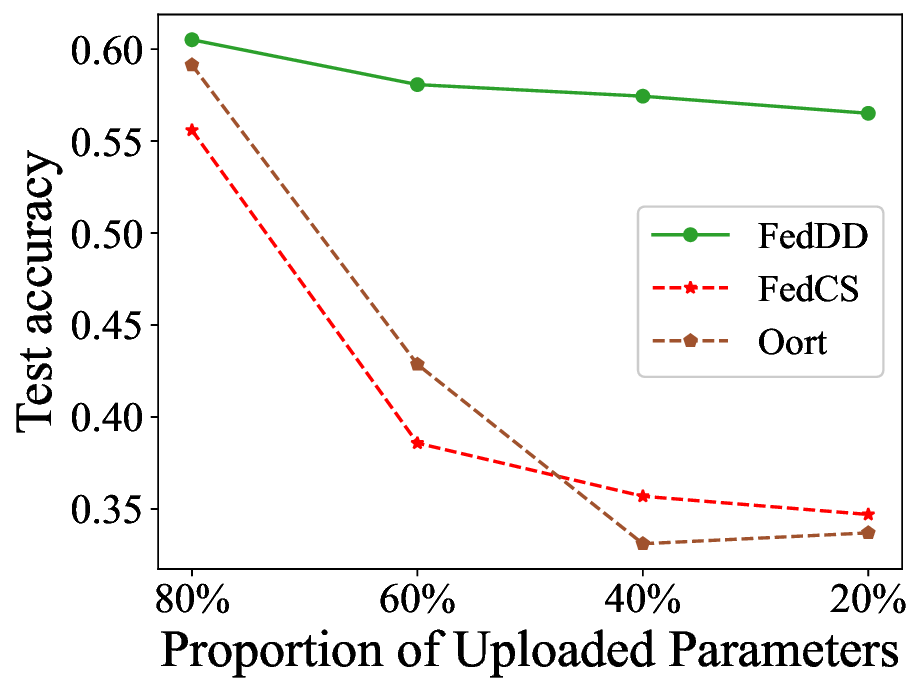}}
	\caption{The final test accuracy with different proportion of uploaded parameters under model-heterogeneous setting in simulation.}
	\label{Sensitivity_heterogeneousA_noniid}
\end{figure}

\subsection{Sensitivity Analysis}

\textbf{Robustness to diverse communication budgets.} To clearly show the superiority of FedDD, we further investigate the parameter sensitivity for different FL methods. We evaluate how the testing performance of different methods change as the proportion of uploaded parameters decreases in this section. Fig. \ref{Sensitivity_homogeneity_noniid} and Fig. \ref{Sensitivity_heterogeneousA_noniid} illustrate that the testing performance of FedDD stay relative stable as $A_{\rm server}$ decreases, while the performance of client-selection-based methods deteriorates rapidly. When the proportion of uploaded parameters decreases from 80\% to 20\%, the final test accuracy of FedDD decreases by 0.1\% to 11.5\% under different setting, while the final test accuracies of FedCS and Oort decrease by 3.7\% to 47.5\% and 2.3\% to 43.0\%, respectively. We can conclude from the sensitivity analysis that in the case of limited communication resources, we should select more clients to participate in FL through the parameter dropout, rather than reducing the number of clients participating in FL, in order to improve the generalization of global model. Thus, FedDD is more communication-efficient and is much more robust to the dynamic communication budgets with a stable model accuracy, compared with the client-selection-based schemes.

\begin{figure}[htbp]
	\centering  
	\subfigbottomskip=2pt 
	\subfigcapskip=-5pt 
 	\subfigure[Model-heterogeneous-a]{
		\includegraphics[width=0.45\linewidth]{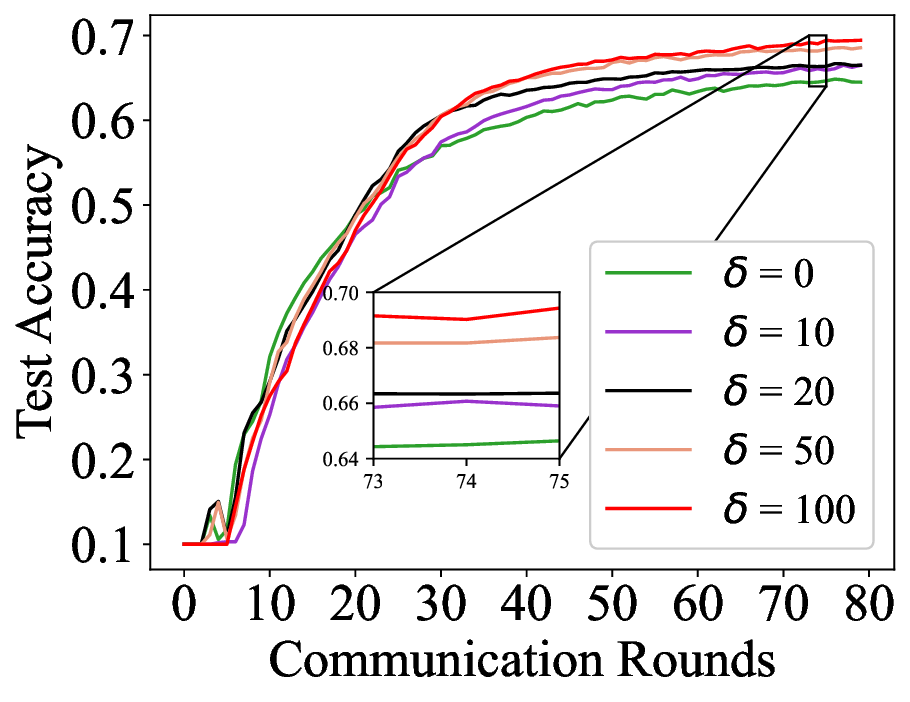}}
  	\subfigure[Model-heterogeneous-b]{
		\includegraphics[width=0.45\linewidth]{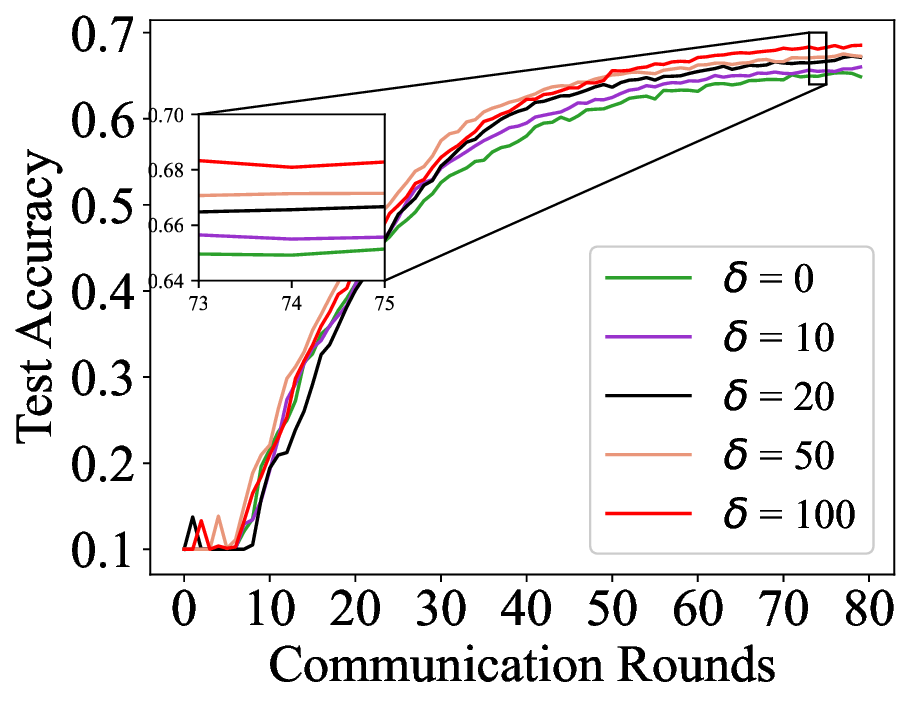}}
	\caption{The test accuracy with different penalty factor $\delta$ under Non-IID-a and model-heterogeneous setting in simulation.}
	\label{Sensitivity_heterogeneous_noniidA_penalty_factor}
\end{figure}

\textbf{Impact of penalty factor $\delta$.} The penalty factor $\delta$ are used to balance the effect among system heterogeneity, data heterogeneity and model heterogeneity. A larger $\delta$ let server pay more attention to data heterogeneity and model heterogeneity. Fig. \ref{Sensitivity_heterogeneous_noniidA_penalty_factor} illustrates the testing performance with different $\delta$ under Non-IID-a and model-heterogeneous setting. Zero $\delta$ means that the server only considers system heterogeneity, while non-zero $\delta$ has a better trade-off between communication efficiency and final test accuracy, and help global model converge to a higher test accuracy scarifying communication efficiency. The server can set the value of $\delta$ based on its preference towards the trade-off between communication efficiency and the final accuracy.

\begin{figure}[htbp]
	\centering
	\subfigbottomskip=2pt 
	\subfigcapskip=-5pt 
	\subfigure[IID]{
		\includegraphics[width=0.45\linewidth]{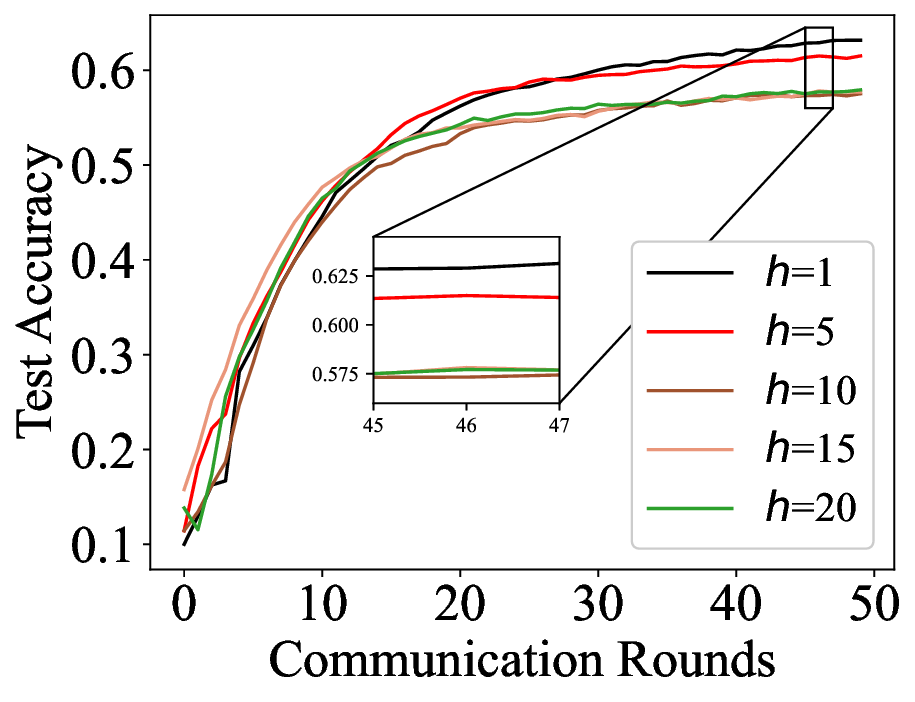}}
	\subfigure[Non-IID-a]{
		\includegraphics[width=0.45\linewidth]{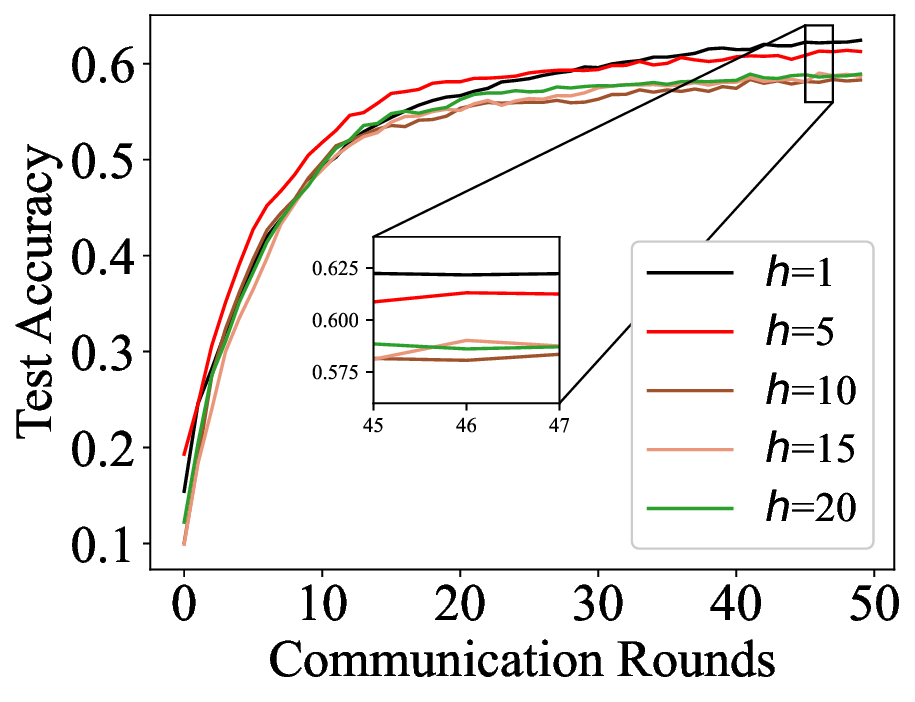}}
	\subfigure[Non-IID-b]{
		\includegraphics[width=0.45\linewidth]{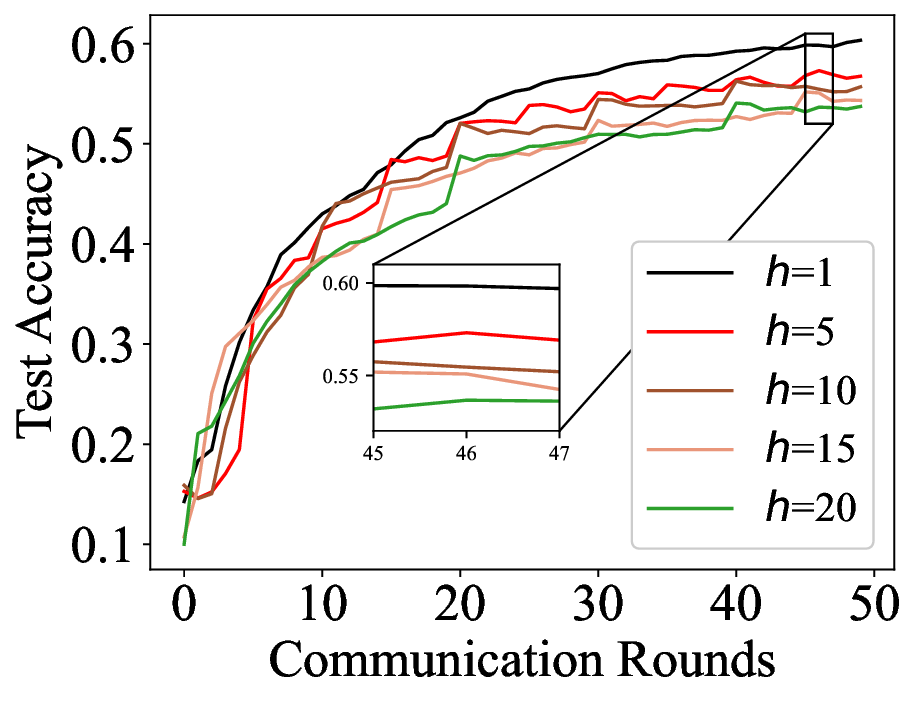}}
 	\caption{The test accuracy with different full model broadcast period $h$ on CIFAR10 under model-homogeneous setting in simulation.}
	\label{Sensitivity_homogeneous_noniidA_H}
\end{figure}

\begin{figure}[htbp]
	\centering
	\subfigbottomskip=2pt 
	\subfigcapskip=-5pt 
	\subfigure[Model-heterogeneous-a]{
		\includegraphics[width=0.45\linewidth]{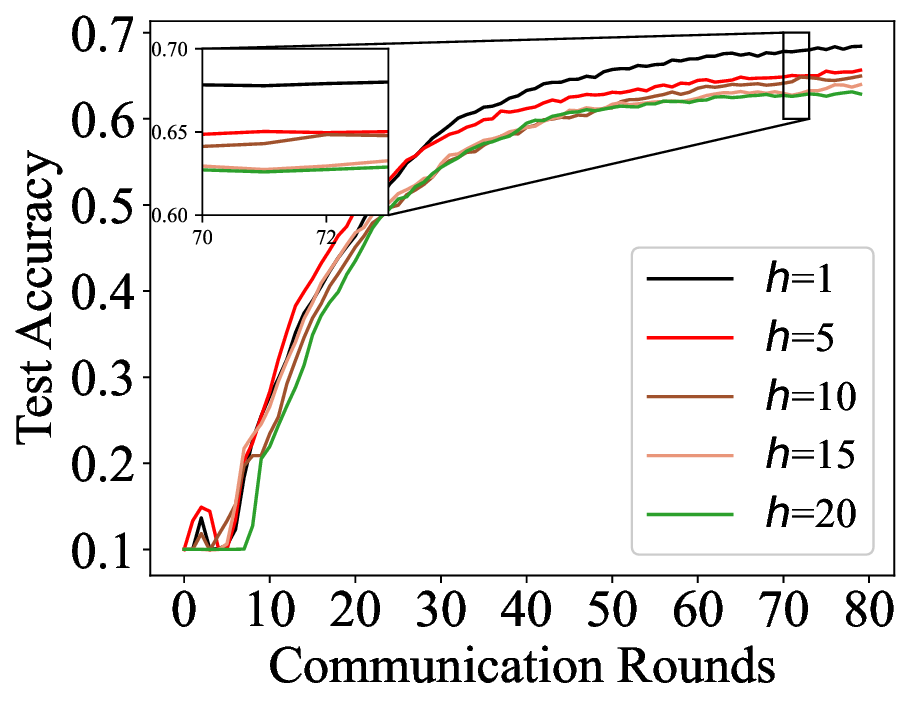}}
	\subfigure[Model-heterogeneous-b]{
		\includegraphics[width=0.45\linewidth]{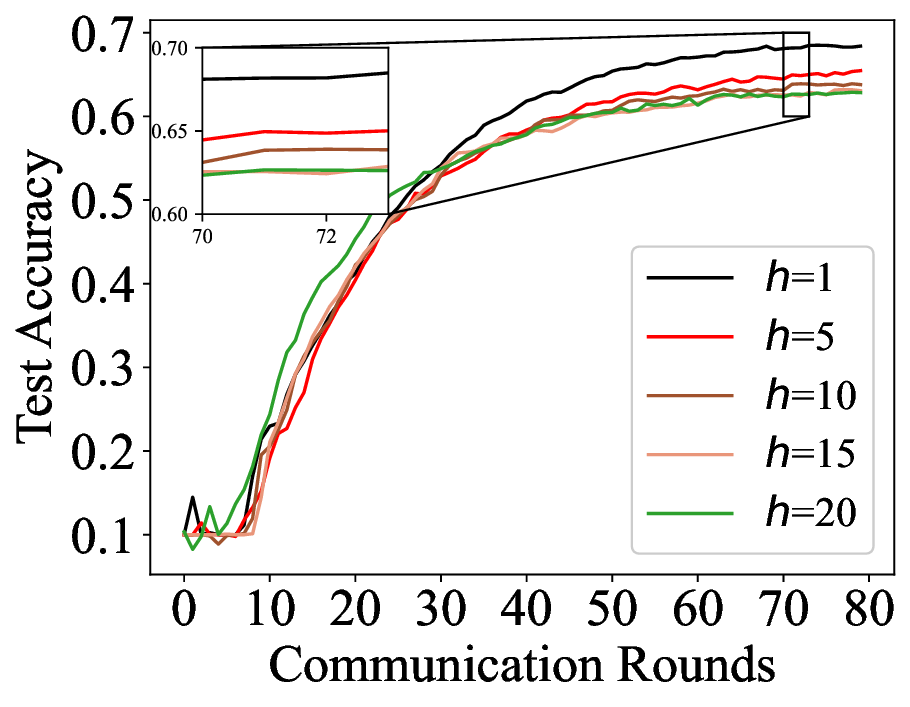}}
	\caption{The test accuracy with different full model broadcast period $h$ on CIFAR10 under Non-IID-a and model-heterogeneous setting in simulation.}
	\label{Sensitivity_heterogeneous_noniidA_H}
\end{figure}

\textbf{Impact of full model broadcast period $h$.} As mentioned in Section \ref{section:Convergence analysis}, the convergence bound increases as $h$ get larger, which is consistent with the experimental results of Fig. \ref{Sensitivity_homogeneous_noniidA_H} and Fig. \ref{Sensitivity_heterogeneous_noniidA_H}. As shown in Fig. \ref{Sensitivity_homogeneous_noniidA_H} and Fig. \ref{Sensitivity_heterogeneous_noniidA_H}, $h$ has little influence on the early stage of model training, but will affect the final test accuracy. Therefore, in order to balance the trade-off between communication efficiency and testing performance, the server can set a larger $h$ at the early stage of model training and a smaller $h$ at the later stage of model training.

\begin{figure*}[htbp]
	\centering  
	\subfigbottomskip=2pt 
	\subfigcapskip=-5pt 
	\subfigure[MNIST]{
		\includegraphics[width=0.3\linewidth]{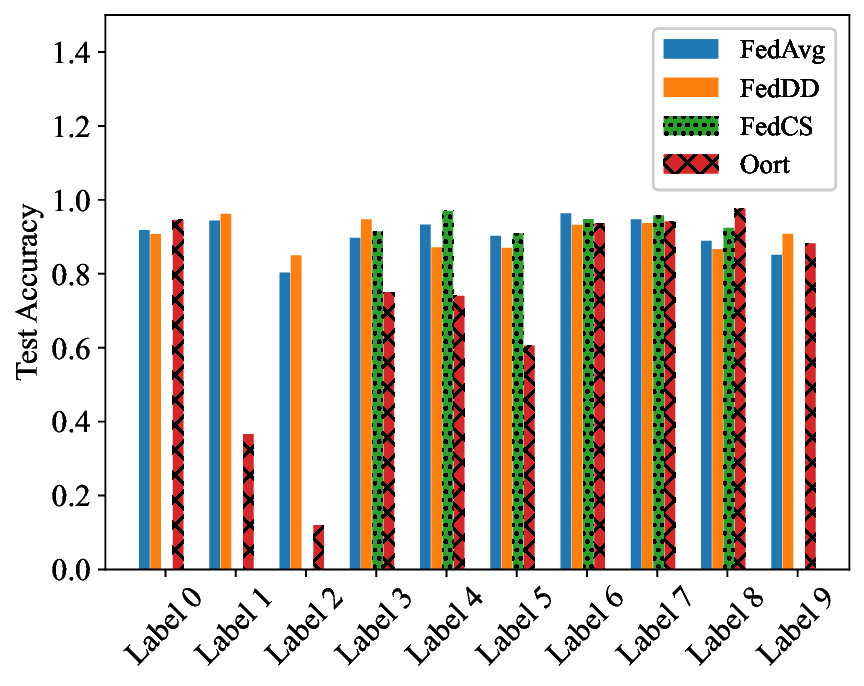}}
	\subfigure[FMNIST]{
		\includegraphics[width=0.3\linewidth]{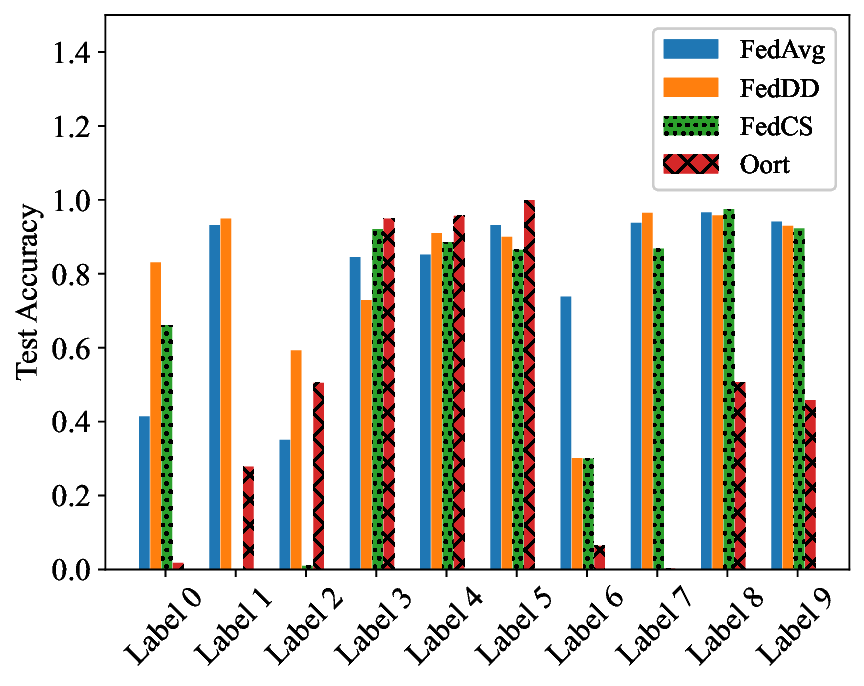}}
	\subfigure[CIFAR10]{
		\includegraphics[width=0.3\linewidth]{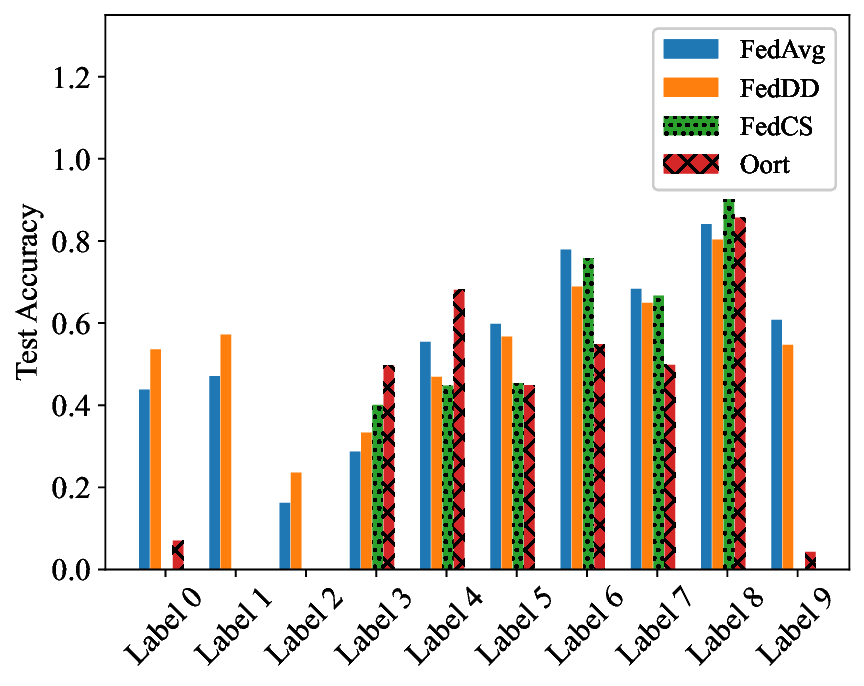}}
	\caption{The test accuracy of different labels on class-imbalanced dataset in simulation.}
	\label{fig:Generalization}
\end{figure*}

\subsection{Generalization on Class-imbalanced Datasets}
Class-imbalance refers to the unequal number of samples of different classes in global dataset, which is the union of all the  available client local datasets in FL\cite{zhang2022fed}. We create a global dataset, in which 7 classes have $n_1$ samples each, and the remaining 3 classes (rare classes) have $n_2$ samples each. The ratio between $n_1$ and $n_2$ is set to 1 : 0.4. Like Non-IID-b setting, each client is assigned with three classes.

Fig. \ref{fig:Generalization} illustrates that the final test accuracy of different FL schemes on three datasets, where label 0, label 1 and label 2 are set as rare classes. FedCS and Oort select 20\% clients to upload full model parameters, while FedDD let all clients upload sparse model parameters and the uploaded proportion $A_{\rm server}$ are also set to 20\%. 

From Fig. \ref{fig:Generalization}(a), we can see that FedCS can not classify the 3 rare classes with test accuracy of 0 and Oort has a poor performance on label 1 and label 2 with test accuracy of 36.7\% and 12.0\%, while FedDD is as good as FedAvg in the classification performance on both common classes and rare classes with test accuracy of 85.0\% to 96.2\%. As shown in Fig. \ref{fig:Generalization}(b), for FMNIST dataset, FedCS can not classify label 1 and label 2 with test accuracy of 0 and 1\%, while Oort has a poor performance on label 0 and label 1 with test accuracy of 1.8\% and 27.8\%. On the contrary, the test accuracy of the 3 rare classes of FedDD are 83.1\%, 94.9\%, and 59.3\%. From Fig. \ref{fig:Generalization}(c), we can see that FedDD has the similar performance with FedAvg, and even outperform FedAvg in testing the 3 rare classes with test accuracy of 23.6\% to 57.2\%. On the contrary, client-selection-based methods fail to classify the 3 rare classes with test accuracy close to 0. From the above results, we can conclude that our method has a good generalization performance on class-imbalanced dataset, even when only a small proportion of model parameters are uploaded.

\section{Related Work}\label{section:Related Work}

\subsection{Generalization in FL}
Traditional distributed machine learning assumes that data is independent and identically distributed (IID), which generally does not hold in FL. Data heterogeneity will significantly degrade the generalization performance of FL. Generalization in FL can be defined as the testing performance toward different data labels or different clients. According to different scenarios, methods for dealing with data heterogeneity are roughly divided into personalized FL\cite{ouyang2021clusterfl,t2020personalized} and non-personalized FL \cite{wang2020tackling,luo2021no,wang2019adaptive,zhou2022efficient}. Personalized FL aims to create a personalized model for each client that is suitable for their own data distribution while the generalization of global model is not the focus. Non-personalized FL focuses on improving the generalization of the global model and reducing the adverse effects of data heterogeneity. Wang \emph{et al.} claimed that naive weighted aggregation in heterogeneous FL will cause the global model converges to a stationary point different from the true objective and propose a normalized averaging method (FedNova) to eliminate objective inconsistency while preserving fast convergence\cite{wang2020tackling}. Motivated by the finding that data heterogeneity brings a greater bias in the classifier than other layers and classification performance can be significantly improved by post-calibrating the classifier after federated training, Luo \emph{et al.} proposed Classifier Calibration with Virtual Representations (CCVR), which adjusts the classifier using virtual representations sampled from an approximated Gaussian mixture model\cite{luo2021no}. However, the above works do not consider the generalization degradation problem under limited resource constraint. Focusing on the frequency of global model aggregation under a fixed resource budget, Wang \emph{et al.} proposed a control algorithm that learns the data distribution, system dynamics and model characteristics, where it dynamically adapts the frequency of global aggregation in real time to minimize the learning loss\cite{wang2019adaptive}. Considering the clients scheduling problem in multi-job FL, Zhou \emph{et al.} combined the capability and data fairness in the cost model to improve the efficiency of the training process and the accuracy of the global model\cite{zhou2022efficient}, but this work lacks convergence analysis.

\subsection{Straggler problem in Federated Learning} 
Various works had been proposed to address the straggler problem in FL, which can be roughly divided into two groups: in an asynchronous manner \cite{ma2021fedsa, avdiukhin2021federated,9562486, 10021868, chen2021fedsa} and in a synchronous manner \cite{nishio2019client, wang2020optimizing, lai2021oort, xu2020client}. Dmitrii Avdyukhin \emph{et al.} proposed an asynchronous version of local SGD wherein the clients can communicate with the server at arbitrary time intervals \cite{avdiukhin2021federated}. In \cite{9562486}, three Asynchronous Learning-aware transmission Scheduling (ALS) algorithms named ALSA-PI, BALSA, and BALSA-PO, are proposed to improve the effectivity score of FL in wireless networks, where statistical information regarding uncertainty is known, unknown, or limited. Q. Ma \emph{et al.} designed a semi-asynchronous FL mechanism called FedSA with convergence analysis, where the  server aggregates a certain number of local updates in each round in the order of arrival \cite{ma2021fedsa}. Although asynchronous FL can mitigate the straggler effect, it brings other problems, such as the staleness effect. The staleness effect means that due to the different frequencies of model parameter uploading, the stale model slows down the convergence of the global model. Therefore, in asynchronous FL, it is often necessary to introduce other complex mechanisms to control the staleness effect \cite{10021868}\cite{chen2021fedsa}, which makes the algorithm more complicated. Compared with asynchronous FL, synchronous FL naturally does not have the problem of staleness effect. To tackle the straggler problem and improve communication efficiency in synchronous FL, many researchers focused on client-selection-based methods, in which partial clients are selected to upload full models to the parameter server for global aggregation. In \cite{nishio2019client}, an update aggregation method named FedCS is developed to aggregate as many client updates as possible under one-round processing time constraint. Wang \emph{et al.} utilized Deep Reinforcement Learning (DRL) to select participating clients with the objective of minimizing the number of required communication rounds under non-IID setting\cite{wang2020optimizing}. This work focuses on data heterogeneity but does not consider system heterogeneity. Motivated by the ideas from \cite{nishio2019client} and \cite{wang2020optimizing}, Lai \emph{et al.} highlighted the tension between statistical efficiency and system efficiency when selecting FL participants and presented Oort to effectively balance the trade-off\cite{lai2021oort}. Xu \emph{et al.} considered client selection and bandwidth allocation in a long time scale and then designed a Lyapunov-based optimization algorithm to solve the optimization problem with long-term constraints\cite{xu2020client}. The existing works \cite{yang2020federated,wang2021cooperative} show the convergence of training can be accelerated with more involved clients. But the above client-selection-based methods require clients to upload the full model, which would reduce the number of participating clients under limited communication resources and degrade the generalization performance of the global model.


\subsection{Parameter Sparsification Methods in FL}
FL requires massive model parameter exchange between the server and clients, which causes a large amount of communication overhead. In order to improve communication efficiency, some works focus on parameter sparsification. Strom first proposed a parameter sparsification method, in which only gradients larger than the predefined threshold are selected for uploading\cite{strom2015scalable}, but the predefined threshold is hard to determine. Aji \emph{et al.} fixed the sparsity rate and only exchange sparse updates instead of dense updates with the biggest magnitude of each gradient\cite{aji-heafield-2017-sparse}. Sattler \emph{et al.} considered that communication-efficient FL should compress both upstream and downstream communications and robust to Non-IID setting, small batch sizes, unbalanced data and partial client participation\cite{sattler2019robust}. Shah \emph{et al.} proposed a communication-efficient FL by separately applying sparse reconstruction for the server model to address downstream communication and two algorithms to produce a compressed model for the client models to address upstream communication depending on the available computational resources\cite{9660377}. Sun \emph{et al.} proposed a mechanism named gradient correction, which makes working nodes exchange optimizer-corrected gradients to ensure convergence to a large extent\cite{sun2020toward}. Yi \emph{et al.} proposed a novel FL framework named QSFL to reduce FL uplink communication overhead at both client-level and model-level\cite{yi2022qsfl}. At the client level, the authors designed a Qualification Judgment (QJ) algorithm to sample high-qualification clients to upload models. At the model level, the authors explored a Sparse Cyclic Sliding Segment (SCSS) algorithm to further compress transmitted models. However, the above works are concerned with how to select parameters, but ignore the issue of how many parameters should be uploaded in heterogeneous scenarios. Several researchers focused on the idea of Federated Dropout\cite{caldas2018expanding,9484526,horvath2021fjord,wen2022federated}, which is similar to model pruning\cite{li2021hermes}, where the structures of local models keep changing as training progresses. Our paper is orthogonal to these works. The structure of local models in this work do not change, because our goal is to train a global model with good generalization performance, rather than training personalized models for different clients.

\section{Conclusion}\label{section:Conclusion and Discussion}

The proposed FedDD framework allocates differential dropout rates to clients by solving a convex optimization problem, considering system heterogeneity, statistical heterogeneity and model heterogeneity, in order to improve communication efficiency. The FedDD speeds up the convergence of global model by cherry-picking the uploaded parameters. Compared to existing client-selection-based schemes, the proposed FedDD can effectively utilize all clients' data and computing resources to improve the generalization of global model. Convergence analysis and extensive experiments verify that the FedDD can get a better performance of time to accuracy with marginal final accuracy degradation, as long as we carefully select the dropout rate and upload parameters. 

For the future work, we are going to explore the novel paradigms based on the combination of both client selection and parameter dropout, in order to fully leverage the benefits of these two approaches for an integrated design.




\begin{IEEEbiography}[{\includegraphics[width=1in,height=1.25in,clip,keepaspectratio]{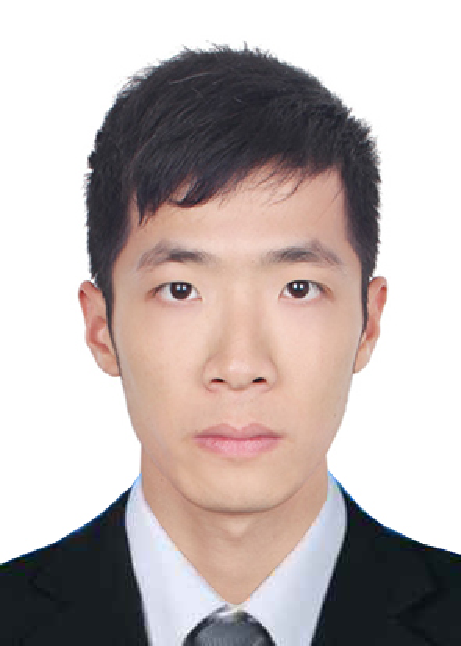}}]{Zhiying Feng} received his B.E. and M.E. degrees in electrical engineering from South China University of Technology (SCUT), Guangzhou, China, in 2016 and 2019, respectively. He is currently working toward the Ph.D. degree with the School of Computer Science and Engineering, Sun Yat-sen University (SYSU). His primary research interests include, mobile edge computing, and federated learning.

\end{IEEEbiography}

\begin{IEEEbiography}[{\includegraphics[width=1in,height=1.25in,clip,keepaspectratio]{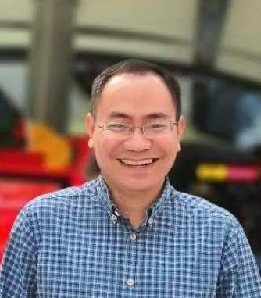}}]{Xu Chen}(Senior Member, IEEE) received his Ph.D. degree in information engineering from the Chinese  University of Hong Kong, in 2012. He is a full professor with Sun Yat-sen University, Guangzhou, China, and the vice director of National and Local Joint  Engineering Laboratory of Digital Home Interactive  Applications. He worked as a postdoctoral research  associate with Arizona State University, Tempe, from  2012 to 2014, and a Humboldt Scholar fellow with the Institute of Computer Science, University of Goettingen, Germany, from 2014 to 2016. He received the prestigious Humboldt research fellowship awarded by the Alexander von  Humboldt Foundation of Germany, 2014 Hong Kong Young Scientist Runner-up  Award, 2017 IEEE Communication Society Asia-Pacific Outstanding Young  Researcher Award, 2017 IEEE ComSoc Young Professional Best Paper Award,  Honorable Mention Award of 2010 IEEE international conference on Intelligence and Security Informatics (ISI), Best Paper Runner-up Award of 2014 IEEE  International Conference on Computer Communications (INFOCOM), and Best  Paper Award of 2017 IEEE International Conference on Communications (ICC).  He is currently an area editor of the IEEE Open Journal of the Communications  Society, an associate editor of the IEEE Transactions Wireless Communications,  IEEE Internet of Things Journal and IEEE Journal on Selected Areas in Communications (JSAC) Series on Network Softwarization and Enablers.
\end{IEEEbiography}

\begin{IEEEbiography}[{\includegraphics[width=1in,height=1.25in,clip,keepaspectratio]{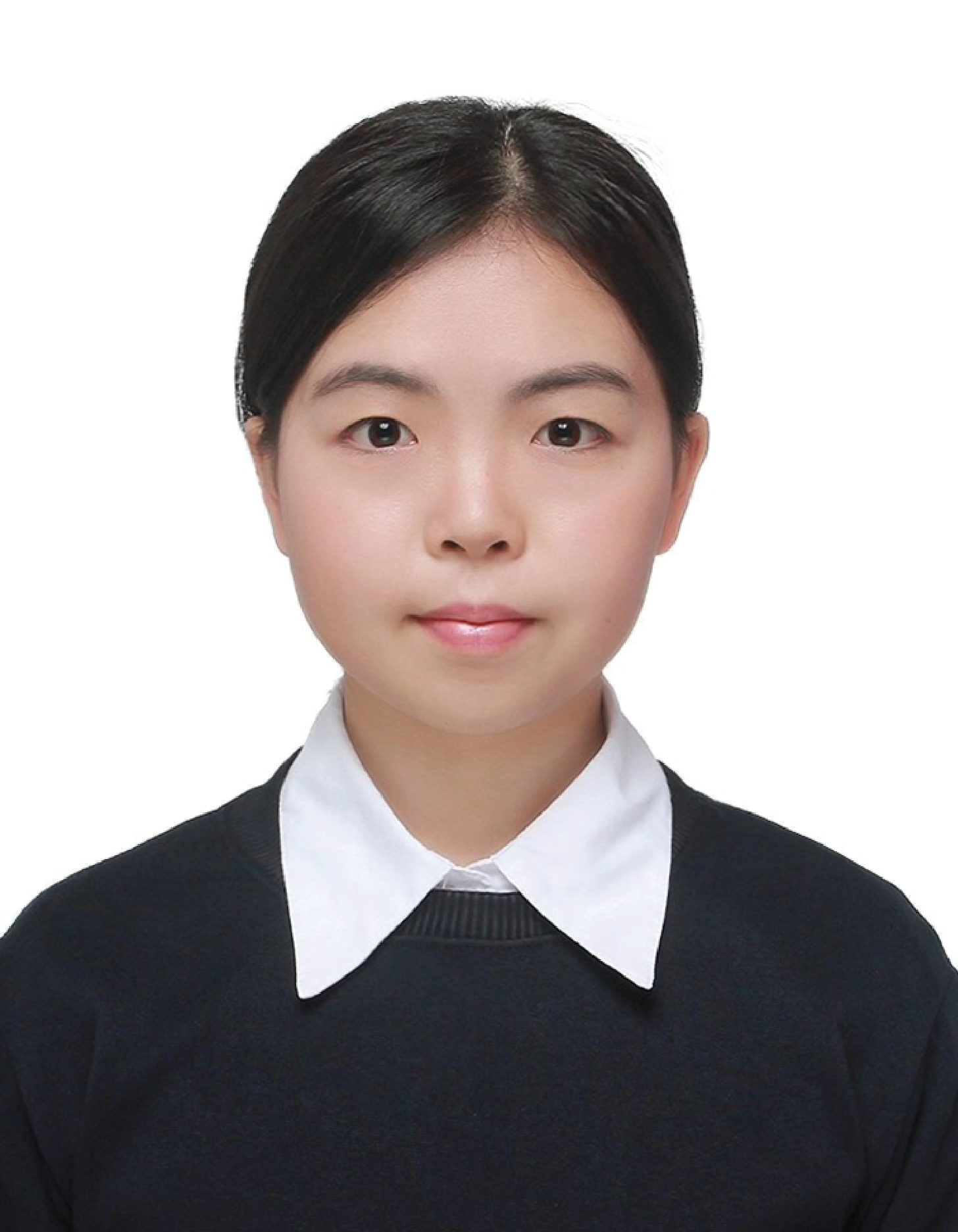}}]{Qiong Wu}
received her B.S. and M.E. degrees from the School of Data and Computer Science, Sun Yat-sen University (SYSU), Guangzhou, China, in 2017 and 2019, respectively. She is currently working toward the Ph.D. degree with the School of Computer Science and Engineering, Sun Yat-sen University. Her primary research interests include social data analysis, mobile edge computing, and federated learning.
\end{IEEEbiography}

\begin{IEEEbiography}[{\includegraphics[width=1in,height=1.25in,clip,keepaspectratio]{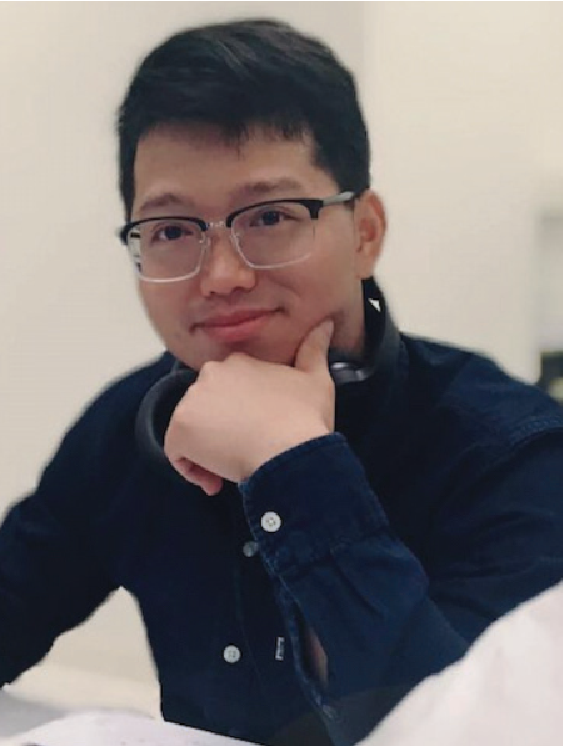}}]{Wen Wu}(Senior Member, IEEE) received his Ph.D. degree in Electrical and Computer Engineering from University of Waterloo, Waterloo, ON, Canada, in  2019. He received his B.E. degree in Information Engineering from South China University of Technology (SCUT), Guangzhou, China, and his M.E. degree in Electrical Engineering from University of Science and Technology of China, Hefei, China, in  2012 and 2015, respectively. He worked as a Postdoctoral Fellow with the Department of Electrical and Computer Engineering, University of Waterloo. He is currently an Associate Researcher at the Frontier Research Center, Peng Cheng Laboratory, Shenzhen, China. His research interests include 6G  networks, network intelligence, and network virtualization.
\end{IEEEbiography}

\begin{IEEEbiography}[{\includegraphics[width=1in,height=1.25in,clip,keepaspectratio]{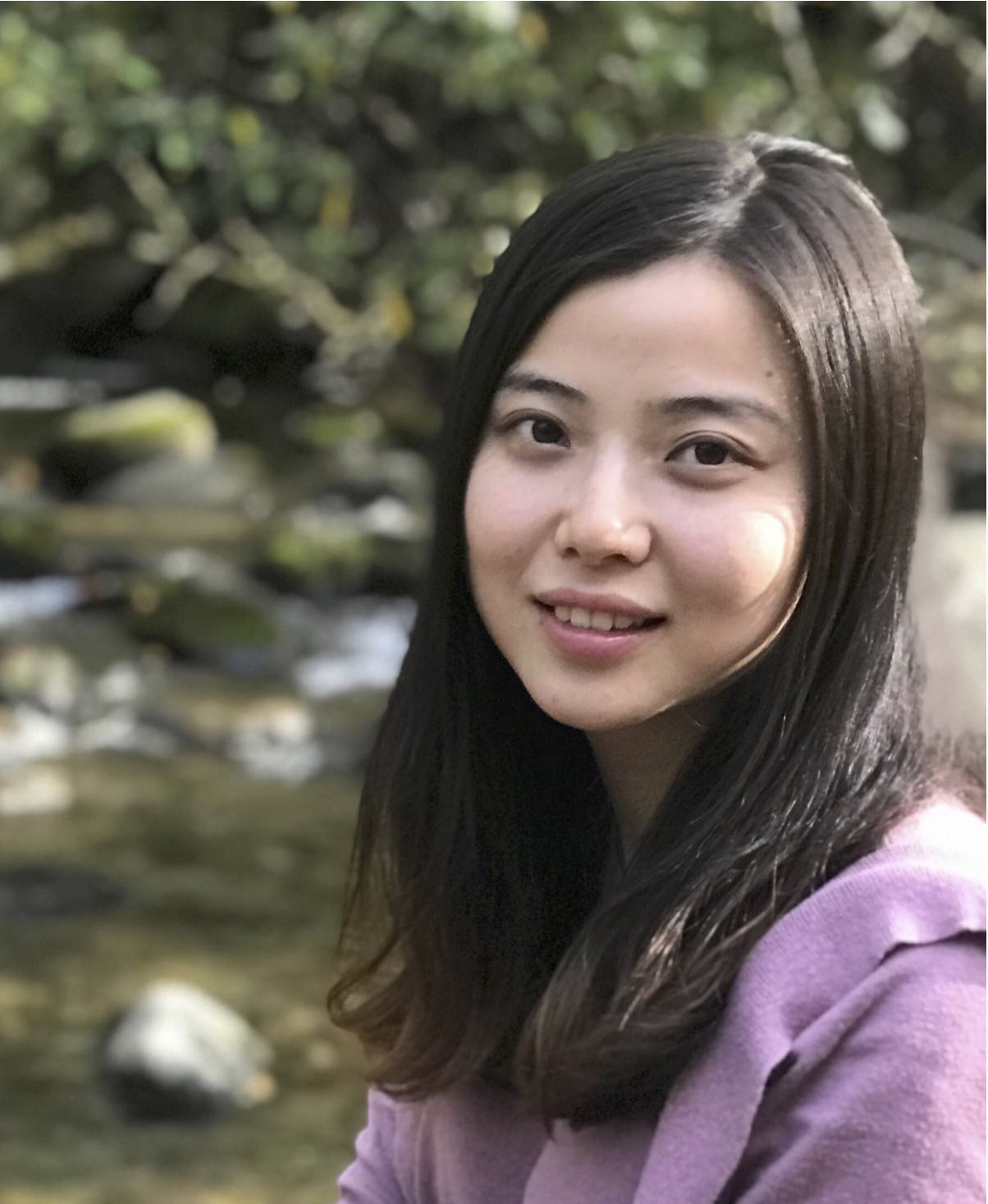}}]{Xiaoxi Zhang}(Member, IEEE) received her B.E. degree in electronics and information engineering from the Huazhong University of Science and Technology in 2013 and her Ph.D. degree in computer science from The University of Hong Kong in 2017. She is currently an Associate Professor with the School of Computer Science and Engineering, Sun Yat-sen University. Before joining SYSU, she was a Post-Doctoral Researcher with the Department of Electrical and Computer Engineering, Carnegie Mellon University. She is broadly interested in optimization and algorithm design for networked systems, including cloud and edge computing networks, NFV systems, and distributed machine learning systems.
\end{IEEEbiography}

\begin{IEEEbiography}[{\includegraphics[width=1in,height=1.25in,clip,keepaspectratio]{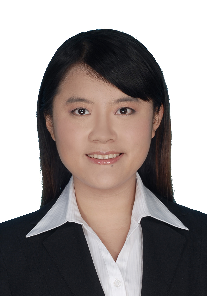}}]{Qianyi Huang}(Member, IEEE) received her bachelor’s degree in computer science from Shanghai Jiao Tong University and her Ph.D. degree from the Department of Computer Science and Engineering, The Hong Kong University of Science and Technology. She is currently an associate professor at Sun Yat-sen University, Guangzhou, China. She has authored or co-authored many papers in top-ranking journals and conferences, including the IEEE/ACM Transactions on Networking (TON), IEEE Transactions on Mobile Computing (TMC), MobiCom, UbiComp, and INFOCOM. Her research interests include mobile computing, integrated sensing and communication, Internet-of-Things, and security.
\end{IEEEbiography}

\onecolumn

\appendices

\section{Configurations of Model-Heterogeneous-b Setting}\label{Configurations of Model-Heterogeneous-b Setting}

\begin{table*}[h]
\centering
\scriptsize
\setlength{\abovecaptionskip}{0pt}
\setlength{\belowcaptionskip}{0pt}
\caption{Summary of Model-Heterogeneous-b Configurations.}\label{tab:modelheterogeneousB}
\begin{tabular}{c|c|c|c|c}
 \hline
\textbf{Full model (Sub-model-1)} & \textbf{Sub-model-2} & \textbf{Sub-model-3}& \textbf{Sub-model-4}& \textbf{Sub-model-5}\\
\hline
Conv(3, 64, kernel=3) & Conv(3, 64, kernel=3)  &  Conv(3, 64, kernel=3)  &  Conv(3, 32, kernel=3)  &   Conv(3, 32, kernel=3)    \\
ReLU   &  ReLU &   ReLU    &   ReLU  &  ReLU    \\
Maxpool  & Maxpool  & Maxpool  &  Maxpool   &   Maxpool   \\
Conv(64, 128, kernel=3) & Conv(64, 128, kernel=3)  & Conv(64, 128, kernel=3) & Conv(32, 96, kernel=3)    &    Conv(32, 96, kernel=3)    \\
ReLU   &  ReLU &   ReLU    &  ReLU   &  ReLU    \\
Maxpool  & Maxpool  &   Maxpool    & Maxpool    & Maxpool     \\
Conv(128, 256, kernel=3) & Conv(128, 256, kernel=3)  & Conv(128, 256, kernel=3)    &   Conv(96, 256, kernel=3)  &  Conv(96, 128, kernel=3)      \\
ReLU   &  ReLU &   ReLU    &  ReLU   &  ReLU    \\
Maxpool  &  Maxpool &    Maxpool   &  Maxpool   & Maxpool     \\
Conv(256, 512, kernel=3) &  Conv(256, 256, kernel=3) &  Conv(256, 256, kernel=3)     &  Conv(256, 256, kernel=3)   & Conv(128, 128, kernel=3)       \\
ReLU   & ReLU   &   ReLU    &  ReLU   &    ReLU  \\
Maxpool  & Maxpool  &  Maxpool   & Maxpool    &  Maxpool    \\
Conv(512, 512, kernel=3) & Conv(256, 256, kernel=3)  & Conv(256, 256, kernel=3)   &  Conv(256, 256, kernel=3)   &     Conv(128, 256, kernel=3)   \\
ReLU   &  ReLU & ReLU   &  ReLU   &  ReLU    \\
Maxpool  &  Maxpool &  Maxpool     & Maxpool    &    Maxpool  \\
FC(512, 100) & FC(512, 100)  &   FC(512, 80)    &   FC(512, 80)  &  FC(512, 80)      \\
ReLU   &  ReLU & ReLU &  ReLU   &  ReLU    \\
FC(100, 100) & FC(100, 100)  &    FC(80, 80)   &   FC(80, 80)  &  FC(80, 80)      \\
ReLU   & ReLU  &   ReLU    &   ReLU  &  ReLU   \\
FC(100, 10) & FC(100, 10)  &   FC(80, 10)    &  FC(80, 10)   & FC(80, 10)       \\
Softmax   & Softmax   & Softmax      &  Softmax   & Softmax \\
 \hline
\end{tabular}
\end{table*}

\section{Proof of Theorem 2}\label{Proof of Theorem 2}
To prove Theorem 2, we introduce virtual auxiliary variables $\bar{\bm{W}}^{t+1}$ and $\bar{\bar{\bm{W}}}^{t+1}$ to help analyze the convergence. $\bar{\bm{W}}^{t+1}$ is formulated by uploading and downloading full models. $\bar{\bar{\bm{W}}}^{t+1}$ is formulated by uploading sparse models and downloading full models. There two virtual auxiliary variables are defined as follows:

\begin{equation}\label{eq:appbar}
\begin{aligned}
\bar{\bm{W}}^{t+1}   =  \frac{1}{N} \sum_{n=1}^{N}\left(\bm{W}^{t}-\eta  \nabla  F_{n}(\bm{W}^{t})\right).
\end{aligned}
\end{equation}

\begin{equation}\label{eq:appbarbar}
\begin{aligned}
\bar{\bar{\bm{W}}}^{t+1} = \frac{\sum_{n=1}^{N} \left(\bm{W}^t - \eta \nabla  F_{n}(\bm{W}^{t})\right) \odot \bm{M}_{n}^{t+1}}{\sum_{n=1}^{N}\bm{M}_{n}^{t+1}}.
\end{aligned}
\end{equation}

From the property of the $L$-smoothness, we have:

\begin{equation}\label{eq:appsmoo1}
\begin{aligned}
F(\bm{W}^{t+1}) \leq F(\bar{\bar{\bm{W}}}^{t+1}) + (\bm{W}^{t+1}-\bar{\bar{\bm{W}}}^{t+1})^{\rm T} \nabla F(\bar{\bar{\bm{W}}}^{t+1}) + \frac{L}{2}\left\| \bm{W}^{t+1}- \bar{\bar{\bm{W}}}^{t+1}\right\|^2.
\end{aligned}
\end{equation}

\begin{equation}\label{eq:appsmoo2}
\begin{aligned}
F(\bar{\bar{\bm{W}}}^{t+1}) \leq F(\bar{\bm{W}}^{t+1}) + (\bar{\bar{\bm{W}}}^{t+1}-\bar{\bm{W}}^{t+1})^{\rm T} \nabla F(\bar{\bm{W}}^{t+1}) + \frac{L}{2}\left\| \bar{\bar{\bm{W}}}^{t+1}- \bar{\bm{W}}^{t+1}\right\|^2.
\end{aligned}
\end{equation}

\begin{equation}\label{eq:appsmoo3}
\begin{aligned}
F(\bar{\bm{W}}^{t+1}) \leq F(\bm{W}^{t}) + (\bar{\bm{W}}^{t+1}-\bm{W}^{t})^{\rm T} \nabla F(\bm{W}^{t}) + \frac{L}{2}\left\| \bar{\bm{W}}^{t+1}- \bm{W}^{t}\right\|^2.
\end{aligned}
\end{equation}

Taking the expectations of both sides of Eq. (\ref{eq:appsmoo1}), Eq. (\ref{eq:appsmoo2}) and Eq. (\ref{eq:appsmoo3}), we have the three following equations:

\begin{equation}\label{eq:app4}
\begin{aligned}
\mathbb{E}F(\bm{W}^{t+1}) \leq \mathbb{E}F(\bar{\bar{\bm{W}}}^{t+1}) + \frac{L}{2}\mathbb{E}\left\| \bm{W}^{t+1}- \bar{\bar{\bm{W}}}^{t+1}  \right\|^2.
\end{aligned}
\end{equation}

\begin{equation}\label{eq:app5}
\begin{aligned}
\mathbb{E}F(\bar{\bar{\bm{W}}}^{t+1}) \leq \mathbb{E}F(\bar{\bm{W}}^{t+1})  + \frac{L}{2}\mathbb{E}\| \bar{\bar{\bm{W}}}^{t+1}- \bar{\bm{W}}^{t+1}  \|^2.
\end{aligned}
\end{equation}

\begin{equation}\label{eq:app6}
\begin{aligned}
\mathbb{E}F(\bar{\bm{W}}^{t+1}) \leq \mathbb{E}F(\bm{W}^{t}) + \mathbb{E}((\bar{\bm{W}}^{t+1}-\bm{W}^{t})^{\rm T} \nabla F(\bm{W}^{t})) + \frac{L}{2}\mathbb{E}\left\| \bar{\bm{W}}^{t+1}- \bm{W}^{t}  \right\|^2.
\end{aligned}
\end{equation}

Adding the left and right sides of Eq. (\ref{eq:app4}), Eq. (\ref{eq:app5}) and Eq. (\ref{eq:app6}) respectively, we get

\begin{equation}\label{eq:app_aggregation}
\begin{aligned}
\mathbb{E}F(\bm{W}^{t+1})-\mathbb{E}F(\bm{W}^{t}) \leq  &   \mathbb{E}\left((\bar{\bm{W}}^{t+1}-\bm{W}^{t})^{\rm T} \nabla F(\bm{W}^{t})\right) +  \frac{L}{2}\mathbb{E}\left\| \bm{W}^{t+1}- \bar{\bar{\bm{W}}}^{t+1}  \right\|^2 \\
+ & \frac{L}{2}\mathbb{E}\left\| \bar{\bar{\bm{W}}}^{t+1}- \bar{\bm{W}}^{t+1}  \right\|^2 + \frac{L}{2}\mathbb{E}\left\| \bar{\bm{W}}^{t+1}- \bm{W}^{t}\right\|^2.
\end{aligned}
\end{equation}

The relation between $\bar{\bm{W}}^{t+1}$ and $\bm{W}^t$ can be shown as follows:

\begin{equation}\label{}
\begin{aligned}
\bar{\bm{W}}^{t+1} -  \bm{W}^t = - \frac{1}{N} \sum_{{n}=1}^{N} \eta \nabla  F_{n}(\bm{W}^t).
\end{aligned}
\end{equation}

\begin{equation}\label{eq:appWWT}
\begin{aligned}
\mathbb{E}\left((\bar{\bm{W}}^{t+1}-\bm{W}^{t})^{\rm T} \nabla F(\bm{W}^{t})\right) &= -\mathbb{E}\left(\left( \frac{1}{N} \sum_{{n}=1}^{N} \eta \nabla  F_{n}(\bm{W}^t)\right)^{\rm T} \nabla F(\bm{W}^{t})\right)\\
&= -\eta \mathbb{E}\left\| \nabla F(\bm{W}^{t})\right\|^2.
\end{aligned}
\end{equation}

\begin{equation}\label{eq:appWWT2L}
\begin{aligned}
\frac{L}{2}\mathbb{E}\left\| \bar{\bm{W}}^{t+1} - \bm{W}^t  \right\|^2 = \frac{L}{2}\mathbb{E} \left\| - \frac{1}{N} \sum_{{n}=1}^{N} \eta \nabla  F_{n}(\bm{W}^t) \right\|^2
=  \frac{L}{2} \eta^2 \mathbb{E}\left\| \nabla  F(\bm{W}^t)  \right\|^2.
\end{aligned}
\end{equation}

We introduce the following lemmas to prove Theorem 2.

\textbf{Lemma 1:} With Assumptions 1, 2 and 3, $\frac{L}{2}\mathbb{E}\left\| \bm{W}^{t+1}- \bar{\bar{\bm{W}}}^{t+1}  \right\|^2$ is bounded as follows:
\begin{equation}\label{eq:Lemma 1}
\begin{aligned}
\frac{L}{2}\mathbb{E}\left\| \bm{W}^{t+1}- \bar{\bar{\bm{W}}}^{t+1} \right \|^2 \leq \frac{L}{2}(1+ \eta^2 L^2) 2( \epsilon +1)\epsilon \frac{1}{N}\sum_{n=1}^{N}\sigma_n^2 \eta^2 + 2( \epsilon +1)\epsilon \eta^2 \mathbb{E}\left\|   \nabla F(\bm{W}^{t-1})  \right\|^2.
\end{aligned}
\end{equation}

\textbf{Lemma 2:} With Assumptions 1, 2 and 3, $\frac{L}{2}\mathbb{E}\| \bar{\bar{\bm{W}}}^{t+1}- \bar{\bm{W}}^{t+1}  \|^2 $ is bounded as follows:
\begin{equation}\label{eq:Lemma 2}
\begin{aligned}
\frac{L}{2}\mathbb{E}\left\| \bar{\bar{\bm{W}}}^{t+1}- \bar{\bm{W}}^{t+1}  \right\|^2 \leq \frac{L}{2}\epsilon \eta^2 \frac{1}{N}\sum_{n=1}^{N}\sigma_n^2  +  \frac{L}{2}\epsilon \eta^2 \mathbb{E}\left\|   \nabla  F(\bm{W}^{t}) \right\|^2.
\end{aligned}
\end{equation}

\textbf{Lemma 3:} With Assumptions 1, 2 and 3, $\mathbb{E}F(\bm{W}^{t+1})-\mathbb{E}F(\bm{W}^{t})  $ is bounded as follows:

\begin{equation}\label{eq:appTH2Final_lemma3}
\begin{aligned}
\mathbb{E}F(\bm{W}^{t+1})-\mathbb{E}F(\bm{W}^{t}) 
& \leq  ( -\eta  +\frac{L}{2} \eta^2 +  \frac{L}{2}\epsilon \eta^2) \mathbb{E}\| \nabla F(\bm{W}^{t})\|^2 + 2( \epsilon +1)\epsilon \eta^2 \mathbb{E}\left\|   \nabla F(\bm{W}^{t-1})  \right\|^2    \\
& \qquad \qquad +  \frac{L}{2}(1+ \eta^2 L^2) 2( \epsilon +1)\epsilon \frac{1}{N}\sum_{n=1}^{N}\sigma_n^2 \eta^2 + \frac{L}{2}\epsilon \eta^2 \frac{1}{N}\sum_{n=1}^{N}\sigma_n^2.
\end{aligned}
\end{equation}

\textbf{Lemma 4:} With Assumptions 1, 2 and 3, $\mathbb{E}F(\bar{\bar{\bm{W}}}^{t+1}) -\mathbb{E}F(\bar{\bar{\bm{W}}}^{t}) $ is bounded as follows:

\begin{equation}\label{eq:appTheorem4_aggregation_further_lemma4}
\begin{aligned}
\mathbb{E}F(\bar{\bar{\bm{W}}}^{t+1}) -\mathbb{E}F(\bar{\bar{\bm{W}}}^{t})  \leq \frac{L}{2}\epsilon \eta^2 \frac{1}{N}\sum_{n=1}^{N}\sigma_n^2  +\left(  \frac{L}{2}\epsilon \eta^2    -\eta+ \frac{L}{2}\eta^2\right)\mathbb{E}\left\|  \nabla  F(\bar{\bar{\bm{W}}}^{t})  \right\|^2.\\
\end{aligned}
\end{equation}

\textbf{\emph{Proof} of Lemma 1:}
\begin{equation}\label{}
\begin{aligned}
\frac{L}{2}\mathbb{E} \left \| \bm{W}^{t+1}- \bar{\bar{\bm{W}}}^{t+1}  \right\|^2 &   =  \frac{L}{2}\mathbb{E} \left \|\frac{ \sum_{n=1}^{N}\hat{\bm{W}}_n^{t+1}\odot \bm{M}_n^{t+1} }{\sum_{n=1}^{N}\bm{M}_{n}^{t+1}}  - \frac{\sum_{n=1}^{N} \left(\bm{W}^t - \eta \nabla  F_{n}(\bm{W}^{t})\right) \odot \bm{M}_{n}^{t+1}}{\sum_{n=1}^{N}\bm{M}_{n}^{t+1}} \right \|^2  \\
&  =  \frac{L}{2}\mathbb{E} \left\|\frac{ \sum_{n=1}^{N} \left(\bm{W}_n^{t+1} - \eta \nabla  F_{n}(\bm{W}_{n}^{t+1} ) \right)\odot \bm{M}_n^{t+1} }{\sum_{n=1}^{N}\bm{M}_{n}^{t+1}}  - \frac{\sum_{n=1}^{N} \left(\bm{W}^t - \eta \nabla  F_{n}(\bm{W}^{t}) \right) \odot \bm{M}_{n}^{t+1}}{\sum_{n=1}^{N}\bm{M}_{n}^{t+1}}  \right\|^2   \\
&   =  \frac{L}{2}\mathbb{E} \Bigg\|\frac{ \sum_{n=1}^{N} \left(\bm{W}^t \odot \bm{M}_n^t + \hat{\bm{W}}_n^t \odot (\bm{1} - \bm{M}_n^t) - \eta \nabla  F_{n}(\bm{W}_{n}^{t+1})\right) \odot \bm{M}_n^{t+1}}{\sum_{n=1}^{N}\bm{M}_{n}^{t+1}}  \\
 & \qquad \qquad - \frac{\sum_{n=1}^{N} \left(\bm{W}^t - \eta \nabla  F_{n}(\bm{W}^{t})\right) \odot \bm{M}_{n}^{t+1}}{\sum_{n=1}^{N}\bm{M}_{n}^{t+1}}  \Bigg\|^2   \\
&  = \frac{L}{2}\underbrace{\mathbb{E} \left \| \frac{ \sum_{n=1}^{N}\left( (\bm{W}^t - \hat{\bm{W}}_n^t )\odot \bm{M}_n^t + (\hat{\bm{W}}_n^t -\bm{W}^t ) \right) \odot \bm{M}_{n}^{t+1}}{\sum_{n=1}^{N}\bm{M}_{n}^{t+1}} \right \|^2}_{A1}  \\
& \qquad \qquad +  \frac{L}{2}\underbrace{\mathbb{E} \left\| \frac{ \sum_{n=1}^{N}\left(\eta \nabla  F_{n}(\bm{W}^{t}) - \eta \nabla  F_{n}(\bm{W}_{n}^{t+1})\right) \odot \bm{M}_{n}^{t+1}}{\sum_{n=1}^{N}\bm{M}_{n}^{t+1}} \right\|^2}_{A2}.
\end{aligned}
\end{equation}

\begin{equation}\label{}
\begin{aligned}
A1 &  = \mathbb{E}  \left \| \frac{ \sum_{n=1}^{N} (\bm{W}^t - \hat{\bm{W}}_n^t )\odot \bm{M}_n^t \odot \bm{M}_{n}^{t+1} + (\hat{\bm{W}}_n^t -\bm{W}^t )\odot \bm{M}_{n}^{t+1}  }{\sum_{n=1}^{N}\bm{M}_{n}^{t+1}} \right\|^2\\
 & =  \mathbb{E} \left\| \frac{ \sum_{n=1}^{N} (\bm{W}^t - \hat{\bm{W}}_n^t )\odot \bm{M}_n^t \odot \bm{M}_{n}^{t+1}  }{\sum_{n=1}^{N}\bm{M}_{n}^{t+1}} \right\|^2    +  \mathbb{E} \left \| \frac{ \sum_{n=1}^{N} (\hat{\bm{W}}_n^t -\bm{W}^t )\odot \bm{M}_{n}^{t+1}  }{\sum_{n=1}^{N}\bm{M}_{n}^{t+1}} \right\|^2  \\
 &  = \mathbb{E} \left\| \frac{ \sum_{n=1}^{N} (\bm{W}^t - \hat{\bm{W}}_n^t )\odot \bm{M}_n^t \odot \bm{M}_{n}^{t+1}  }{\sum_{n=1}^{N}\bm{M}_{n}^{t+1}} - \frac{1}{N}\sum_{n=1}^{N} (\bm{W}^t - \hat{\bm{W}}_n^t )\odot \bm{M}_n^t  + \frac{1}{N}\sum_{n=1}^{N} (\bm{W}^t - \hat{\bm{W}}_n^t )\odot \bm{M}_n^t \right\|^2    \\
 & \qquad \qquad + \mathbb{E}  \left\| \frac{ \sum_{n=1}^{N} (\hat{\bm{W}}_n^t -\bm{W}^t )\odot \bm{M}_{n}^{t+1}  }{\sum_{n=1}^{N}\bm{M}_{n}^{t+1}}  - \frac{1}{N}\sum_{n=1}^{N} (\bm{W}^t - \hat{\bm{W}}_n^t )  + \frac{1}{N}\sum_{n=1}^{N} (\bm{W}^t - \hat{\bm{W}}_n^t ) \right\|^2  \\
& =  \mathbb{E} \left\|  \frac{ \sum_{n=1}^{N} (\bm{W}^t - \hat{\bm{W}}_n^t )\odot \bm{M}_n^t \odot \bm{M}_{n}^{t+1}  }{\sum_{n=1}^{N}\bm{M}_{n}^{t+1}} - \frac{1}{N}\sum_{n=1}^{N} (\bm{W}^t - \hat{\bm{W}}_n^t )\odot \bm{M}_n^t \right\|^2  +  \mathbb{E}  \left\|  \frac{1}{N}\sum_{n=1}^{N} (\bm{W}^t - \hat{\bm{W}}_n^t )\odot \bm{M}_n^t  \right\|^2\\
 & \qquad \qquad + \mathbb{E}   \left \| \frac{ \sum_{n=1}^{N} (\hat{\bm{W}}_n^t -\bm{W}^t )\odot \bm{M}_{n}^{t+1}  }{\sum_{n=1}^{N}\bm{M}_{n}^{t+1}}  - \frac{1}{N}\sum_{n=1}^{N} (\bm{W}^t - \hat{\bm{W}}_n^t ) \right\|^2  + \mathbb{E}  \left\|  \frac{1}{N}\sum_{n=1}^{N} (\bm{W}^t - \hat{\bm{W}}_n^t ) \right\|^2\\
& \leq (\epsilon +1) \mathbb{E} \left\|  \frac{1}{N}\sum_{n=1}^{N} (\bm{W}^t - \hat{\bm{W}}_n^t )\odot \bm{M}_n^t  \right\|^2   + ( \epsilon +1) \mathbb{E}  \left\|  \frac{1}{N}\sum_{n=1}^{N} (\bm{W}^t - \hat{\bm{W}}_n^t ) \right\|^2  \leq 2( \epsilon +1) \mathbb{E} \left\|  \frac{1}{N}\sum_{n=1}^{N} (\bm{W}^t - \hat{\bm{W}}_n^t ) \right\|^2\\
& =   2( \epsilon +1) \mathbb{E}  \left\|  \frac{\sum_{n=1}^{N}\left( \bm{W}^{t-1}- \eta \nabla F_{n}(\bm{W}^{t-1} )\right) \odot \bm{M}_n^t} {\sum_{n=1}^{N} \bm{M}_n^t}    -  \frac{1}{N}\sum_{n=1}^{N}  \left( \bm{W}^{t-1}  -  \eta \nabla F_{n}(\bm{W}^{t-1} ) \right)    \right\|^2\\
& =  2( \epsilon +1) \mathbb{E} \left\|  \frac{\sum_{n=1}^{N}- \eta \nabla F_{n}(\bm{W}^{t-1} ) \odot \bm{M}_n^t} {\sum_{n=1}^{N} \bm{M}_n^t}    -  \frac{1}{N}\sum_{n=1}^{N}  \left(   -  \eta \nabla F_{n}(\bm{W}^{t-1} ) \right)   \right \|^2 \\
& \leq 2( \epsilon +1)\epsilon  \mathbb{E} \left \| \frac{1}{N}\sum_{n=1}^{N}  \eta \nabla F_{n}(\bm{W}^{t-1}) \right\|^2 
=   2( \epsilon +1)\epsilon  \mathbb{E} \left\| \frac{1}{N}\sum_{n=1}^{N}  \eta \nabla F_{n}(\bm{W}^{t-1})   -   \eta \nabla F(\bm{W}^{t-1})  +\eta \nabla F(\bm{W}^{t-1})  \right\|^2\\
& =  2( \epsilon +1)\epsilon  \mathbb{E}  \left\| \frac{1}{N}\sum_{n=1}^{N}  \eta \nabla F_{n}(\bm{W}^{t-1})   -   \eta \nabla F(\bm{W}^{t-1})  \right\|^2 +  2( \epsilon +1)\epsilon  \mathbb{E}  \left\|  \eta \nabla F(\bm{W}^{t-1}) \right \|^2\\
& \leq 2( \epsilon +1)\epsilon \frac{1}{N}\sum_{n=1}^{N}\sigma_n^2 \eta^2 + 2( \epsilon +1)\epsilon \eta^2  \mathbb{E}\left\|  \nabla F(\bm{W}^{t-1})  \right\|^2.
\end{aligned}
\end{equation}

From the property of the $L$-smoothness $\forall \bm{V}, \bm{W}, \|\nabla  F_{n_k}(\bm{V})-\nabla F_{n_k}(\bm{W})\|\leq L\|\bm{V}-\bm{W}\|$, we have:

\begin{equation}\label{}
\begin{aligned}
A2 &= \mathbb{E} \left\| \frac{ \sum_{n=1}^{N}\left(\eta \nabla  F_{n}(\bm{W}^{t}) - \eta \nabla  F_{n}(\bm{W}_{n}^{t+1}) \right)\odot \bm{M}_{n}^{t+1}}{\sum_{n=1}^{N}\bm{M}_{n}^{t+1}} \right\|^2\\
&\leq \eta^2 L^2 \mathbb{E} \left\| \frac{ \sum_{n=1}^{N} (  \bm{W}^{t} -    \bm{W}_{n}^{t+1}) \odot\bm{M}_{n}^{t+1}}{\sum_{n=1}^{N}\bm{M}_{n}^{t+1}} \right\|^2\\
& =  \eta^2 L^2 \mathbb{E} \left\|    \frac{\sum_{n=1}^{N} \left(  \bm{W}^{t} -  \bm{W}^{t} \odot \bm{M}_{n}^t - \hat{\bm{W}}_{n}^t \odot (\bm{1}-\bm{M}_{n}^t) \right) \odot \bm{M}_{n}^{t+1}}{\sum_{n=1}^{N}\bm{M}_{n}^{t+1}}  \right\|^2 \\
& = \eta^2 L^2 A1.
\end{aligned}
\end{equation}

Combined with the derivation of A1 and A2, $\frac{L}{2}\mathbb{E}\| \bm{W}^{t+1}- \bar{\bar{\bm{W}}}^{t+1}  \|^2 $ can be bounded as follows:

\begin{equation}\label{eq:appA1A2}
\begin{aligned}
\frac{L}{2}\mathbb{E}\| \bm{W}^{t+1}- \bar{\bar{\bm{W}}}^{t+1}  \|^2 & =\frac{L}{2}(A1+A2) \leq  \frac{L}{2}(1+ \eta^2 L^2)A1\\
& \leq \frac{L}{2}(1+ \eta^2 L^2) 2( \epsilon +1)\epsilon \frac{1}{N}\sum_{n=1}^{N}\sigma_n^2 \eta^2 + 2( \epsilon +1)\epsilon \eta^2 \mathbb{E}\left\| \nabla F(\bm{W}^{t-1}) \right\|^2.
\end{aligned}
\end{equation}
$\hfill\qedsymbol$

\textbf{\emph{Proof} of Lemma 2:}
\begin{equation}\label{eq:41w_t1w_t1}
\begin{aligned}
\frac{L}{2}\mathbb{E} \left\| \bar{\bar{\bm{W}}}^{t+1}- \bar{\bm{W}}^{t+1}  \right\|^2 & = \frac{L}{2}\mathbb{E}\left\|  \frac{\sum_{n=1}^{N} (\bm{W}^t - \eta \nabla  F_{n}(\bm{W}^{t})) \odot \bm{M}_{n}^{t+1}}{\sum_{n=1}^{N}\bm{M}_{n}^{t+1}}   -    \frac{1}{N} \sum_{n=1}^{N}(\bm{W}^{t}-\eta  \nabla  F_{n}(\bm{W}^{t}))   \right\|^2\\
& =  \frac{L}{2}\mathbb{E}\left\|  \frac{\sum_{n=1}^{N} ( - \eta \nabla  F_{n}(\bm{W}^{t})) \odot \bm{M}_{n}^{t+1}}{\sum_{n=1}^{N}\bm{M}_{n}^{t+1}}   -    \frac{1}{N} \sum_{n=1}^{N}(-\eta  \nabla  F_{n}(\bm{W}^{t}))   \right\|^2\\
&\leq  \frac{L}{2}\epsilon \mathbb{E}\left\|  \frac{1}{N} \sum_{n=1}^{N}\eta  \nabla  F_{n}(\bm{W}^{t}) \right\|^2\\
& =  \frac{L}{2}\epsilon \mathbb{E}\left\|  \frac{1}{N} \sum_{n=1}^{N}\left(\eta  \nabla  F_{n}(\bm{W}^{t})  +  \eta  \nabla  F(\bm{W}^{t})   - \eta  \nabla  F(\bm{W}^{t}) \right) \right\|^2\\
& = \frac{L}{2}\epsilon \eta^2  \mathbb{E}\left\| \frac{1}{N} \sum_{n=1}^{N}\left( \nabla  F_{n}(\bm{W}^{t})  -    \nabla  F(\bm{W}^{t})\right) \right\|^2   +  \frac{L}{2}\epsilon \eta^2  \mathbb{E}\left\|   \nabla  F(\bm{W}^{t}) \right\|^2\\
& \leq \frac{L}{2}\epsilon \eta^2 \frac{1}{N}\sum_{n=1}^{N}\sigma_n^2  +  \frac{L}{2}\epsilon \eta^2 \mathbb{E} \left\|   \nabla  F(\bm{W}^{t}) \right\|^2.
\end{aligned}
\end{equation}
$\hfill\qedsymbol$

\textbf{\emph{Proof} of Lemma 3:}
Plugging Eq. (\ref{eq:appWWT}), Eq. (\ref{eq:appWWT2L}), Eq. (\ref{eq:appA1A2}) and Eq. (\ref{eq:41w_t1w_t1}) into Eq. (\ref{eq:app_aggregation}), we have

\begin{equation}\label{eq:appTH2Final}
\begin{aligned}
\mathbb{E}F(\bm{W}^{t+1})-\mathbb{E}F(\bm{W}^{t}) 
& \leq \mathbb{E}\left((\bar{\bm{W}}^{t+1}-\bm{W}^{t})^{\rm T} \nabla F(\bm{W}^{t})\right) +  \frac{L}{2}\mathbb{E}\left\| \bm{W}^{t+1}- \bar{\bar{\bm{W}}}^{t+1} \right\|^2 \\
& \qquad \qquad + \frac{L}{2}\mathbb{E}\left\| \bar{\bar{\bm{W}}}^{t+1}- \bar{\bm{W}}^{t+1} \right\|^2 + \frac{L}{2}\mathbb{E}\left\| \bar{\bm{W}}^{t+1}- \bm{W}^{t}  \right\|^2\\
&\leq -\eta \mathbb{E}\left\| \nabla F(\bm{W}^{t})\right\|^2   +  \frac{L}{2}(1+ \eta^2 L^2) 2( \epsilon +1)\epsilon \frac{1}{N}\sum_{n=1}^{N}\sigma_n^2 \eta^2 + 2( \epsilon +1)\epsilon \eta^2 \mathbb{E}\left\|   \nabla F(\bm{W}^{t-1})  \right\|^2  \\
& \qquad \qquad + \frac{L}{2}\epsilon \eta^2 \frac{1}{N}\sum_{n=1}^{N}\sigma_n^2  +  \frac{L}{2}\epsilon \eta^2 \mathbb{E}\left\|   \nabla  F(\bm{W}^{t}) \right\|^2 +  \frac{L}{2} \eta^2 \mathbb{E}\left\| \nabla  F(\bm{W}^t)  \right\|^2 \\
& = ( -\eta  +\frac{L}{2} \eta^2 +  \frac{L}{2}\epsilon \eta^2) \mathbb{E}\| \nabla F(\bm{W}^{t})\|^2 + 2( \epsilon +1)\epsilon \eta^2 \mathbb{E}\left\|   \nabla F(\bm{W}^{t-1})  \right\|^2    \\
& \qquad \qquad +  \frac{L}{2}(1+ \eta^2 L^2) 2( \epsilon +1)\epsilon \frac{1}{N}\sum_{n=1}^{N}\sigma_n^2 \eta^2 + \frac{L}{2}\epsilon \eta^2 \frac{1}{N}\sum_{n=1}^{N}\sigma_n^2.
\end{aligned}
\end{equation}
$\hfill\qedsymbol$

\textbf{\emph{Proof} of Lemma 4:} Continue to use the definitions of Eq. (\ref{eq:appbar}) and Eq. (\ref{eq:appbarbar}). From the property of the $L$-smoothness, we have:

\begin{equation}\label{eq:appTheorem4smoo1}
\begin{aligned}
F(\bar{\bar{\bm{W}}}^{t+1}) \leq F(\bar{\bm{W}}^{t+1}) + (\bar{\bar{\bm{W}}}^{t+1}-\bar{\bm{W}}^{t+1})^{\rm T} \nabla F(\bar{\bm{W}}^{t+1}) + \frac{L}{2}\left\| \bar{\bar{\bm{W}}}^{t+1}- \bar{\bm{W}}^{t+1}\right\|^2.
\end{aligned}
\end{equation}

\begin{equation}\label{eq:appTheorem4smoo2}
\begin{aligned}
F(\bar{\bm{W}}^{t+1}) \leq F(\bar{\bar{\bm{W}}}^{t}) + (\bar{\bm{W}}^{t+1}-\bar{\bar{\bm{W}}}^{t})^{\rm T} \nabla F(\bar{\bar{\bm{W}}}^{t}) + \frac{L}{2}\left\| \bar{\bm{W}}^{t+1}- \bar{\bar{\bm{W}}}^{t}\right\|^2.
\end{aligned}
\end{equation}

Taking the expectations of both sides of Eq. (\ref{eq:appTheorem4smoo1}), Eq. (\ref{eq:appTheorem4smoo2}) and Eq. (\ref{eq:appsmoo3}), we have the three following equations:

\begin{equation}\label{eq:appTheorem45}
\begin{aligned}
\mathbb{E}F(\bar{\bar{\bm{W}}}^{t+1}) \leq \mathbb{E}F(\bar{\bm{W}}^{t+1})  + \frac{L}{2}\mathbb{E}\left\| \bar{\bar{\bm{W}}}^{t+1}- \bar{\bm{W}}^{t+1}  \right\|^2.
\end{aligned}
\end{equation}

\begin{equation}\label{eq:appTheorem46}
\begin{aligned}
\mathbb{E}F(\bar{\bm{W}}^{t+1}) \leq \mathbb{E}F(\bar{\bar{\bm{W}}}^{t}) + \mathbb{E}\left((\bar{\bm{W}}^{t+1}-\bar{\bar{\bm{W}}}^{t})^{\rm T} \nabla F(\bar{\bar{\bm{W}}}^{t})\right) + \frac{L}{2}\mathbb{E}\left\| \bar{\bm{W}}^{t+1}- \bar{\bar{\bm{W}}}^{t}  \right\|^2.\\
\end{aligned}
\end{equation}

Adding the left and right sides of Eq. (\ref{eq:appTheorem45}) and Eq. (\ref{eq:appTheorem46}) respectively, we get

\begin{equation}\label{eq:appTheorem4_aggregation}
\begin{aligned}
\mathbb{E}F(\bar{\bar{\bm{W}}}^{t+1}) -\mathbb{E}F(\bar{\bar{\bm{W}}}^{t})  \leq \frac{L}{2}\mathbb{E}\left\| \bar{\bar{\bm{W}}}^{t+1}- \bar{\bm{W}}^{t+1}  \right\|^2 + \mathbb{E}\left((\bar{\bm{W}}^{t+1}-\bar{\bar{\bm{W}}}^{t})^{\rm T} \nabla F(\bar{\bar{\bm{W}}}^{t})\right) + \frac{L}{2}\mathbb{E}\left\| \bar{\bm{W}}^{t+1}- \bar{\bar{\bm{W}}}^{t}  \right\|^2.\\
\end{aligned}
\end{equation}

The relation between $\bar{\bm{W}}^{t+1}$ and $\bar{\bar{\bm{W}}}^{t} $ can be shown as follows:

\begin{equation}\label{}
\begin{aligned}
\bar{\bm{W}}^{t+1} -  \bar{\bar{\bm{W}}}^{t}  = - \frac{1}{N} \sum_{{n}=1}^{N} \eta \nabla  F_{n}(\bar{\bar{\bm{W}}}^{t}).\\
\end{aligned}
\end{equation}

We first bound the second and third terms in Eq. (\ref{eq:appTheorem4_aggregation}):

\begin{equation}\label{eq:appTheorem42nd}
\begin{aligned}
\mathbb{E}\left((\bar{\bm{W}}^{t+1}-\bar{\bar{\bm{W}}}^{t})^{\rm T} \nabla F(\bar{\bar{\bm{W}}}^{t})\right) = \mathbb{E} \left(- \frac{1}{N} \sum_{{n}=1}^{N} \eta \nabla  F_{n}(\bar{\bar{\bm{W}}}^{t} )^{\rm T} \nabla F(\bar{\bar{\bm{W}}}^{t})\right) = -\eta\mathbb{E} \left\|\nabla F(\bar{\bar{\bm{W}}}^{t}) \right\|^2.\\
\end{aligned}
\end{equation}

\begin{equation}\label{eq:appTheorem43rd}
\begin{aligned}
\frac{L}{2}\mathbb{E}\left\| \bar{\bm{W}}^{t+1}- \bar{\bar{\bm{W}}}^{t}  \right\|^2 = \frac{L}{2}\mathbb{E}\left\|  - \frac{1}{N} \sum_{{n}=1}^{N} \eta \nabla  F_{n}(\bar{\bar{\bm{W}}}^{t})  \right\|^2 = \frac{L}{2}\eta^2\mathbb{E}\left \|  \nabla  F(\bar{\bar{\bm{W}}}^{t})  \right\|^2.\\
\end{aligned}
\end{equation}

Combining Eq. (\ref{eq:appTheorem42nd}), Eq. (\ref{eq:appTheorem43rd}) and \textbf{Lemma 2}, we have the following bound:

\begin{equation}\label{eq:appTheorem4_aggregation_further}
\begin{aligned}
\mathbb{E}F(\bar{\bar{\bm{W}}}^{t+1}) -\mathbb{E}F(\bar{\bar{\bm{W}}}^{t})  \leq \frac{L}{2}\epsilon \eta^2 \frac{1}{N}\sum_{n=1}^{N}\sigma_n^2  +  \frac{L}{2}\epsilon \eta^2 \mathbb{E}\left\|   \nabla  F(\bar{\bar{\bm{W}}}^{t})\right \|^2   -\eta\mathbb{E}\left\| \nabla F(\bar{\bar{\bm{W}}}^{t}) \right\|^2 + \frac{L}{2}\eta^2\mathbb{E}\left\|  \nabla  F(\bar{\bar{\bm{W}}}^{t})  \right\|^2.\\
\end{aligned}
\end{equation}
$\hfill\qedsymbol$

\textbf{\emph{Proof} of Theorem 2:}

Combining Lemma 3 and Lemma 4, we have the following bound:

\begin{equation}\label{eq:prooft0}
\begin{aligned}
\mathbb{E}F(\bm{W}^{kh}) -\mathbb{E}F(\bm{W}^{(k-1)h})
& = \sum_{t=(k-1)h+1}^{kh-1}\left(\mathbb{E}F(\bm{W}^{t+1})-\mathbb{E}F(\bm{W}^{t})\right) + \mathbb{E}F(\bm{W}^{(k-1)h+1}) -\mathbb{E}F(\bm{W}^{(k-1)h})\\
& \leq \sum_{t=(k-1)h+1}^{kh-1} \left( \left( -\eta  +\frac{L}{2} \eta^2 +  \frac{L}{2}\epsilon \eta^2 \right) \mathbb{E}\| \nabla F(\bm{W}^{t})\|^2 + 2( \epsilon +1)\epsilon \eta^2 \mathbb{E}\left\|   \nabla F(\bm{W}^{t-1})  \right\|^2\right)  \\
& \qquad\qquad +  \sum_{t=(k-1)h+1}^{kh-1}\frac{L}{2}\epsilon \eta^2 \frac{1}{N}\sum_{n=1}^{N}\sigma_n^2\left( 2\epsilon + 2\epsilon\eta^2 L^2 +  2\eta^2 L^2  +3 \right)  \\
 & \qquad\qquad + \left( -\eta  +\frac{L}{2} \eta^2 +  \frac{L}{2}\epsilon \eta^2 \right)\mathbb{E}\left\|  \nabla  F(\bm{W}^{(k-1)h})  \right\|^2 + \frac{L}{2}\epsilon \eta^2 \frac{1}{N}\sum_{n=1}^{N}\sigma_n^2 \\
& =  \left( -\eta  +\frac{L}{2} \eta^2 +  \frac{L}{2}\epsilon \eta^2 \right) \sum_{t=(k-1)h}^{kh-1} \mathbb{E}\| \nabla F(\bm{W}^{t})\|^2 +  2( \epsilon +1)\epsilon \eta^2 \sum_{t=(k-1)h+1}^{kh-1} \mathbb{E}\left\|   \nabla F(\bm{W}^{t-1})  \right\|^2 \\
& \qquad\qquad + \frac{L}{2}\epsilon \eta^2 \frac{1}{N}\sum_{n=1}^{N}\sigma_n^2 (h-1) \left( 2\epsilon + 2\epsilon\eta^2 L^2 +  2\eta^2 L^2  +3 \right) + \frac{L}{2}\epsilon \eta^2 \frac{1}{N}\sum_{n=1}^{N}\sigma_n^2 
\end{aligned}
\end{equation}

When $\eta$ is small sufficiently, and $\eta < \frac{2}{L+L\epsilon + 4(\epsilon + 1)\epsilon}$, Eq. (\ref{eq:prooft0}) can be rewritten as follows:

\begin{equation}\label{eq:prooft0_AB}
\begin{aligned}
\frac{\mathbb{E}F(\bm{W}^{kh}) -\mathbb{E}F(\bm{W}^{(k-1)h})}{A+B} 
& \geq \frac{A \sum_{t=(k-1)h}^{kh-1} \mathbb{E}\| \nabla F(\bm{W}^{t})\|^2 + B  \sum_{t=(k-1)h+1}^{kh-1} \mathbb{E}\left\|   \nabla F(\bm{W}^{t-1})  \right\|^2 + C  }{A+B}\\
& = \frac{A \sum_{t=(k-1)h}^{kh-1} \mathbb{E}\| \nabla F(\bm{W}^{t})\|^2 + B  \sum_{t=(k-1)h}^{kh-2} \mathbb{E}\left\|   \nabla F(\bm{W}^{t})  \right\|^2 + C  }{A+B}\\
& = \frac{(A +B) \sum_{t=(k-1)h}^{kh-1} \mathbb{E}\| \nabla F(\bm{W}^{t})\|^2 - B  \mathbb{E}\left\|   \nabla F(\bm{W}^{kh-1})  \right\|^2  + C  }{A+B}\\
&\geq \frac{(A +B) \sum_{t=(k-1)h}^{kh-1} \mathbb{E}\| \nabla F(\bm{W}^{t})\|^2 + C  }{A+B}\\
\end{aligned}
\end{equation}
where $A =  -\eta  +\frac{L}{2} \eta^2 +  \frac{L}{2}\epsilon \eta^2 $, $B = 2( \epsilon +1)\epsilon \eta^2 $, and $C= \frac{L}{2}\epsilon \eta^2 \frac{1}{N}\sum_{n=1}^{N}\sigma_n^2 (h-1) \left( 2\epsilon + 2\epsilon\eta^2 L^2 +  2\eta^2 L^2  +3 \right) + \frac{L}{2}\epsilon \eta^2 \frac{1}{N}\sum_{n=1}^{N}\sigma_n^2 $. Considering $\eta < \frac{2}{L+L\epsilon + 4(\epsilon + 1)\epsilon}$, it is easy to derive that $A < 0$ and $A+B<0$.

By summing Eq. (\ref{eq:prooft0_AB}) over the $k=1,2,3,...,K$, we have
\begin{equation}\label{eq:prooft0_ABsum}
\begin{aligned}
\frac{\mathbb{E}F(\bm{W}^{Kh}) - F(\bm{W}^{0})}{A+B} 
& \geq \sum_{k=1}^{K}\sum_{t=(k-1)h}^{kh-1} \mathbb{E}\| \nabla F(\bm{W}^{t})\|^2 
+ \sum_{k=1}^{K}\frac{C}{A+B}
\end{aligned}
\end{equation}

Dividing both sides of Eq. (\ref{eq:prooft0_ABsum}) by $Kh$, we have

\begin{equation}\label{eq:prooft0_ABsumDividedN}
\begin{aligned}
\frac{1}{Kh} \sum_{k=1}^{K}\sum_{t=(k-1)h}^{kh-1} \mathbb{E}\| \nabla F(\bm{W}^{t})\|^2 
& \leq \frac{F(\bm{W}^{0}) -\mathbb{E}F(\bm{W}^{Kh}) }{-Kh\left(A+B\right)} + \frac{C}{-h(A+B)}\\
& \leq \frac{F(\bm{W}^{0}) - F(\bm{W}^{*}) }{-Kh\left(-\eta  +\frac{L}{2} \eta^2 +  \frac{L}{2}\epsilon \eta^2+2( \epsilon +1)\epsilon \eta^2\right)}\\
& \qquad\qquad + \frac{\frac{L}{2}\epsilon \eta^2 \frac{1}{N}\sum_{n=1}^{N}\sigma_n^2 (h-1) \left( 2\epsilon + 2\epsilon\eta^2 L^2 +  2\eta^2 L^2  +3 \right) + \frac{L}{2}\epsilon \eta^2 \frac{1}{N}\sum_{n=1}^{N}\sigma_n^2}{-h\left(-\eta  +\frac{L}{2} \eta^2 +  \frac{L}{2}\epsilon \eta^2+2( \epsilon +1)\epsilon \eta^2\right)}\\
& = \frac{2\left(F(\bm{W}^{0}) - F(\bm{W}^{*})\right) }{Kh\left(2\eta  -L \eta^2-  L\epsilon \eta^2- 4( \epsilon +1)\epsilon \eta^2\right)}\\
& \qquad\qquad + \frac{L\epsilon \eta^2 \frac{1}{N}\sum_{n=1}^{N}\sigma_n^2 (h-1) \left( 2\epsilon + 2\epsilon\eta^2 L^2 +  2\eta^2 L^2  +3 \right) + L\epsilon \eta^2 \frac{1}{N}\sum_{n=1}^{N}\sigma_n^2}{h\left(2\eta  -L \eta^2-  L\epsilon \eta^2- 4( \epsilon +1)\epsilon \eta^2\right)}\\
\end{aligned}
\end{equation}

We can complete the proof of Theorem 2 by replacing $Kh$ with $T$. $\hfill\qedsymbol$\\

\end{document}